\documentclass[11pt]{article}

\usepackage[T1]{fontenc}
\pdfmapline{=ecss0900 SFSS0900 "T1Encoding ReEncodeFont" <cm-super-t1.enc <sfss0900.pfb}
\usepackage[utf8]{inputenc}
\usepackage{tgpagella}
\usepackage{mathpazo}
\usepackage[varqu]{inconsolata}
\usepackage{amsmath}
\usepackage{amssymb}
\usepackage[nopatch=footnote]{microtype}
\usepackage{booktabs}
\usepackage{graphicx}
\usepackage{pdfpages}
\usepackage[percent]{overpic}
\usepackage{tikz}
\usepackage{caption}
\usepackage{subcaption}
\usepackage{tabularx}
\usepackage{array}
\usepackage{multirow}
\usepackage{makecell}
\usepackage[numbers]{natbib}
\usepackage{fvextra}
\usepackage{needspace}
\usepackage{placeins}
\usepackage{tcolorbox}
\tcbuselibrary{skins,breakable}
\usepackage{ovisocr3_report}

\newcolumntype{Y}{>{\raggedright\arraybackslash}X}

\newcommand{\ours}{OvisOCR3}

\newif\ifcrosspagereview
\crosspagereviewfalse

\newcommand{\up}{\(\uparrow\)}

\newcommand{\bestcell}[1]{\textbf{#1}}
\newcommand{\secondcell}[1]{\underline{#1}}
\newcounter{taskprompt}
\newcommand{\inputprompt}[3]{%
  \par\Needspace{4\baselineskip}%
  \refstepcounter{taskprompt}\label{#1}%
  \begin{tcolorbox}[enhanced,breakable,lines before break=4,sharp corners,
    title after break={Prompt (continued)},
    colback=white,colbacktitle=black!5,colframe=ovisgray,colupper=black,boxrule=0.4pt,boxsep=0pt,
    left=2mm,right=2mm,top=3mm,bottom=1.5mm,before skip=6pt,after skip=6pt,
    title=Prompt,fonttitle=\normalfont\small,coltitle=ovisgray,
    attach boxed title to top left={xshift=0pt,yshift=-\tcboxedtitleheight/2},
    boxed title style={sharp corners,colback=black!5,colframe=ovisgray,
      boxrule=0.4pt,boxsep=0pt,left=2mm,right=2mm,top=0.8mm,bottom=0.8mm}]
  \VerbatimInput[fontsize=\footnotesize,breaklines=true,breakanywhere=true,
    breaksymbolleft={},frame=none,
    listparameters={\setlength{\topsep}{0pt}\setlength{\partopsep}{0pt}%
      \clubpenalty=10000\widowpenalty=10000}]{prompts/#3.txt}%
  \end{tcolorbox}%
  \par
}

\title{OvisOCR3 Technical Report}
\author{
  ATH-MaaS, Alibaba Group\\[0.4em]
  \small Full author list is provided at the end of this report.
}
\date{}

\begin{document}

\maketitle

\begin{abstract}
We introduce \ours{}, a 0.8B model for a broad range of OCR-related tasks. \ours{} unifies document parsing, general OCR, key information extraction (KIE), document figure captioning, and cross-page table reconstruction in a single model. We train the model on real and synthetic data through continued pretraining and use domain-balanced mid-training to balance supervision across tasks with different data volumes and response lengths. Multi-teacher on-policy distillation transfers expertise from five domain-specific 4B teachers into the shared model, followed by mixed-task reinforcement learning with domain-specific rewards. We compare \ours{} with general-purpose vision-language models and specialized OCR models on public and in-house benchmarks covering all five task families. In these comparisons, it achieves an overall score of 96.74 on OmniDocBench v1.6 and the highest overall scores on PureDocBench, Real5-OmniDocBench, and MDPBench. On OCRBench v2, it ranks first, scoring 65.9 and 61.3 on the English and Chinese splits, respectively. It also outperforms the baselines on OCRBench and all four KIE benchmarks. On DocFigCapBench, it achieves the highest localization F1 and a caption score close to that of Qwen3.8-Max. For cross-page tables, it leads in Relation F1 on CrossTabBench and truncated-table TEDS on MPDocBench-Parse. These results demonstrate that a compact 0.8B model can achieve strong performance across diverse OCR-related tasks. \ours{} is available at \url{https://huggingface.co/ATH-MaaS/OvisOCR3}.

\end{abstract}

\begin{figure}[!htbp]
\centering
\includegraphics[width=\linewidth]{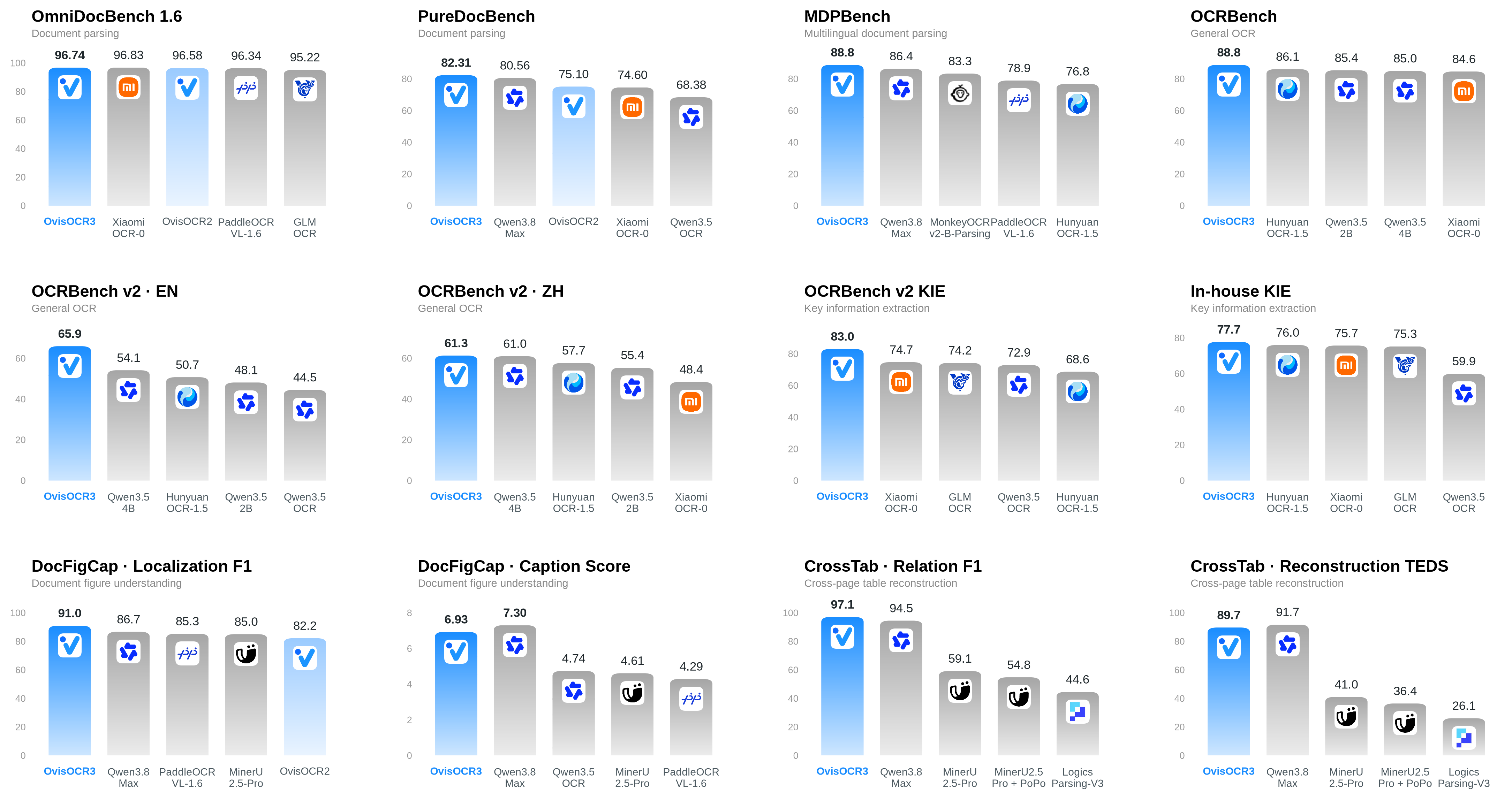}
\caption{Performance of \ours{} across benchmarks for OCR-related tasks.}
\label{fig:homepage_performance}
\end{figure}

\section{Introduction}
\label{sec:introduction}

Optical character recognition (OCR) converts text in images into machine-readable content. In practical applications, however, accurate transcription is only a starting point. Users may need the structure of an entire page, an answer to a question about its content, or a set of fields extracted according to a specified schema~\cite{fu2025ocrbenchv2,yang2024ccocr}. Figures can carry information absent from the surrounding text, while tables spanning multiple pages must be recovered as complete structures~\cite{zhou2026mpdocbench}. These requirements call for models that combine accurate reading with task-directed understanding and structured generation, producing representations suited to different downstream needs.

Our previous work, OvisOCR2~\cite{docparse_ovis2_report}, demonstrated that a compact 0.8B end-to-end model can achieve leading document parsing performance through carefully designed data and post-training. Extending this approach beyond page parsing is not simply a matter of adding more tasks to the training mixture. Different OCR tasks vary substantially in response length and in what constitutes a correct output. A full-page reconstruction can contain thousands of tokens, whereas a field extraction may require only a few; balanced sample counts therefore do not ensure balanced supervision under a token-averaged loss. Moreover, accurate transcription, informative figure descriptions, and correct cross-page table associations require different forms of supervision and feedback. The challenge is to achieve specialist-level performance across a broader range of tasks with a compact model.

We introduce \ours{}, a 0.8B model that unifies document parsing, general OCR, key information extraction, document figure captioning, and cross-page table reconstruction. Our goal is to achieve leading performance across these tasks while retaining a single compact model for deployment. Built on Qwen3.5-0.8B~\cite{qwen2026qwen3508b}, \ours{} performs each task according to the supplied instructions and shares the same parameters across all five domains. Continued pretraining on real and synthetic data builds a common foundation for the five domains. During domain-balanced mid-training, we rebalance the task mixture and train two branches with token-level and sample-level loss normalization. We fuse their weights at the end of this stage. Multi-teacher on-policy distillation (MOPD) then integrates expertise from five domain-specific 4B teachers into the shared student. Finally, mixed-task reinforcement learning (MixRL) jointly refines the model with domain-specific rewards. The teachers are used only during training; inference requires no separate task-specific models.

We evaluate \ours{} on public and in-house benchmarks covering all five task families. In document parsing, it achieves an overall score of 96.74 on OmniDocBench v1.6 and the highest overall scores among the compared models on PureDocBench, Real5-OmniDocBench, and MDPBench. It also leads our comparisons on OCRBench, both language splits of OCRBench v2, and all four KIE benchmarks. On DocFigCapBench, \ours{} achieves the highest localization F1 among the compared models and a caption score close to that of Qwen3.8-Max, leading specialized OCR models on both metrics. For cross-page tables, it achieves the highest relation F1 on CrossTabBench and the highest truncated-table TEDS on MPDocBench-Parse. These results demonstrate that broad OCR task coverage and strong task-specific performance can be combined in a single 0.8B model.

\section{Model Capabilities}
\label{sec:capabilities}

The capabilities of \ours{} span text recognition and localization, document structure recovery, and visual content understanding. We organize them into five task families according to their prediction targets and output representations (Table~\ref{tab:capability_overview}). These families operate at different levels of granularity, from individual text regions to complete pages and tables spanning multiple pages. All five share the same model parameters, with task-specific instructions defining the information to be recovered and its representation.

\begin{table}[!htbp]
\centering
\small
\setlength{\tabcolsep}{4pt}
\renewcommand{\arraystretch}{1.15}
\caption{Task coverage and output representations of \ours{}.}
\label{tab:capability_overview}
\begin{tabularx}{\linewidth}{@{}>{\raggedright\arraybackslash}p{0.21\linewidth}YY@{}}
\toprule
\textbf{Task family} & \textbf{Task scope} & \textbf{Output representation} \\
\midrule
Document parsing & Recovery of page content, layout, and reading order & Region-level JSON or Markdown with LaTeX and HTML \\
General OCR & Recognition, localization, translation, and visual reasoning & Transcriptions, bounding boxes, LaTeX, chart data, labels, or answers \\
Key information extraction & Extraction and organization of fields from images & Key-value records or template-guided JSON \\
Document figure captioning & Localization and description of document figures & Localized captions and Mermaid representations \\
Cross-page table reconstruction & Continuation detection and table reconstruction across pages & HTML tables with source-page associations, or a no-continuation prediction \\
\bottomrule
\end{tabularx}
\end{table}

Tasks involving localization use a common spatial representation: bounding boxes are specified by their upper-left and lower-right corners, with horizontal and vertical coordinates normalized to integers in 0--999 relative to the input image width and height, respectively.

\subsection{Document Parsing}
\label{sec:capability_document_parsing}

Document parsing aims to recover the content and spatial organization of a page from its image. While OvisOCR2 supports only Markdown document parsing, \ours{} introduces two additional capabilities: structured document parsing and document layout detection. Beyond expanding task coverage, \ours{} further improves parsing quality, especially for multilingual documents and photographed pages.

\subsubsection{Structured Document Parsing}

Structured parsing represents a page as a JSON array of region-level records in reading order. Each record associates a bounding box (\texttt{bbox}) with a category from a predefined taxonomy (\texttt{category}) and the corresponding region content (\texttt{content}). The content representation depends on the region category: text uses Markdown, mathematical expressions use LaTeX, and tables use HTML. Images and charts are referenced by HTML image tags whose paths encode their bounding boxes.

\inputprompt{prompt:document_structured}{Structured document parsing}{document_structured}

\subsubsection{Document Layout Detection}

Document layout detection predicts the locations, semantic categories, and reading order of page regions without transcribing their content. Its output consists of JSON records containing \texttt{bbox} and \texttt{category} fields. This formulation shares the coordinate convention and category taxonomy of structured parsing but omits the \texttt{content} field.

\inputprompt{prompt:document_layout}{Document layout detection}{document_layout}

\subsubsection{Markdown Document Parsing}

Markdown document parsing converts a page image into a single Markdown document in reading order, preserving the source text without translation or paraphrasing. Mathematical expressions and tables are represented in LaTeX and HTML, respectively, while images and charts are referenced by coordinate-bearing HTML image tags. Unlike structured parsing, this formulation represents page content as a continuous document rather than a sequence of region-level JSON records, without assigning explicit bounding boxes or category labels to individual text regions.

\inputprompt{prompt:document_markdown}{Markdown document parsing}{document_markdown}
\subsection{General OCR}
\label{sec:capability_general_ocr}

General OCR covers ten tasks spanning text recognition and localization, structured content extraction, translation, and visual understanding. The task formulations differ in both the information supplied with the image and the prediction target, supporting outputs ranging from plain text and spatial coordinates to structured data, category labels, and answers. Table~\ref{tab:general_ocr_interfaces} summarizes these interfaces.

\begin{table}[!htbp]
\centering
\small
\setlength{\tabcolsep}{4pt}
\renewcommand{\arraystretch}{1.12}
\caption{The ten General OCR interfaces. Every task receives an image; the second column lists additional task inputs. Bounding boxes use the order $[x_{\min},y_{\min},x_{\max},y_{\max}]$, with integer coordinates normalized independently by image width and height to $[0,1000)$.}
\label{tab:general_ocr_interfaces}
\begin{tabularx}{\linewidth}{@{}lYY@{}}
\toprule
\textbf{Task} & \textbf{Additional input} & \textbf{Output} \\
\midrule
Plain OCR & None & Plain-text transcription \\
Word-level text spotting & None & JSON records with word boxes and text \\
Reference grounding & Ordered reference list & JSON records with reference boxes \\
Region OCR & Target bounding box & Text within the region \\
Formula recognition & None & LaTeX expressions \\
Chart parsing & None & Chart metadata and nested data values \\
Text translation & Target region and language & Translated text \\
Image classification & Target description, candidate labels & One supplied label \\
Visual question answering & Question, optionally with choices & Reasoning and a boxed final answer \\
Position-aware VQA & Question about visible answer text & Answer and its bounding box \\
\bottomrule
\end{tabularx}
\end{table}
\FloatBarrier

\subsubsection{Plain OCR}

Plain OCR transcribes all readable text from scanned pages, screenshots, signs, handwritten notes, or text crops. It preserves the original wording, language, and reading order, using spaces and line breaks rather than document markup or bounding boxes.

\inputprompt{prompt:plain_ocr}{Plain OCR}{plain_ocr}

\subsubsection{Word-Level Text Spotting}

Word-level text spotting jointly localizes and transcribes each readable word or text token in an image. It returns a JSON array in reading order, with each record pairing a transcription with a bounding box that tightly encloses the corresponding word or token.

\inputprompt{prompt:word_spotting}{Word-level text spotting}{word_spotting}

\subsubsection{Reference Grounding}

Reference grounding localizes image content specified by user-provided references. Unlike text spotting, it takes the target references as input and predicts their spatial locations. The output is a JSON array ordered by the input references. Each record pairs a reference with the bounding box of a corresponding visible region, with separate records when a reference corresponds to multiple regions.

\inputprompt{prompt:reference_grounding}{Reference grounding}{reference_grounding}

\subsubsection{Region OCR}

Region OCR reads the text within a supplied bounding box while retaining the surrounding image as context. The response contains only the text in the requested region. This complements reference grounding: region OCR maps a location to text, whereas reference grounding maps a supplied reference to its location.

\inputprompt{prompt:region_ocr}{Region OCR}{region_ocr}

\subsubsection{Formula Recognition}

Formula recognition transcribes printed or handwritten mathematical expressions into LaTeX, preserving their symbols and spatial structure, including superscripts, subscripts, matrices, and alignments. Each standalone or display expression is enclosed in \verb|$$ ... $$|, with multiple standalone expressions returned in natural reading order.

\inputprompt{prompt:formula_recognition}{Formula recognition}{formula_recognition}

\subsubsection{Chart Parsing}

Chart parsing recovers chart metadata and data values as structured JSON, preserving the organization of series and categories. The output follows a fixed schema with \texttt{title}, \texttt{source}, \texttt{x\_title}, \texttt{y\_title}, and \texttt{values}. Scalar data values are serialized as strings, and unavailable metadata is represented by the literal string \texttt{"None"}. For dual-axis charts, \texttt{y\_title} may contain two titles in visual order.

\inputprompt{prompt:chart_parsing}{Chart parsing}{chart_parsing}

\subsubsection{Text Translation}

Text translation renders the readable text within a specified image region in a target language. It returns only the translated text, preserving the source meaning, names, numbers, symbols, and line order.

\inputprompt{prompt:text_translation}{Text translation}{text_translation}

\subsubsection{Image Classification}

Image classification assigns an input image to one of the user-provided candidate categories. The candidate set defines the label space, and the model returns a single label exactly as supplied.

\inputprompt{prompt:image_classification}{Candidate-label image classification}{image_classification}

\subsubsection{Visual Question Answering}
\label{sec:general_vqa}

Visual question answering (VQA) answers natural-language questions by reading and reasoning over image content. It covers text-centric questions about documents, scene text, interfaces, and handwritten content, together with problems involving diagrams, mathematics, science, and visual reasoning. Questions may be open-ended or include answer choices. The prompt requests reasoning followed by a final answer enclosed in \verb|\boxed{}|.

\inputprompt{prompt:visual_question_answering}{Visual question answering}{visual_question_answering}

\subsubsection{Position-Aware Visual Question Answering}

Position-aware VQA combines question answering with localization of the answer text in the image. Unlike reference grounding, the target text is not specified in advance but must be identified from the question and image. The output is a JSON object with \texttt{answer} and \texttt{bbox} fields. Here, \texttt{bbox} is a string in the form \texttt{"(x1,y1,x2,y2)"}, encoding the normalized coordinates of a tight box around the answer text, rather than the numeric array used by text spotting.

\inputprompt{prompt:position_aware_vqa}{Position-aware visual question answering}{position_aware_vqa}
\subsection{Key Information Extraction}
\label{sec:capability_kie}

Key information extraction (KIE) extracts information from an input image and organizes it into structured records. \ours{} supports open extraction, specified-field extraction, and schema-based extraction, which differ in how the target fields and output structure are specified.
All three modes request JSON-only output, with field values preserved as written in the image. Empty or missing values are represented by empty strings. Open and specified-field extraction use arrays for repeated values, while schema-based extraction follows the supplied JSON structure.

\subsubsection{Open Extraction}

Open extraction identifies key-value pairs in an input image without a predefined field list or output schema. The model determines both the output keys and the organization of the resulting JSON record.

\inputprompt{prompt:kie_open}{Open key information extraction}{kie_open}

\subsubsection{Specified-Field Extraction}

Specified-field extraction takes a user-provided JSON array of field names, such as \texttt{"company"}, \texttt{"date"}, and \texttt{"total"}. The model returns a JSON object containing exactly these keys, with values extracted from the image and empty strings for missing fields.

\inputprompt{prompt:kie_fields}{Specified-field key information extraction}{kie_fields}

\subsubsection{Schema-Based Extraction}

Schema-based extraction specifies both the target fields and their organization through a user-provided JSON template. Empty strings serve as value placeholders, while nested objects and arrays define the output structure. For example, a receipt template can group each item's name, quantity, and price within an array element and place payment totals in a separate object.

\inputprompt{prompt:kie_schema}{Schema-based key information extraction}{kie_schema}
\subsection{Document Figure Captioning}
\label{sec:capability_captioning}

Document figure captioning jointly localizes and describes visual regions in a document page, including photographs, illustrations, charts, and flowcharts. Given the full-page image, \mbox{\ours{}} returns a JSON array of page regions in reading order, using the same \texttt{bbox}, \texttt{category}, and \texttt{content} fields as structured document parsing.

For visual regions, the \texttt{content} field contains an English caption of no more than 100 words enclosed in \verb|<caption>...</caption>|, preserving visible labels in their original language when needed and replacing the image references used in structured parsing. For flowcharts, a Mermaid representation of their nodes and connections may follow within \verb|<mermaid>...</mermaid>|. Text, formulas, and tables retain their Markdown, LaTeX, and HTML representations, respectively.

\inputprompt{prompt:figure_captioning}{Document figure captioning}{figure_captioning}
\subsection{Cross-Page Table Reconstruction}
\label{sec:capability_crosspage}

Cross-page table reconstruction recovers complete tables from fragments distributed across consecutive pages, requiring both continuation detection and reconstruction of cell content and structure. \mbox{\ours{}} performs page parsing and cross-page reconstruction with the same model.

\subsubsection{Pairwise Table Reconstruction}

\begin{figure}[!htbp]
\centering
\includegraphics[width=\linewidth]{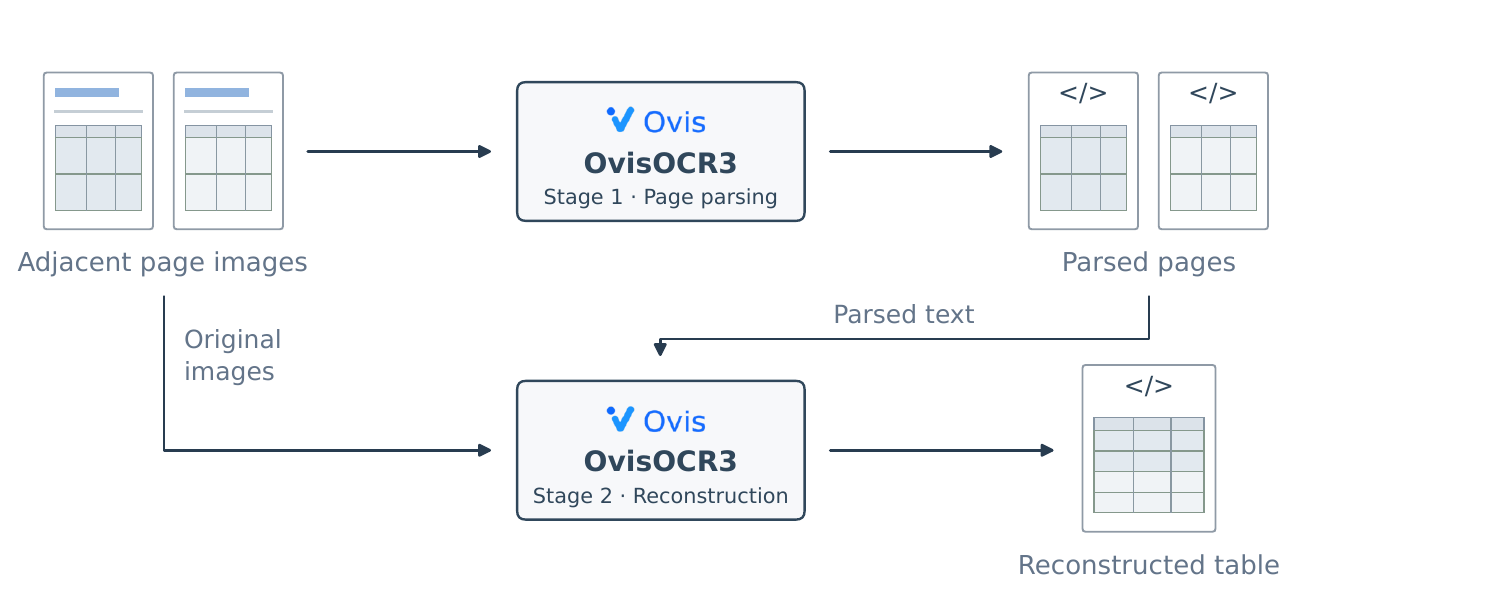}
\caption{Two-stage cross-page table reconstruction with a shared model. Page-level parsing produces Markdown with HTML tables; reconstruction uses these representations together with adjacent page images to identify continuations and recover complete tables.}
\label{fig:crosspage_workflow}
\end{figure}

We use a two-stage procedure (Figure~\ref{fig:crosspage_workflow}). First, \mbox{\ours{}} parses each page independently into Markdown, with tables represented in HTML. Second, the same model takes two adjacent page images and their parsed representations to identify cross-page continuations and reconstruct each continued table. The output is a JSON array whose records contain the page indices within the input pair (\texttt{source\_pages}) and the reconstructed HTML table (\texttt{html}). If no continuation is detected, the model returns an explicit no-continuation result.

\inputprompt{prompt:crosspage_table}{Cross-page table reconstruction}{crosspage_table}

\subsubsection{Recursive Reconstruction}

\begin{figure}[!htbp]
\centering
\includegraphics[width=\linewidth]{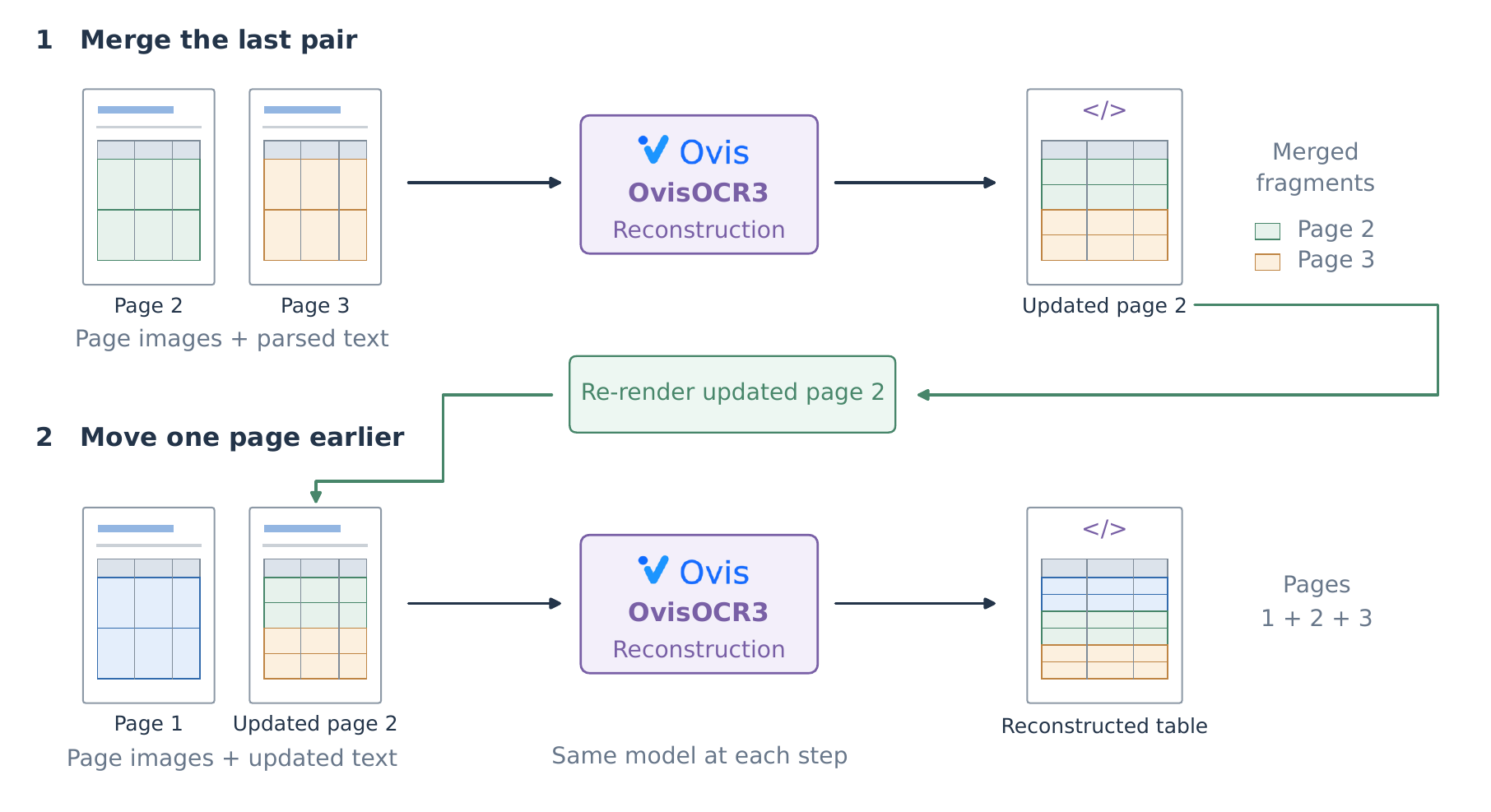}
\caption{Recursive reconstruction of a table spanning three pages. After merging the fragments on pages 2 and 3, the updated Markdown for page 2 is re-rendered and paired with page 1. Colors denote the source pages of the table fragments. Each step uses two page images and their current parsed representations.}
\label{fig:crosspage_recursive}
\end{figure}

For tables spanning more than two pages, we apply pairwise reconstruction recursively within contiguous sequences of pages containing detected tables (Figure~\ref{fig:crosspage_recursive}). We process adjacent pairs from the end of each sequence toward its beginning. After each merge, the reconstructed table is written back to the earlier page's Markdown, which is then re-rendered before that page is paired with its predecessor. Each step uses two page images and their current parsed representations, allowing recovered content to accumulate across multiple pages while retaining the pairwise input format.

\section{Training}
\label{sec:training}

The training recipe of \ours{} builds on the staged learning paradigm of OvisOCR2~\cite{docparse_ovis2_report}, extending its use of supervised learning, on-policy distillation, and model fusion to a unified multi-domain model. The central objective is to develop and jointly refine five complementary capabilities: document parsing, general OCR, key information extraction, document figure captioning, and cross-page table reconstruction. These tasks share a foundation in text recognition and document structure understanding, but place different emphasis on content reconstruction, field extraction, image-text association, and cross-page structure reconstruction. We therefore adopt a staged training process that first builds general capabilities, then integrates domain expertise and jointly refines the model.

We organize training into four stages: continued pretraining builds a common foundation for the five domains; domain-balanced mid-training rebalances supervision across tasks; multi-teacher on-policy distillation (MOPD) transfers domain-specific expertise into a shared student; and mixed-task reinforcement learning (MixRL) jointly refines this student with domain-specific rewards.

\subsection{Stage 1: Continued Pretraining}
\label{sec:training-stage1}

We initialize \ours{} from Qwen3.5-0.8B~\cite{qwen2026qwen3508b} and continue training on a broad mixture of visual-text tasks. The vision encoder, multimodal connector, and language model are updated jointly. Training examples retain their task-specific instructions and target representations, allowing the same autoregressive model to learn short text answers, structured extractions, captions, and long-form reconstructions. This stage provides a common foundation for the five domains before more targeted balancing and post-training.

The Stage~1 mixture contains approximately 72 million sampled training examples. Real and synthetic data span diverse document genres, languages, layouts, and acquisition conditions. Document parsing covers textbooks, reports, news, and handwritten notes, together with synthetic layouts, multilingual pages, and photographed or degraded documents. General OCR includes printed and handwritten text recognition, mathematical expressions, chart and document question answering, text translation, and grounding. Key information extraction covers forms, receipts, invoices, and medical documents with both schema-guided and open-ended extraction targets. Document figure captioning includes figures in real documents and flowcharts, while cross-page table reconstruction includes multi-page tables and positive and negative examples of cross-page relations. Document parsing and general OCR account for approximately 57\% and 29\% of the sampling budget, respectively, with the remaining budget allocated to the other three domains.

We use next-token prediction over reference responses, averaging the loss over supervised response tokens. Sequence packing improves training utilization while preserving the boundaries of individual examples.

\subsection{Stage 2: Domain-Balanced Mid-Training}
\label{sec:training-stage2}

Starting from Stage~1, we rebalance the supervised mixture across the five domains. This is particularly important for a unified model: datasets vary substantially in size, while an individual example may range from a few extracted fields to a lengthy structured response.

The Stage~2 mixture contains approximately 8.8 million sampled examples, with each domain contributing about 16--23\% of the sample budget. It retains the five-domain coverage while incorporating challenging cases such as multilingual photographed pages, small or densely arranged figures, extraction targets with empty fields, and hard negative cross-page relations.

However, more balanced sample counts do not imply balanced supervision at the token level. General OCR contributes approximately 22.8\% of examples but only 5.2\% of supervised response tokens; key information extraction contributes 16.1\% of examples but only 2.5\% of those tokens. The longer responses in document parsing, figure captioning, and cross-page table reconstruction therefore still dominate the supervised token budget.

We train two branches with token-level and sample-level loss normalization. Let $\mathcal{B}$ index the valid examples in a training batch, $x_i$ and $y_i$ denote the input and reference response, and $S_i$ contain supervised response positions. The token-level loss is:
\begin{equation}
  \mathcal{L}_{\mathrm{token}}(\theta)
  =-\frac{\sum_{i\in\mathcal{B}}\sum_{t\in S_i}
  \log p_\theta(y_{i,t}\mid x_i,y_{i,<t})}{\sum_{i\in\mathcal{B}} |S_i|}.
  \label{eq:training-token-loss}
\end{equation}
Here, $p_\theta$ is the model's next-token distribution with parameters $\theta$, and $y_{i,t}$ and $y_{i,<t}$ denote the current token and its preceding response prefix. The sample-level loss averages each example's token loss:
\begin{equation}
  \mathcal{L}_{\mathrm{sample}}(\theta)
  =-\frac{1}{|\mathcal{B}|}
   \sum_{i\in\mathcal{B}}\frac{1}{|S_i|}
   \sum_{t\in S_i}\log p_\theta(y_{i,t}\mid x_i,y_{i,<t}).
  \label{eq:training-sample-loss}
\end{equation}
Token-level normalization gives longer responses more weight, whereas sample-level normalization weights valid examples equally. Under sequence packing, sample-level loss follows the original example boundaries. Empirically, sample-level loss outperforms token-level loss on short-response tasks, including general OCR and key information extraction.

At the end of Stage~2, we merge the weights of the two independently trained branches to initialize subsequent post-training.

\subsection{Stage 3: Multi-Teacher On-Policy Distillation}
\label{sec:training-stage3}

We train five domain-specific 4B teachers, one for each domain. Each teacher is initialized from Qwen3.5-4B~\cite{qwen2026qwen354b} and undergoes supervised fine-tuning on its corresponding domain, followed by reinforcement learning with domain-specific rewards.

Following on-policy distillation (OPD)~\cite{agarwal2024onpolicy} and the recipe of OvisOCR2~\cite{docparse_ovis2_report}, we distill these teachers into a single 0.8B student initialized from the merged Stage~2 model. For an input $x_i$ from domain $d_i$, the student generates a response $y_i$, and the corresponding teacher $q_{d_i}$ provides token-level supervision conditioned on the same input and student-generated prefix. Each example is supervised only by its domain teacher, while all domains update the shared student.

Our MOPD experiments compare two implementations: policy-gradient optimization and top-$K$ distillation. The policy-gradient implementation uses the teacher--student log-probability difference on each sampled token as feedback for a clipped policy-gradient objective~\cite{schulman2017ppo}. Top-$K$ distillation~\cite{li2026rethinkingopd} instead directly matches the teacher and student distributions over multiple candidate tokens at each student-generated prefix. In our experiments, top-$K$ distillation achieves better overall performance, particularly on document parsing and general OCR tasks. Relative preferences over multiple candidates may help the capacity-constrained 0.8B student better absorb teacher knowledge for coordinated text and structure generation. We therefore adopt this implementation for Stage~3.

For top-$K$ distillation, let $\bar p_{\theta,i,t}$ and $\bar q_{d_i,i,t}$ denote the student and teacher distributions at response position $t$, renormalized over the rollout student's top-$K$ tokens. This support remains fixed during each update. We minimize the response-token-averaged reverse KL divergence, with a per-position cap $c$:
\begin{equation}
  \mathcal{L}_{\mathrm{MOPD}}(\theta)
  =\frac{\sum_i\sum_{t=1}^{|y_i|}
    \min\!\left\{D_{\mathrm{KL}}
    \!\left(\bar p_{\theta,i,t}\Vert\bar q_{d_i,i,t}\right),c\right\}}
    {\sum_i |y_i|}.
  \label{eq:training-mopd}
\end{equation}

\subsection{Stage 4: Mixed-Task Reinforcement Learning}
\label{sec:training-stage4}

After teacher-guided capability integration, MixRL jointly optimizes the model across the five domains using task-specific rewards and a domain-balanced input mixture.

\paragraph{Domain-specific rewards.}
MixRL brings diverse tasks into a common optimization process while preserving the criteria that define success in each domain. For example, text recognition can be assessed directly against a reference transcription, whereas cross-page table reconstruction requires correct table associations before reconstruction quality can be evaluated. Domain-specific rewards encode these distinctions, allowing joint optimization to respect both output fidelity and task-specific structural dependencies:
\begin{itemize}
  \item \textbf{Document parsing.} The reward jointly measures text fidelity, table structure and content, and the visual consistency of mathematical expressions. Text is scored by one minus normalized edit distance~\cite{levenshtein1965binary}, tables by Tree-Edit-Distance-based Similarity (TEDS)~\cite{pubtabnet}, and formulas by Character Detection Matching (CDM)~\cite{cdm} on their rendered representations. Scores are aggregated and normalized over the components present in the reference document, encouraging accurate transcription together with faithful table topology and mathematical notation.
  \item \textbf{General OCR.} Rewards target recognition accuracy, answer correctness, or spatial alignment according to the subtask. Text recognition combines edit similarity with exact matching to reward both partial improvements and fully correct transcriptions. Question answering uses answer-type-aware matching for text, numeric values, and categorical answers. Localization uses one-to-one matching with overlap-based scores and duplicate-prediction penalties; text spotting additionally combines hard matching scores with continuous IoU feedback. Formula and table recognition use CDM and TEDS, respectively, to assess their content and structure.
  \item \textbf{Key information extraction.} Field-value matching forms the main reward component, combined with field-type and list-structure consistency and modulated by field-path coverage. This design evaluates both which fields are extracted and whether their values are correct, rather than rewarding a correct schema alone. Additional penalties target incorrect filling of empty fields, omission of nonempty fields, values assigned to the wrong fields, and excessively long extractions. Together, these terms encourage accurate extraction while distinguishing content that should be retained, left empty, or excluded.
  \item \textbf{Document figure captioning.} The reward focuses on identifying and spatially grounding the figures to be described, using a weighted combination of figure-matching F1 and geometric IoU credit. F1 penalizes both missed and spurious detections, while the geometric term provides graded feedback for imperfect localization. Their combination encourages complete figure identification and accurate spatial extent, including localization improvements below the hard matching threshold.
  \item \textbf{Cross-page table reconstruction.} We distinguish relation identification from reconstruction quality. Relation-oriented examples use F1 to penalize both missed and incorrect cross-page associations. Reconstruction-oriented examples follow a relation-first reward: TEDS evaluates the merged table's structure and content only when all candidate responses in the group correctly identify the cross-page relations; otherwise, the reward reflects relation correctness. For page-level reconstruction, the quality score also incorporates F1 over non-cross-page content to discourage omissions outside the reconstructed table.
\end{itemize}

\paragraph{Joint policy optimization.}
We use group-relative policy optimization (GRPO)~\cite{shao2024deepseekmath} to jointly optimize the shared model with domain-specific rewards. Advantages are normalized within each response group, allowing heterogeneous reward signals to guide a unified policy.

Response-level advantages guide token generation through a clipped policy objective. An independent KL regularizer constrains policy drift, balancing task improvement with training stability.

\section{Evaluation}
\label{sec:evaluation}

\subsection{Document Parsing}
\label{sec:eval_document_parsing}

We evaluate the document parsing capability of \ours{} on four public benchmarks: OmniDocBench v1.6, PureDocBench, Real5-OmniDocBench, and MDPBench. Together, they assess general page parsing, robustness to image degradation, and multilingual parsing. \ours{} uses the \hyperref[prompt:document_markdown]{Markdown document parsing prompt} across all four benchmarks.

We compare \ours{} with both general-purpose vision-language models (VLMs) and specialized OCR models. Locally evaluated baselines are marked with $\dagger$, while other baseline scores are taken from the cited reports or benchmark leaderboards. In each column, bold and underlined values denote the best and second-best results, respectively, including ties. Parameter counts follow the cited sources: a slash separates total and active parameters for mixture-of-experts (MoE) models, and a dash indicates that the parameter count is not reported.

\subsubsection{General Document Parsing}

OmniDocBench v1.6~\cite{ouyang2025omnidocbench,docparse_odb_leaderboard} contains 1,651 pages with diverse document layouts. We report normalized text edit distance, formula CDM, table TEDS, structure-only TEDS-S, and reading-order edit distance. The Overall score averages text accuracy (one minus normalized edit distance), formula CDM, and table TEDS after expressing all three on a percentage scale. TEDS-S and reading-order edit distance separately assess table structure and reading order and are not included in Overall.

\begin{table}[!htbp]
\centering
\fontsize{9}{11.2}\selectfont
\setlength{\tabcolsep}{2pt}
\renewcommand{\arraystretch}{1.08}
\caption{Document parsing results on OmniDocBench v1.6. Baseline scores without $\dagger$ are taken from the official leaderboard~\cite{docparse_odb_leaderboard} and model reports~\cite{docparse_ovis2_report,docparse_hunyuan_report,docparse_monkey_report,docparse_xiaomi_report}. RO denotes reading order.}
\label{tab:docparse_odb}
\begin{tabular*}{\linewidth}{@{\extracolsep{\fill}}llcrrrrrr@{}}
\toprule
\textbf{Model Type} & \textbf{Model} & \textbf{Parameters} & \textbf{Overall}$\uparrow$ & \makecell{Text\\Edit$\downarrow$} & \makecell{Formula\\CDM$\uparrow$} & \makecell{Table\\TEDS$\uparrow$} & \makecell{Table\\TEDS-S$\uparrow$} & \makecell{RO\\Edit$\downarrow$} \\
\midrule
\multirow{8}{*}{\makecell[l]{General\\VLMs}} & InternVL3.5-241B~\cite{model_intern35} & 241B/28B & 83.76 & 0.1300 & 89.95 & 74.35 & 79.78 & 0.2150 \\
 & Kimi K2.5~\cite{model_kimi25} & 1T/32B & 84.53 & 0.1070 & 83.50 & 80.76 & 84.00 & 0.2110 \\
 & GPT-5.2~\cite{model_gpt52} & -- & 86.59 & 0.1140 & 88.21 & 82.95 & 87.93 & 0.1930 \\
 & Qwen3-VL-235B~\cite{model_qwen3vl} & 235B/22B & 89.78 & 0.0630 & 92.55 & 83.07 & 86.75 & 0.1660 \\
 & Gemini 3 Flash~\cite{model_gem3flash} & -- & 92.62 & 0.0660 & 95.16 & 89.29 & 93.51 & 0.1720 \\
 & Gemini 3 Pro~\cite{model_gem3pro} & -- & 92.91 & 0.0640 & 95.99 & 89.15 & 92.96 & 0.1650 \\
 & Qwen3.8-Max$^{\dagger}$~\cite{model_qwen38} & -- & 93.70 & 0.0431 & 95.41 & 89.99 & 92.87 & 0.1379 \\
 & Ovis2.6-30B-A3B~\cite{model_ovis26} & 30B/3B & 93.70 & 0.0350 & 95.17 & 89.44 & 92.40 & 0.1350 \\
\midrule
\multirow{18}{*}{\makecell[l]{Specialized\\OCR Models}} & Dolphin-v2~\cite{model_dolphin2} & 3B & 89.50 & 0.0690 & 91.01 & 84.40 & 87.44 & 0.1500 \\
 & DeepSeek-OCR 2~\cite{model_deep2} & 3B/0.5B & 90.25 & 0.0500 & 91.84 & 83.89 & 87.75 & 0.1440 \\
 & dots.ocr~\cite{model_dotsocr} & 3B & 90.77 & 0.0480 & 89.95 & 87.18 & 90.58 & 0.1380 \\
 & Qwen3.5-OCR$^{\dagger}$~\cite{model_qwenocr} & -- & 91.27 & 0.0636 & 92.47 & 87.70 & 91.04 & 0.2234 \\
 & MonkeyOCRv2-B-Parsing~\cite{docparse_monkey_report} & 0.7B & 91.57 & 0.0530 & 91.83 & 88.24 & 91.38 & 0.1310 \\
 & MinerU2.5~\cite{model_miner25} & 1.2B & 93.04 & 0.0450 & 95.77 & 87.88 & 91.47 & 0.1300 \\
 & Logics-Parsing-v2~\cite{model_logicsv2} & 4B & 93.33 & 0.0410 & 95.65 & 88.42 & 91.98 & 0.1370 \\
 & Youtu-Parsing~\cite{model_youtu} & 2.5B & 93.74 & 0.0440 & 93.63 & 92.02 & 95.00 & 0.1160 \\
 & Qianfan-OCR~\cite{model_qianfan} & 4B & 93.90 & 0.0400 & 95.08 & 90.53 & 93.31 & 0.1300 \\
 & PaddleOCR-VL~\cite{model_paddle} & 0.9B & 94.18 & 0.0400 & 95.91 & 90.65 & 93.74 & 0.1350 \\
 & HunyuanOCR-1.5~\cite{docparse_hunyuan_report} & 1B & 94.74 & 0.0390 & 94.50 & 93.67 & 94.71 & 0.1290 \\
 & PaddleOCR-VL-1.5~\cite{model_paddle15} & 0.9B & 94.93 & 0.0380 & 96.89 & 91.67 & 94.37 & 0.1300 \\
 & GLM-OCR~\cite{model_glmocr} & 0.9B & 95.22 & 0.0440 & 97.18 & 92.83 & 95.39 & 0.1330 \\
 & MinerU2.5-Pro~\cite{model_minerpro} & 1.2B & 95.75 & 0.0360 & 97.45 & 93.42 & 95.92 & 0.1200 \\
 & PaddleOCR-VL-1.6~\cite{model_paddle16} & 0.9B & 96.34 & 0.0326 & 97.53 & 94.76 & 97.10 & 0.1278 \\
 & OvisOCR2~\cite{docparse_ovis2_report} & 0.8B & 96.58 & \bestcell{0.0250} & 97.53 & 94.76 & 97.16 & \bestcell{0.1110} \\
 & Xiaomi-OCR-0~\cite{docparse_xiaomi_report} & 0.8B & \bestcell{96.83} & 0.0310 & \bestcell{98.50} & \secondcell{95.11} & \secondcell{97.19} & 0.1220 \\
\cmidrule(l){2-9}
 & \textbf{\ours{}} & 0.8B & \secondcell{96.74} & \secondcell{0.0261} & \secondcell{97.66} & \bestcell{95.16} & \bestcell{97.46} & \secondcell{0.1137} \\
\bottomrule
\end{tabular*}
\end{table}
\Needspace{3\baselineskip}

As shown in Table~\ref{tab:docparse_odb}, \ours{} achieves an Overall score of 96.74, exceeding OvisOCR2 and PaddleOCR-VL-1.6 and closely matching Xiaomi-OCR-0. It ranks among the top two on all five individual metrics, leading in both TEDS and TEDS-S, a result that highlights its strength in table parsing.

\subsubsection{Robustness to Image Degradation}

We evaluate robustness to image degradation on PureDocBench and Real5-OmniDocBench. PureDocBench~\cite{li2026puredocbench,docparse_pdb_leaderboard} comprises 1,475 source pages spanning 10 domains and 66 subcategories. Each page has three matched views (Clean, Digital, and Real), yielding 4,425 images in total. The Clean, Digital, and Real tracks contain rendered, digitally degraded, and recaptured pages, respectively. Each track aggregates text accuracy, formula CDM, and table TEDS as described above; Overall averages the three track scores.

\begin{table}[!htbp]
\centering
\fontsize{9}{11.2}\selectfont
\setlength{\tabcolsep}{2pt}
\renewcommand{\arraystretch}{1.08}
\caption{Document parsing results on PureDocBench. Overall averages the Clean, Digital, and Real track scores. Baseline scores without $\dagger$ follow the benchmark leaderboard~\cite{docparse_pdb_leaderboard}.}
\label{tab:docparse_pdb}
\begin{tabular*}{\linewidth}{@{\extracolsep{\fill}}llcrrrr@{}}
\toprule
\textbf{Model Type} & \textbf{Model} & \textbf{Parameters} & \textbf{Overall}$\uparrow$ & Clean$\uparrow$ & Digital$\uparrow$ & Real$\uparrow$ \\
\midrule
\multirow{8}{*}{\makecell[l]{General\\VLMs}} & GPT-5.6 Sol~\cite{model_gpt56} & -- & 69.16 & 74.15 & 71.73 & 61.59 \\
 & Kimi K2.6~\cite{model_kimi26} & 1T/32B & 70.10 & 72.32 & 69.95 & 68.02 \\
 & Qwen3.5-122B-A10B~\cite{model_qwen35} & 122B/10B & 74.11 & 76.14 & 76.34 & 69.85 \\
 & Claude Opus 5~\cite{model_opus5} & -- & 77.07 & 82.99 & 72.94 & 75.27 \\
 & Qwen3.8-27B~\cite{model_qwen3827b} & 27B & 77.63 & 80.45 & 78.86 & 73.59 \\
 & Gemini 3.6 Flash~\cite{model_gem36} & -- & 78.27 & 79.97 & 79.11 & 75.73 \\
 & GLM-5.3-Flash~\cite{model_glm53} & 320B/18B & 79.69 & 83.04 & 81.03 & 74.99 \\
 & Qwen3.8-Max$^{\dagger}$~\cite{model_qwen38} & -- & \secondcell{80.56} & \secondcell{83.31} & \secondcell{81.79} & \secondcell{76.58} \\
\midrule
\multirow{13}{*}{\makecell[l]{Specialized\\OCR Models}} & DeepSeek-OCR 2~\cite{model_deep2} & 3B/0.5B & 49.51 & 55.53 & 49.41 & 43.60 \\
 & GLM-OCR~\cite{model_glmocr} & 0.9B & 63.34 & 68.65 & 63.06 & 58.31 \\
 & MonkeyOCRv2-B-Parsing~\cite{docparse_monkey_report} & 0.7B & 64.91 & 69.82 & 66.08 & 58.83 \\
 & PaddleOCR-VL-1.6$^{\dagger}$~\cite{model_paddle16} & 0.9B & 65.44 & 70.52 & 64.97 & 60.83 \\
 & HunyuanOCR-1.5~\cite{docparse_hunyuan_report} & 1B & 67.58 & 73.09 & 69.36 & 60.28 \\
 & Qwen3.5-OCR$^{\dagger}$~\cite{model_qwenocr} & -- & 68.38 & 74.87 & 68.33 & 61.93 \\
 & MinerU2.5-Pro~\cite{model_minerpro} & 1.2B & 70.07 & 75.87 & 71.77 & 62.56 \\
 & dots.mocr~\cite{model_dotsmocr} & 3B & 70.39 & 76.27 & 73.16 & 61.73 \\
 & FD-RL~\cite{model_fdrl} & 4B & 73.92 & 78.38 & 76.33 & 67.04 \\
 & Xiaomi-OCR-0$^{\dagger}$~\cite{docparse_xiaomi_report} & 0.8B & 74.60 & 79.48 & 75.81 & 68.51 \\
 & OvisOCR2~\cite{docparse_ovis2_report} & 0.8B & 75.10 & 81.53 & 77.29 & 66.49 \\
 & WeVisDoc-4B~\cite{model_wevis} & 4B & 75.57 & 79.59 & 77.99 & 69.14 \\
\cmidrule(l){2-7}
 & \textbf{\ours{}} & 0.8B & \bestcell{82.31} & \bestcell{85.89} & \bestcell{82.98} & \bestcell{78.05} \\
\bottomrule
\end{tabular*}
\end{table}
\Needspace{3\baselineskip}

As shown in Table~\ref{tab:docparse_pdb}, \ours{} achieves the highest Overall score of 82.31 and leads on all three tracks, outperforming both Qwen3.8-Max and OvisOCR2. The Clean-to-Real performance drop is 7.84 points for \ours{}, compared with 15.04 points for OvisOCR2, indicating improved robustness to physical recapture.

\Needspace{3\baselineskip}
Real5-OmniDocBench~\cite{zhou2026real5omnidocbench,docparse_real5_leaderboard} extends 1,355 pages from OmniDocBench v1.5 to five acquisition conditions: Scanning, Warping, Screen-Photography, Illumination, and Skew, yielding 6,775 images. We evaluate against the original Real5 annotations using the OmniDocBench v1.5 scoring protocol. Each condition aggregates text accuracy, formula CDM, and table TEDS; Overall averages the five condition scores.

\begin{table}[!htbp]
\centering
\fontsize{9}{11.2}\selectfont
\setlength{\tabcolsep}{2pt}
\renewcommand{\arraystretch}{1.08}
\caption{Document parsing results on Real5-OmniDocBench. Overall averages the scores across the five acquisition conditions; Illum. denotes illumination. Baseline scores without $\dagger$ follow the official leaderboard~\cite{docparse_real5_leaderboard}.}
\label{tab:docparse_real5}
\begin{tabular*}{\linewidth}{@{\extracolsep{\fill}}llcrrrrrr@{}}
\toprule
\textbf{Model Type} & \textbf{Model} & \textbf{Parameters} & \textbf{Overall}$\uparrow$ & Scanning$\uparrow$ & Warping$\uparrow$ & \makecell{Screen\\Photo.$\uparrow$} & Illum.$\uparrow$ & Skew$\uparrow$ \\
\midrule
\multirow{8}{*}{\makecell[l]{General\\VLMs}} & GPT-5.2~\cite{model_gpt52} & -- & 78.66 & 84.43 & 76.26 & 76.75 & 80.88 & 75.00 \\
 & Qwen3.8-Max$^{\dagger}$~\cite{model_qwen38} & -- & 86.86 & 91.80 & 90.58 & 80.59 & 86.18 & 85.17 \\
 & Qwen2.5-VL-72B~\cite{model_qwen25vl} & 72B & 86.92 & 86.19 & 87.77 & 86.48 & 87.25 & 86.90 \\
 & Qwen3-VL-235B~\cite{model_qwen3vl} & 235B/22B & 88.90 & 89.43 & 89.99 & 89.27 & 89.27 & 86.56 \\
 & Doubao-Seed-2.1-Pro~\cite{model_seed21} & -- & 89.02 & 88.85 & 89.36 & 88.99 & 89.13 & 88.79 \\
 & Gemini 3 Pro~\cite{model_gem3pro} & -- & 89.24 & 89.47 & 88.90 & 88.86 & 89.53 & 89.45 \\
 & Kimi K2.6~\cite{model_kimi26} & 1T/32B & 89.76 & 90.08 & 89.62 & 89.58 & 89.91 & 89.61 \\
 & GLM-5.3-Flash~\cite{model_glm53} & 320B/18B & 90.76 & 91.43 & 89.84 & 90.36 & 91.40 & 90.76 \\
\midrule
\multirow{12}{*}{\makecell[l]{Specialized\\OCR Models}} & DeepSeek-OCR 2~\cite{model_deep2} & 3B/0.5B & 73.01 & 89.59 & 66.53 & 71.65 & 76.02 & 61.28 \\
 & Qwen3.5-OCR$^{\dagger}$~\cite{model_qwenocr} & -- & 82.48 & 91.06 & 83.31 & 81.98 & 89.18 & 66.86 \\
 & dots.ocr~\cite{model_dotsocr} & 3B & 86.38 & 86.87 & 86.01 & 87.18 & 87.57 & 84.27 \\
 & HunyuanOCR-1.5$^{\dagger}$~\cite{docparse_hunyuan_report} & 1B & 88.53 & 90.05 & 87.90 & 88.59 & 89.01 & 87.09 \\
 & MinerU2.5-Pro~\cite{model_minerpro} & 1.2B & 88.94 & 92.11 & 88.72 & 91.29 & 91.31 & 81.26 \\
 & MonkeyOCRv2-B-Parsing~\cite{docparse_monkey_report} & 0.7B & 89.22 & 89.49 & 89.70 & 88.40 & 88.54 & 89.97 \\
 & GLM-OCR~\cite{model_glmocr} & 0.9B & 90.32 & 92.67 & 90.68 & 91.75 & 91.12 & 85.39 \\
 & PaddleOCR-VL-1.5~\cite{model_paddle15} & 0.9B & 92.05 & 93.43 & 91.25 & 91.76 & 92.16 & 91.66 \\
 & OvisOCR2~\cite{docparse_ovis2_report} & 0.8B & 92.29 & \secondcell{93.77} & 91.40 & \secondcell{93.09} & 92.88 & 90.33 \\
 & Xiaomi-OCR-0~\cite{docparse_xiaomi_report} & 0.8B & 92.73 & 93.73 & \secondcell{92.76} & 92.80 & 92.68 & 91.69 \\
 & PaddleOCR-VL-1.6~\cite{model_paddle16} & 0.9B & \secondcell{93.19} & \bestcell{94.74} & 92.48 & 92.78 & \secondcell{93.28} & \secondcell{92.66} \\
\cmidrule(l){2-9}
 & \textbf{\ours{}} & 0.8B & \bestcell{93.63} & 93.76 & \bestcell{93.20} & \bestcell{93.62} & \bestcell{94.10} & \bestcell{93.47} \\
\bottomrule
\end{tabular*}
\end{table}
\Needspace{3\baselineskip}

As shown in Table~\ref{tab:docparse_real5}, \ours{} achieves the highest Overall score of 93.63, exceeding PaddleOCR-VL-1.6 and OvisOCR2. It leads in four of the five conditions, with a maximum score difference of 0.90 points across all five, indicating consistent parsing performance under varied acquisition conditions.

\subsubsection{Multilingual Document Parsing}

MDPBench~\cite{li2026mdpbench,docparse_mdp_leaderboard} evaluates multilingual document parsing on digital and photographed documents. Its public split contains 2,720 images spanning 17 language categories. We report Overall and breakdowns by acquisition type (Digital and Photo) and script group (Latin and Non-Latin), following the benchmark's official scoring and aggregation procedures.

\begin{table}[!htbp]
\centering
\fontsize{9}{11.2}\selectfont
\setlength{\tabcolsep}{2pt}
\renewcommand{\arraystretch}{1.08}
\caption{Document parsing results on the MDPBench public split, with breakdowns by acquisition type and script group. Baseline scores without $\dagger$ follow the official leaderboard~\cite{docparse_mdp_leaderboard}.}
\label{tab:docparse_mdp}
\begin{tabular*}{\linewidth}{@{\extracolsep{\fill}}llcrrrrr@{}}
\toprule
\textbf{Model Type} & \textbf{Model} & \textbf{Parameters} & \textbf{Overall}$\uparrow$ & Digital$\uparrow$ & Photo$\uparrow$ & Latin$\uparrow$ & Non-Latin$\uparrow$ \\
\midrule
\multirow{7}{*}{\makecell[l]{General\\VLMs}} & Qwen3-VL-8B~\cite{model_qwen3vl} & 8B & 68.3 & 78.4 & 65.0 & 73.6 & 62.5 \\
 & GPT-5.2~\cite{model_gpt52} & -- & 68.6 & 85.6 & 63.0 & 75.2 & 61.1 \\
 & Claude Sonnet 4.6~\cite{model_sonnet46} & -- & 73.1 & 85.0 & 69.3 & 79.2 & 66.2 \\
 & Doubao-Seed-2.0-Pro~\cite{model_seed20} & -- & 74.2 & 78.9 & 72.8 & 75.7 & 72.5 \\
 & Kimi K3~\cite{model_kimi3} & 2.8T/104B & 83.6 & 90.8 & 81.2 & 86.2 & 80.7 \\
 & Gemini 3 Pro~\cite{model_gem3pro} & -- & \secondcell{86.4} & 90.4 & \secondcell{85.1} & \secondcell{88.4} & \secondcell{84.1} \\
 & Qwen3.8-Max$^{\dagger}$~\cite{model_qwen38} & -- & \secondcell{86.4} & \secondcell{91.7} & 84.6 & \secondcell{88.4} & \secondcell{84.1} \\
\midrule
\multirow{14}{*}{\makecell[l]{Specialized\\OCR Models}} & Qwen3.5-OCR$^{\dagger}$~\cite{model_qwenocr} & -- & 60.6 & 81.3 & 53.8 & 67.6 & 52.8 \\
 & GLM-OCR~\cite{model_glmocr} & 0.9B & 67.3 & 77.9 & 63.7 & 78.7 & 54.3 \\
 & OvisOCR2$^{\dagger}$~\cite{docparse_ovis2_report} & 0.8B & 69.0 & 83.7 & 64.1 & 80.1 & 56.4 \\
 & olmOCR2~\cite{model_olm2} & 7B & 70.4 & 79.9 & 67.2 & 76.7 & 63.3 \\
 & MinerU2.5-Pro~\cite{model_minerpro} & 1.2B & 71.0 & 86.2 & 66.1 & 74.6 & 67.0 \\
 & Xiaomi-OCR-0$^{\dagger}$~\cite{docparse_xiaomi_report} & 0.8B & 73.0 & 86.5 & 68.6 & 79.8 & 65.4 \\
 & HunyuanOCR-1.5~\cite{docparse_hunyuan_report} & 1B & 76.8 & 86.2 & 73.6 & 79.7 & 73.5 \\
 & Mistral-OCR-4.0~\cite{model_mistral4} & -- & 78.2 & 87.5 & 75.2 & 80.0 & 76.1 \\
 & PaddleOCR-VL-1.6~\cite{model_paddle16} & 0.9B & 78.9 & 87.1 & 76.3 & 81.4 & 76.1 \\
 & Chandra-OCR-2~\cite{model_chandra} & 5.3B & 79.7 & 87.8 & 77.1 & 82.7 & 76.4 \\
 & dots.mocr~\cite{model_dotsmocr} & 3B & 80.5 & 90.5 & 77.2 & 81.7 & 79.2 \\
 & PepperOCR-VL~\cite{model_pepper} & 4.5B & 82.3 & 87.3 & 80.7 & 85.3 & 78.9 \\
 & MonkeyOCRv2-B-Parsing~\cite{docparse_monkey_report} & 0.7B & 83.3 & 88.1 & 81.7 & 84.2 & 82.1 \\
\cmidrule(l){2-8}
 & \textbf{\ours{}} & 0.8B & \bestcell{88.8} & \bestcell{93.0} & \bestcell{87.4} & \bestcell{89.5} & \bestcell{88.0} \\
\bottomrule
\end{tabular*}
\end{table}
\Needspace{3\baselineskip}

As shown in Table~\ref{tab:docparse_mdp}, \ours{} achieves the highest Overall score of 88.8, exceeding both Gemini 3 Pro and Qwen3.8-Max by 2.4 points. It also leads across both acquisition types and both script groups. Compared with OvisOCR2, the larger gains on non-Latin documents narrow the Latin/non-Latin performance gap from 23.7 to 1.5 points, indicating more balanced performance across the two script groups.

\subsection{General OCR}
\label{sec:evaluation_general_ocr}
We evaluate \ours{} on OCRBench~\citep{liu2023ocrbench} and OCRBench v2~\citep{fu2025ocrbenchv2}, which together assess text recognition, localization, information extraction, and visual reasoning. OCRBench contains 1,000 samples across five task categories. Its Overall score is the percentage of accepted answers. OCRBench v2 contains 7,400 English samples across eight task categories and 2,600 Chinese samples across five, totaling 10,000 samples. We follow the benchmark's task-specific scoring rules and aggregate subtask results within each category. For each language, the Overall score is the macro-average of the category scores. All scores are reported on a 0--100 scale.

For OCRBench, we use the Qwen3.5 and Xiaomi-OCR-0 scores reported in their official model cards~\citep{qwen2026qwen3508b,qwen2026qwen352b,qwen2026qwen354b,model_xiaomiocr0_card}, with non-thinking results for Qwen3.5-0.8B and Qwen3.5-2B. The HunyuanOCR-1.5 score follows Table~15 of its technical report~\citep{docparse_hunyuan_report}, rescaled from 861/1,000 to 86.1.

\begin{table}[!htbp]
\centering
\small
\setlength{\tabcolsep}{4pt}
\renewcommand{\arraystretch}{1.12}
\caption{Performance comparison on OCRBench and OCRBench v2.}
\label{tab:general_ocr_overall}
\begin{tabularx}{\linewidth}{@{}Ycrrr@{}}
\toprule
\textbf{Model} & \textbf{Parameters} & \makecell{\textbf{OCRBench}} & \makecell{\textbf{v2 English}} & \makecell{\textbf{v2 Chinese}} \\
\midrule
Qwen3.5-0.8B~\citep{qwen2026qwen3508b} & 0.8B & 79.1 & 41.5 & 43.9 \\
Qwen3.5-2B~\citep{qwen2026qwen352b} & 2B & 85.4 & 48.1 & 55.4 \\
Qwen3.5-4B~\citep{qwen2026qwen354b} & 4B & 85.0 & \underline{54.1} & \underline{61.0} \\
Qwen3.5-OCR~\citep{model_qwenocr} & -- & 76.8 & 44.5 & 43.8 \\
PaddleOCR-VL-1.6~\citep{model_paddle16} & 0.9B & 55.9 & 23.8 & 30.4 \\
HunyuanOCR-1.5~\citep{docparse_hunyuan_report} & 1B & \underline{86.1} & 50.7 & 57.7 \\
GLM-OCR~\citep{model_glmocr} & 0.9B & 83.8 & 32.9 & 48.1 \\
Xiaomi-OCR-0~\citep{model_xiaomiocr0_card} & 0.8B & 84.6 & 40.1 & 48.4 \\
\midrule
\textbf{\ours{}} & 0.8B & \textbf{88.8} & \textbf{65.9} & \textbf{61.3} \\
\bottomrule
\end{tabularx}
\end{table}
\Needspace{3\baselineskip}

As shown in Table~\ref{tab:general_ocr_overall}, \ours{} ranks first on OCRBench and the English and Chinese splits of OCRBench v2, with scores of 88.8, 65.9, and 61.3, respectively. These results demonstrate strong performance on both language splits. Notably, on the English split, \ours{} outperforms Qwen3.5-4B by 11.8 points while using one-fifth as many parameters.

\begin{table}[!htbp]
\centering
\small
\setlength{\tabcolsep}{4pt}
\renewcommand{\arraystretch}{1.12}
\caption{Task-level performance of OvisOCR3 on OCRBench.}
\label{tab:general_ocr_classic}
\begin{tabularx}{\linewidth}{@{}Yrrr@{}}
\toprule
\textbf{Task category} & \textbf{Samples} & \textbf{Accepted} & \textbf{Score (\%)} \\
\midrule
Text Recognition & 300 & 284 & 94.7 \\
Scene Text-centric VQA & 200 & 167 & 83.5 \\
Doc-oriented VQA & 200 & 180 & 90.0 \\
Key Information Extraction & 200 & 181 & 90.5 \\
Handwritten Mathematical Expression Recognition & 100 & 76 & 76.0 \\
\midrule
Total & 1,000 & 888 & 88.8 \\
\bottomrule
\end{tabularx}
\end{table}
\begin{table}[!htbp]
\centering
\small
\setlength{\tabcolsep}{4pt}
\renewcommand{\arraystretch}{1.12}
\caption{Task-level performance of OvisOCR3 on OCRBench v2.}
\label{tab:general_ocr_v2_tasks}
\begin{tabularx}{\linewidth}{@{}Ylrr@{}}
\toprule
\textbf{Task category} & \textbf{Language} & \textbf{Samples} & \textbf{Score} \\
\midrule
Text recognition & EN & 1,200 & 76.9 \\
Text detection & EN & 500 & 66.5 \\
Text spotting & EN & 200 & 69.9 \\
Relationship extraction & EN & 700 & 87.7 \\
Element parsing & EN & 1,600 & 59.2 \\
Mathematical calculation & EN & 500 & 34.6 \\
Visual text understanding & EN & 1,300 & 73.2 \\
Knowledge reasoning & EN & 1,400 & 59.2 \\
Text recognition & CN & 200 & 80.4 \\
Relationship extraction & CN & 600 & 75.9 \\
Element parsing & CN & 800 & 47.9 \\
Visual text understanding & CN & 200 & 62.0 \\
Knowledge reasoning & CN & 800 & 40.4 \\
\bottomrule
\end{tabularx}
\end{table}
\Needspace{3\baselineskip}

Tables~\ref{tab:general_ocr_classic} and~\ref{tab:general_ocr_v2_tasks} provide task-level breakdowns for OCRBench and OCRBench v2, respectively. Strong performance across a range of OCR tasks, including text recognition, information extraction, question answering, and text spotting, demonstrates the versatility of \ours{} as a general OCR model.

\subsection{Key Information Extraction}
\label{sec:evaluation_kie}
We evaluate KIE on three public benchmarks and our in-house KIE Bench. We use all KIE samples from the three public benchmarks: 200 from OCRBench~\citep{liu2023ocrbench}, 400 English and 400 Chinese samples from OCRBench v2~\citep{fu2025ocrbenchv2}, and 2,008 from CC-OCR~\citep{yang2024ccocr}. The in-house KIE Bench contains 644 images across various business-document categories.

When evaluating \ours{}, we use specified-field prompts for OCRBench, OCRBench v2, and the in-house KIE Bench, and schema-based prompts for CC-OCR. We adapt the benchmark interfaces and scoring protocols to align field names, output formats, and missing-value representations across models while retaining each benchmark's extraction task and metric. Text normalization and empty-value handling follow benchmark-specific rules. Empty or truncated responses are included in the evaluation.

For OCRBench, we report the percentage of questions answered correctly under the benchmark's answer-matching rule. OCRBench v2 measures agreement between predicted and reference key-value pairs after text normalization; we average the English and Chinese scores with equal weight. For CC-OCR, we use normalized tree-edit-distance accuracy~\citep{kim2022donut} to assess both JSON structure and field values, and report the mean across its six KIE subsets.

For the in-house KIE Bench, we measure character-level similarity between each extracted field value and its reference as $1-d(p,g)/\max(|p|,|g|)$, where $p$ and $g$ are the normalized prediction and reference strings, $d$ is the Levenshtein edit distance~\citep{levenshtein1965binary}, and $|\cdot|$ denotes string length. A pair of empty strings receives a score of one. We average field scores within each document category, then take the unweighted mean across categories. All scores are reported on a 0--100 scale.

\begin{table}[!htbp]
\centering
\small
\setlength{\tabcolsep}{5pt}
\renewcommand{\arraystretch}{1.15}
\caption{Performance comparison on KIE benchmarks.}
\label{tab:kie_results}
\begin{tabularx}{\linewidth}{@{}Xrrrr@{}}
\toprule
\textbf{Model} & \makecell{\textbf{OCRBench}\\\textbf{KIE}} & \makecell{\textbf{OCRBench v2}\\\textbf{KIE}} & \makecell{\textbf{CC-OCR}\\\textbf{KIE}} & \makecell{\textbf{In-house}\\\textbf{KIE Bench}} \\
\midrule
Qwen3.5-OCR~\citep{model_qwenocr} & 78.50 & 72.90 & \secondcell{88.43} & 59.90 \\
GLM-OCR~\citep{model_glmocr} & 79.00 & 74.22 & 68.36 & 75.25 \\
HunyuanOCR-1.5~\citep{docparse_hunyuan_report} & \secondcell{81.00} & 68.63 & 66.08 & \secondcell{75.96} \\
Xiaomi-OCR-0~\citep{docparse_xiaomi_report} & 76.50 & \secondcell{74.66} & 79.53 & 75.68 \\
\midrule
\ours{} & \bestcell{90.50} & \bestcell{83.01} & \bestcell{88.69} & \bestcell{77.73} \\
\bottomrule
\end{tabularx}
\end{table}

As shown in Table~\ref{tab:kie_results}, \ours{} ranks first on all four KIE benchmarks. The largest gains are on OCRBench and OCRBench v2, where \ours{} exceeds the strongest baselines by 9.50 and 8.35 points, respectively. \ours{} also leads in the schema-based evaluation on CC-OCR, achieving 88.69 and outperforming Qwen3.5-OCR by 0.26 points. Beyond the public benchmarks, \ours{} achieves 77.73 on the in-house KIE Bench, exceeding HunyuanOCR-1.5 by 1.77 points. Together, these results demonstrate \ours{}'s strong key information extraction capabilities across diverse document types and extraction settings.

\subsection{Document Figure Captioning}

To evaluate document figure captioning, we introduce DocFigCapBench, an in-house benchmark built from PureDocBench, MDPBench, and OmniDocBench. We randomly sampled candidate pages from these sources, then annotated and reviewed them. The final benchmark contains 182 pages with 372 figure targets and 161 ignored visual regions.

\begin{figure}[!htbp]
\centering
\begin{subfigure}[t]{0.515\linewidth}
\vspace{0pt}
\centering
\begin{tikzpicture}
\node[anchor=south west,inner sep=0] (page) at (0,0)
  {\includegraphics[width=\linewidth]{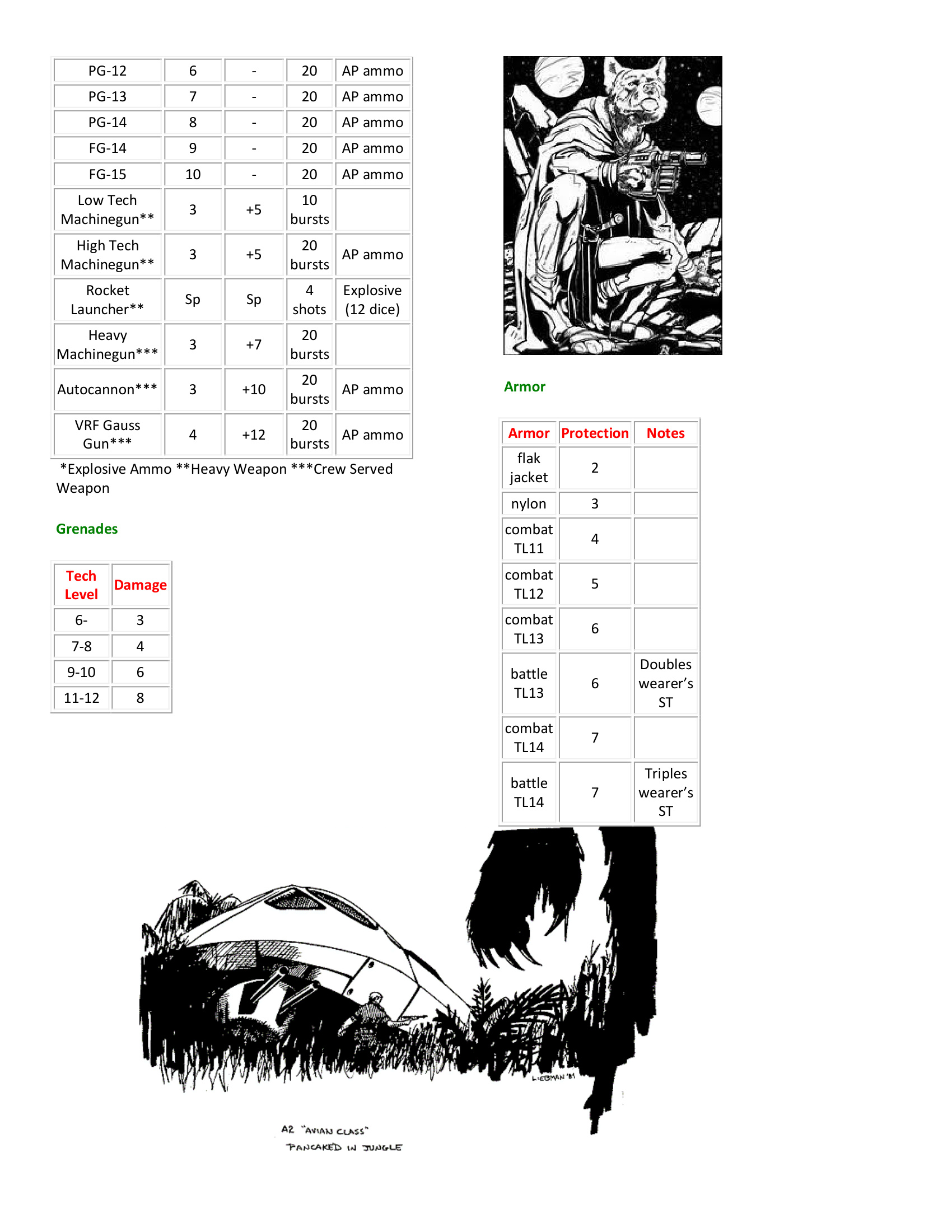}};
\begin{scope}[x={(page.south east)},y={(page.north west)}]
  \draw[gray!75,line width=0.35pt] (0.0522,0.6275) rectangle (0.4358,0.9529);
  \draw[gray!75,line width=0.35pt] (0.0574,0.5963) rectangle (0.4135,0.6259);
  \draw[gray!75,line width=0.35pt] (0.0571,0.5652) rectangle (0.1237,0.5764);
  \draw[gray!75,line width=0.35pt] (0.0534,0.4205) rectangle (0.1818,0.5452);
  \draw[gray!75,line width=0.35pt] (0.5271,0.6795) rectangle (0.5743,0.6916);
  \draw[gray!75,line width=0.35pt] (0.5232,0.3287) rectangle (0.7379,0.6611);
  \node[anchor=north west,fill=white,text=ovisgray,
        font=\tiny\bfseries,inner sep=0.4pt] at (0.0522,0.9529) {1};
  \node[anchor=north west,fill=white,text=ovisgray,
        font=\tiny\bfseries,inner sep=0.4pt] at (0.0574,0.6259) {2};
  \node[anchor=north west,fill=white,text=ovisgray,
        font=\tiny\bfseries,inner sep=0.4pt] at (0.0571,0.5764) {3};
  \node[anchor=north west,fill=white,text=ovisgray,
        font=\tiny\bfseries,inner sep=0.4pt] at (0.0534,0.5452) {4};
  \node[anchor=north west,fill=white,text=ovisgray,
        font=\tiny\bfseries,inner sep=0.4pt] at (0.5271,0.6916) {6};
  \node[anchor=north west,fill=white,text=ovisgray,
        font=\tiny\bfseries,inner sep=0.4pt] at (0.5232,0.6611) {7};
  \draw[red!85!black,line width=0.9pt] (0.5280,0.7110) rectangle (0.7590,0.9550);
  \draw[red!85!black,line width=0.9pt] (0.1430,0.0630) rectangle (0.6980,0.3280);
  \node[anchor=north west,fill=red!85!black,text=white,
        font=\tiny\bfseries,inner sep=1.2pt] at (0.5280,0.9550) {IMAGE 001};
  \node[anchor=north west,fill=red!85!black,text=white,
        font=\tiny\bfseries,inner sep=1.2pt] at (0.1430,0.3280) {IMAGE 002};
\end{scope}
\end{tikzpicture}\par
\scriptsize\textcolor{red!85!black}{Red: figure targets}\quad
\textcolor{gray}{Gray: other page annotations}\par
\normalsize\caption{Document page.}
\end{subfigure}\hfill
\begin{subfigure}[t]{0.465\linewidth}
\vspace{0pt}
\raggedright\footnotesize
\begin{tcolorbox}[colback=red!2,colframe=red!65!black,boxrule=0.4pt,
  arc=0pt,left=3pt,right=3pt,top=2pt,bottom=2pt]
\textbf{I001}\quad \texttt{[528,45,759,289]}\par
\smallskip
This black and white illustration depicts an anthropomorphic canine warrior crouching in a starry space setting. The character has a dog-like head, wears a draped cloak, and holds a futuristic blaster pistol in its right hand. A sword hilt is visible at its waist. The background features a dark, star-filled sky with two planets or moons visible in the upper left and upper right corners. The character is positioned on a rocky, uneven surface.
\end{tcolorbox}
\vspace{2pt}
\begin{tcolorbox}[colback=red!2,colframe=red!65!black,boxrule=0.4pt,
  arc=0pt,left=3pt,right=3pt,top=2pt,bottom=2pt]
\textbf{I002}\quad \texttt{[143,672,698,937]}\par
\smallskip
This black-and-white illustration depicts a futuristic, dome-shaped spacecraft crashed in a dense jungle. A lone figure stands in the foreground, holding a weapon and looking toward the ship. To the right, a large, dark, claw-like silhouette looms over the scene. The text at the bottom reads "A2 'AVIAN CLASS' PANCAKED IN JUNGLE," and the artist's signature "LIEBMAN '81" is visible on the right.
\end{tcolorbox}
\vspace{2pt}
\scriptsize
\textbf{Other page annotations (six entries, summarized)}\par
\textcolor{ovisgray}{\(\triangleright\) [1] table: weapons (HTML)}\par
\textcolor{ovisgray}{\(\triangleright\) [2] table footnote: Explosive Ammo \ldots}\par
\textcolor{ovisgray}{\(\triangleright\) [3] table caption: Grenades}\par
\textcolor{ovisgray}{\(\triangleright\) [4] table: Tech Level / Damage (HTML)}\par
\textcolor{ovisgray}{\(\triangleright\) [6] table caption: Armor}\par
\textcolor{ovisgray}{\(\triangleright\) [7] table: Armor / Protection (HTML)}\par
\normalsize\centering\caption{Annotations.}
\end{subfigure}
\caption{An example document page from DocFigCapBench. (a) The full page, with two figure
targets in red and other page annotations in gray.
(b) Reference figure captions and a summary of the remaining annotations.}
\label{fig:caption_single_sample}
\end{figure}

\begin{figure}[!htbp]
\centering
{\setlength{\fboxsep}{0pt}\setlength{\fboxrule}{1pt}%
\fcolorbox{red}{white}{\includegraphics[width=0.45\linewidth,trim={101bp 242bp 87bp 275bp},clip]{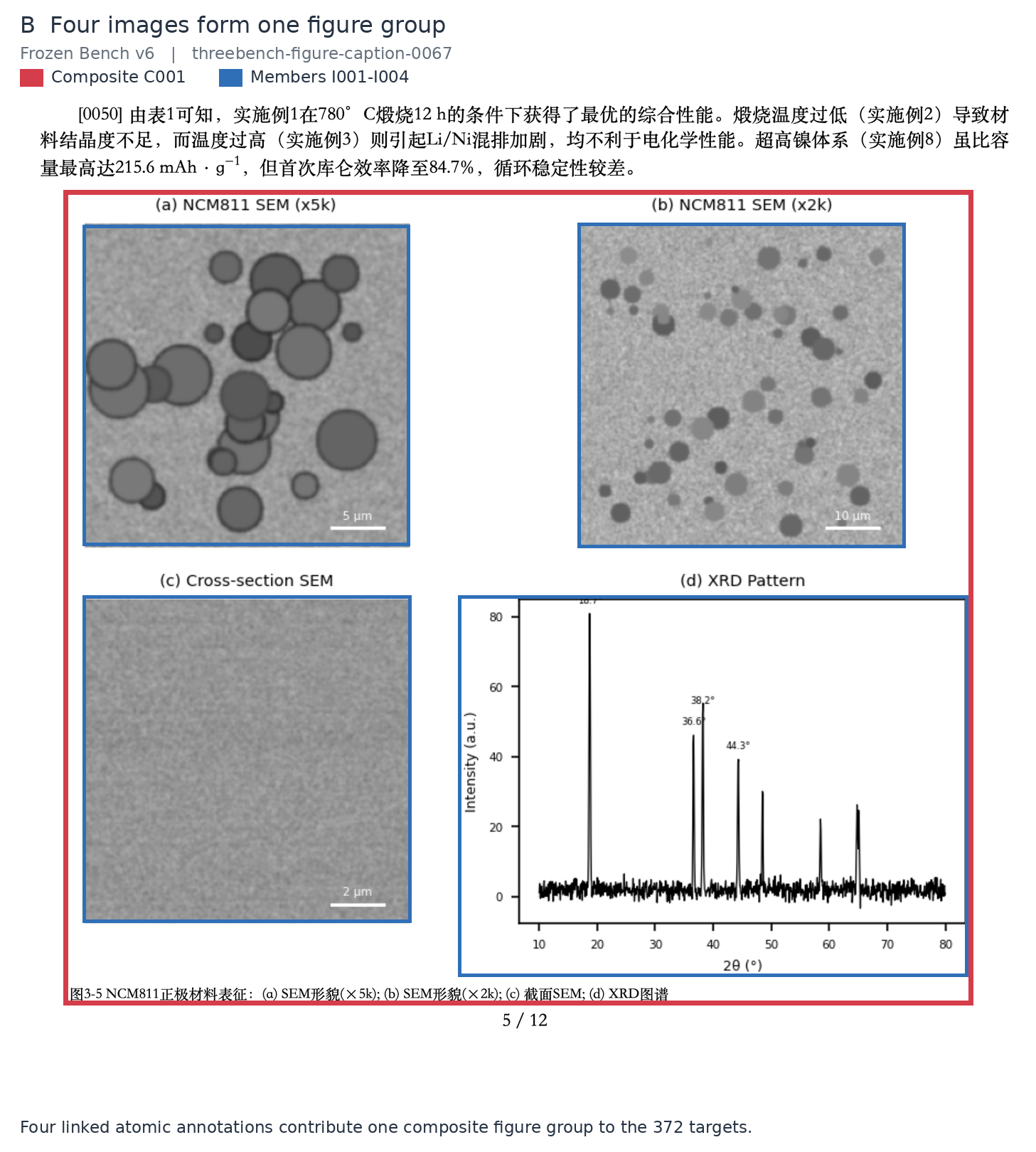}}}
\caption{An example of a four-image figure group from DocFigCapBench. The red outline marks one evaluation target; the blue outlines mark its four annotated members.}
\label{fig:caption_multi_group}
\end{figure}

Annotators marked information-bearing figures using the original pages and candidate locations from document parsers, while designating decorative graphics and header or footer visuals as ignored regions. Reference captions were drafted by a VLM and checked against the images for factual errors, omissions, and hallucinations (Figure~\ref{fig:caption_single_sample}). For multi-image figures, both the enclosing group and its individual members were annotated (Figure~\ref{fig:caption_multi_group}), allowing evaluation to accept either group-level or member-level detections.

Using these annotations, we evaluate figure detection with Localization F1 and end-to-end figure captioning with Caption Score. For localization, predictions are matched to figure targets at an IoU threshold of 0.5. A group-level prediction for a multi-image figure must additionally cover at least 80\% of each member, while member-level matches receive credit proportional to the fraction of members detected. Unmatched predictions are excluded from scoring if their IoU with an ignored region is at least 0.3 or at least 80\% of their area falls within it. We compute F1 as the harmonic mean of precision and recall on each page, then average it across all 182 pages and report the result as a percentage. Pages without figure targets receive full credit if they have no counted predictions, and zero otherwise.

For Caption Score, Qwen3.8-Max~\citep{model_qwen38} scores captions for matched figures on a 0--10 scale, using the image as the basis for judgment and the reference caption as a checklist. Missed figures and empty captions receive zero. For groups matched through their members, we average all member scores, including zeros for missed members. Figure scores are then averaged within each page and across the 177 pages containing figure targets.

\ours{} and the general VLM baselines use the same full-page prompt, which requests English captions of no more than 100 words, preserves visible labels in their original language when needed, and permits Mermaid representations for flowcharts. Invalid outputs are treated as empty predictions and remain in the evaluation denominators, including eight format-failed pages from Qwen3.7-Plus. Specialized OCR baselines use their native document-parsing workflows.

\begin{table}[!htbp]
\centering
\small
\setlength{\tabcolsep}{5pt}
\renewcommand{\arraystretch}{1.13}
\caption{Performance comparison on DocFigCapBench.}
\label{tab:caption_results}
\begin{tabular}{@{}llcc@{}}
\toprule
\textbf{Category} & \textbf{Model} & \textbf{Localization F1 \up} & \textbf{Caption Score \up} \\
\midrule
\multirow{5}{*}{General VLMs} & Qwen3.5-0.8B~\citep{qwen2026qwen3508b} & 5.47 & 0.20 \\
 & Qwen3.5-2B~\citep{qwen2026qwen352b} & 10.09 & 0.42 \\
 & Qwen3.5-4B~\citep{qwen2026qwen354b} & 40.19 & 2.27 \\
 & Qwen3.7-Plus~\citep{model_qwen37plus} & 71.14 & 5.69 \\
 & Qwen3.8-Max~\citep{model_qwen38} & \secondcell{86.67} & \bestcell{7.30} \\
\midrule
\multirow{6}{*}{\makecell[l]{Specialized\\OCR Models}} & HunyuanOCR-1.5~\citep{docparse_hunyuan_report} & 48.24 & -- \\
 & Qwen3.5-OCR~\citep{model_qwenocr} & 78.47 & 4.74 \\
 & OvisOCR2~\citep{docparse_ovis2_report} & 82.18 & -- \\
 & MinerU2.5-Pro~\citep{model_minerpro} & 85.00 & 4.61 \\
 & PaddleOCR-VL-1.6~\citep{model_paddle16} & 85.32 & 4.29 \\
 & \textbf{OvisOCR3} & \bestcell{90.97} & \secondcell{6.93} \\
\bottomrule
\end{tabular}
\end{table}

As reported in Table~\ref{tab:caption_results}, \ours{} achieves the highest Localization F1 on DocFigCapBench, exceeding Qwen3.8-Max by 4.30 points, with a Caption Score close to that of Qwen3.8-Max (6.93 vs.\ 7.30). Among specialized OCR models, \ours{} leads on both metrics, exceeding the strongest baseline by 5.65 points in Localization F1 and 2.19 points in Caption Score. These results show that \ours{} can accurately locate and describe document figures, capturing information that text extraction alone would miss.

\Needspace{8\baselineskip}
\subsection{Cross-Page Table Reconstruction}
\label{sec:eval_crosspage}

\begin{figure}[!htbp]
\centering
\includegraphics[width=\linewidth,height=0.86\textheight,keepaspectratio]{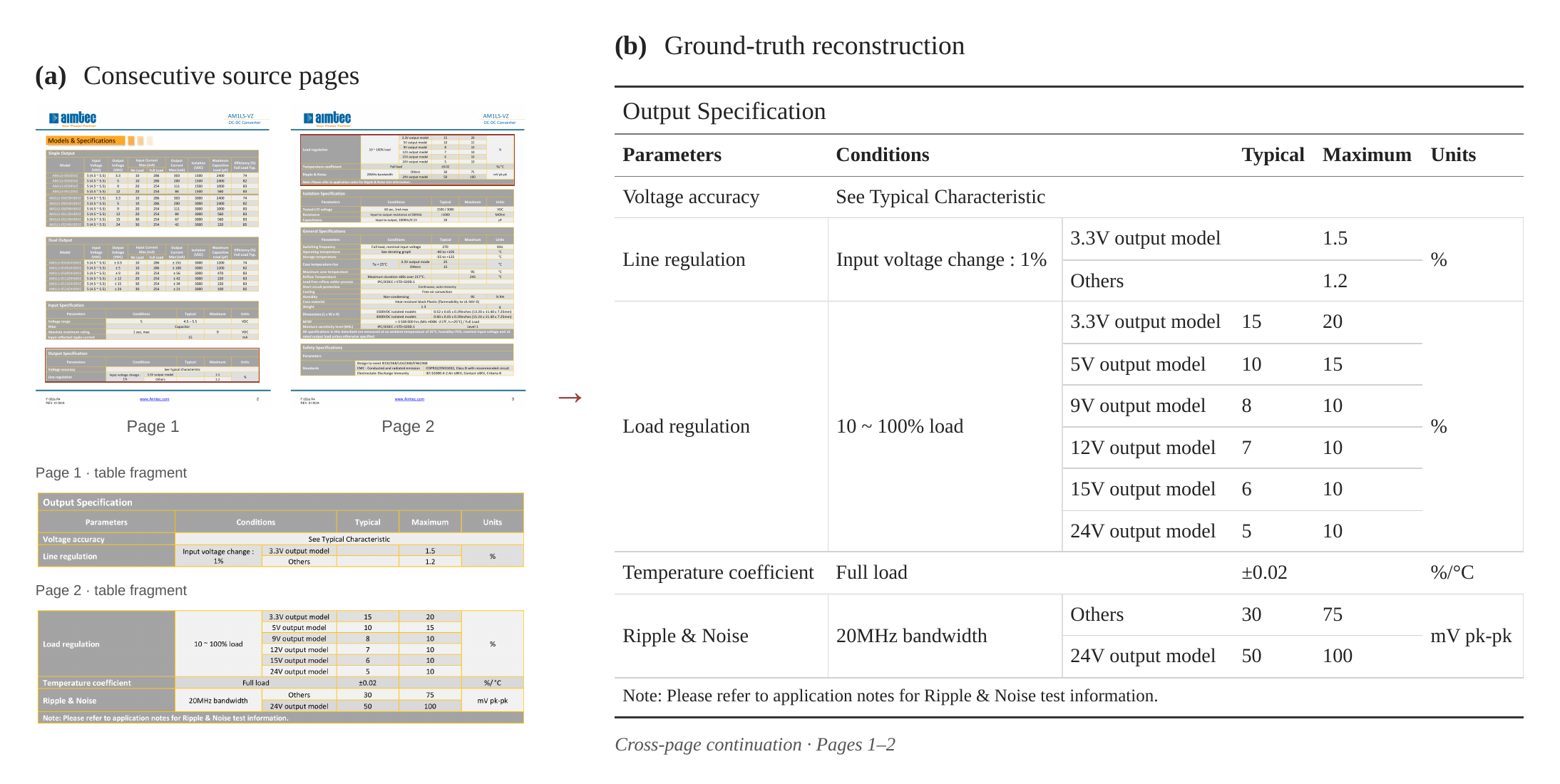}
\caption{An annotated two-page example from CrossTabBench. Red boxes mark fragments of the same table, with enlarged views below the source pages. The ground-truth reconstruction on the right preserves merged cells and numerical values.}
\label{fig:crosspage_benchmark}
\end{figure}

We evaluate cross-page table reconstruction on MPDocBench-Parse~\cite{zhou2026mpdocbench} and CrossTabBench. MPDocBench-Parse covers 15 document types in English and Chinese and includes a truncated-table reconstruction task. To further assess continuation detection and reconstruction across varied layouts, we introduce CrossTabBench, an in-house benchmark drawn from multilingual documents with both vertical continuations and horizontal stitching. It contains 890 consecutive-page pairs comprising 1,780 images: 491 pairs require table reconstruction, while the remaining 399 require no cross-page merge. Figure~\ref{fig:crosspage_benchmark} illustrates an annotated example.

\Needspace{3\baselineskip}
On CrossTabBench, we assess continuation detection with Relation F1 and reconstruction quality with Reconstruction TEDS. Relation F1 is the harmonic mean of precision and recall computed over all 890 page pairs, with false merges counted as false positives and missed continuations as false negatives. Missing or malformed outputs count as incorrect decisions. Evaluation operates at the page-pair level, rather than on individual cells or table fragments.

Reconstruction TEDS measures how accurately the reconstructed tables preserve structure and cell text. For each positive case with a valid output and a correct continuation decision, we normalize the reference and predicted HTML tables and find the one-to-one matching that maximizes their total TEDS. Unmatched reference tables receive zero, as do cases with invalid outputs or incorrect continuation decisions. Scores are summed and divided by the fixed total of 491 reference tables, so missed continuations remain in the denominator. The 399 negative pairs affect Relation F1 but not Reconstruction TEDS. On MPDocBench-Parse, we follow the benchmark's truncated-table TEDS protocol~\cite{zhou2026mpdocbench}, which does not impose this explicit continuation-decision gate. All metrics are reported on a 0--100 scale.

\ours{} and the general VLM baseline follow the same two-stage procedure: each model first parses the pages, then reconstructs cross-page tables using the page images and its own parsed representations. Specialized parsers use their provided inference pipelines. For document-level evaluation, we apply the recursive reconstruction procedure described in Section~\ref{sec:capability_crosspage}, with candidate page sequences determined from model predictions rather than reference annotations.

\begin{table}[!htbp]
\centering
\small
\setlength{\tabcolsep}{4pt}
\renewcommand{\arraystretch}{1.12}
\caption{Performance comparison on cross-page table reconstruction benchmarks.}
\label{tab:crosspage_results}
\begin{tabularx}{\linewidth}{@{}Yrrr@{}}
\toprule
& \multicolumn{2}{c}{\textbf{CrossTabBench}} & \textbf{MPDocBench-Parse} \\
\cmidrule(lr){2-3}\cmidrule(l){4-4}
\textbf{Model} & \makecell{\textbf{Relation F1}\\\textbf{(P/R)}} & \makecell{\textbf{Reconstruction}\\\textbf{TEDS}} & \makecell{\textbf{Truncated Table}\\\textbf{TEDS}} \\
\midrule
PaddleOCR-VL-1.6~\cite{model_paddle16} & 42.61 {\scriptsize (90.13/27.90)} & 24.24 & 83.39 \\
Logics-Parsing-V3~\cite{model_logicsv3} & 44.55 {\scriptsize (90.63/29.53)} & 26.10 & 88.05 \\
MinerU2.5-Pro~\cite{model_minerpro} & 59.14 {\scriptsize (93.81/43.18)} & 41.02 & 91.44 \\
MinerU2.5-Pro + MinerU-PoPo~\cite{model_minerpro,model_minerupopo} & 54.76 {\scriptsize (97.40/38.09)} & 36.42 & \secondcell{91.87} \\
Qwen3.8-Max~\cite{model_qwen38} & \secondcell{94.47} {\scriptsize (90.19/99.19)} & \bestcell{91.71} & 89.57 \\
\ours{} & \bestcell{97.07} {\scriptsize (96.20/97.96)} & \secondcell{89.66} & \bestcell{92.04} \\
\bottomrule
\end{tabularx}
\end{table}

Table~\ref{tab:crosspage_results} shows that \ours{} achieves a Relation F1 of 97.07 on CrossTabBench, exceeding Qwen3.8-Max by 2.60 points. Its higher precision and strong recall support accurate continuation detection while limiting false merges. \ours{} also achieves a Reconstruction TEDS of 89.66, substantially outperforming the specialized parser baselines, and leads on MPDocBench-Parse with a truncated-table TEDS of 92.04. Accurate continuation detection and reconstruction enable tables split across pages to be recovered as complete structures, preserving their row and column relationships in the parsed document.

\section{Conclusion}

In this report, we present \ours{}, a compact 0.8B model that unifies document parsing, general OCR, key information extraction, document figure captioning, and cross-page table reconstruction. We train the model on real and synthetic data using a four-stage recipe comprising continued pretraining, domain-balanced mid-training, multi-teacher on-policy distillation, and mixed-task reinforcement learning. Evaluated on public and in-house benchmarks, \ours{} achieves leading results across a broad range of OCR tasks, outperforming both general-purpose vision-language models and specialized OCR models on multiple benchmarks.

\Needspace{10\baselineskip}
\section*{Authors and Contributions}

\textbf{Core Contributors.}\quad Yu~Xia, Yinglun~Li, Yuhui~Chen, Jinlong~Huang, Yuyang~Wanyan, Shengze~Shi

\medskip
\textbf{Contributors.}\quad Yifei~Song, Yue~Duan, Shubo~Lin, Jianshan~Zhao, En~Lin

\medskip
\textbf{Project Leaders.}\quad Shiyin~Lu, Qing-Guo~Chen, Zhao~Xu, Weihua~Luo

\bibliographystyle{unsrt}
\setlength{\bibsep}{8pt plus 2pt minus 1pt}
\bibliography{references}

\clearpage
\appendix
\section{Qualitative Examples}
\label{sec:qualitative_examples}

This appendix presents qualitative examples of the five task families described in Section~\ref{sec:capabilities}.

\setcounter{figure}{0}
\renewcommand{\thefigure}{\thesection.\arabic{figure}}
\definecolor{exampleborder}{HTML}{8B919B}
\definecolor{examplelight}{HTML}{EAE7F3}
\definecolor{exampledark}{HTML}{26354B}

\newtcolorbox{ExamplePanel}[2][]{%
  sharp corners,
  boxrule=0.4pt,
  colframe=exampleborder,
  colback=white,
  colbacktitle=examplelight,
  coltitle=black,
  fonttitle=\sffamily\fontseries{bx}\selectfont\small,
  halign title=center,
  title={#2},
  left=3mm,right=3mm,top=3mm,bottom=3mm,
  before skip=0pt,after skip=0pt,
  #1}

\newcommand{\exampletext}[1]{%
  \VerbatimInput[fontsize=\footnotesize,breaklines=true,breakanywhere=true,
    breaksymbolleft={},baselinestretch=1.05]{#1}}

\subsection{Document Parsing}
\label{sec:qualitative_document_parsing}

\begin{figure}[!htbp]
\centering
\begin{minipage}[t]{0.49\linewidth}
  \begin{ExamplePanel}[height=8.7cm,valign=center]{Input Image}
    \centering
    \includegraphics[width=\linewidth,height=7.5cm,keepaspectratio]{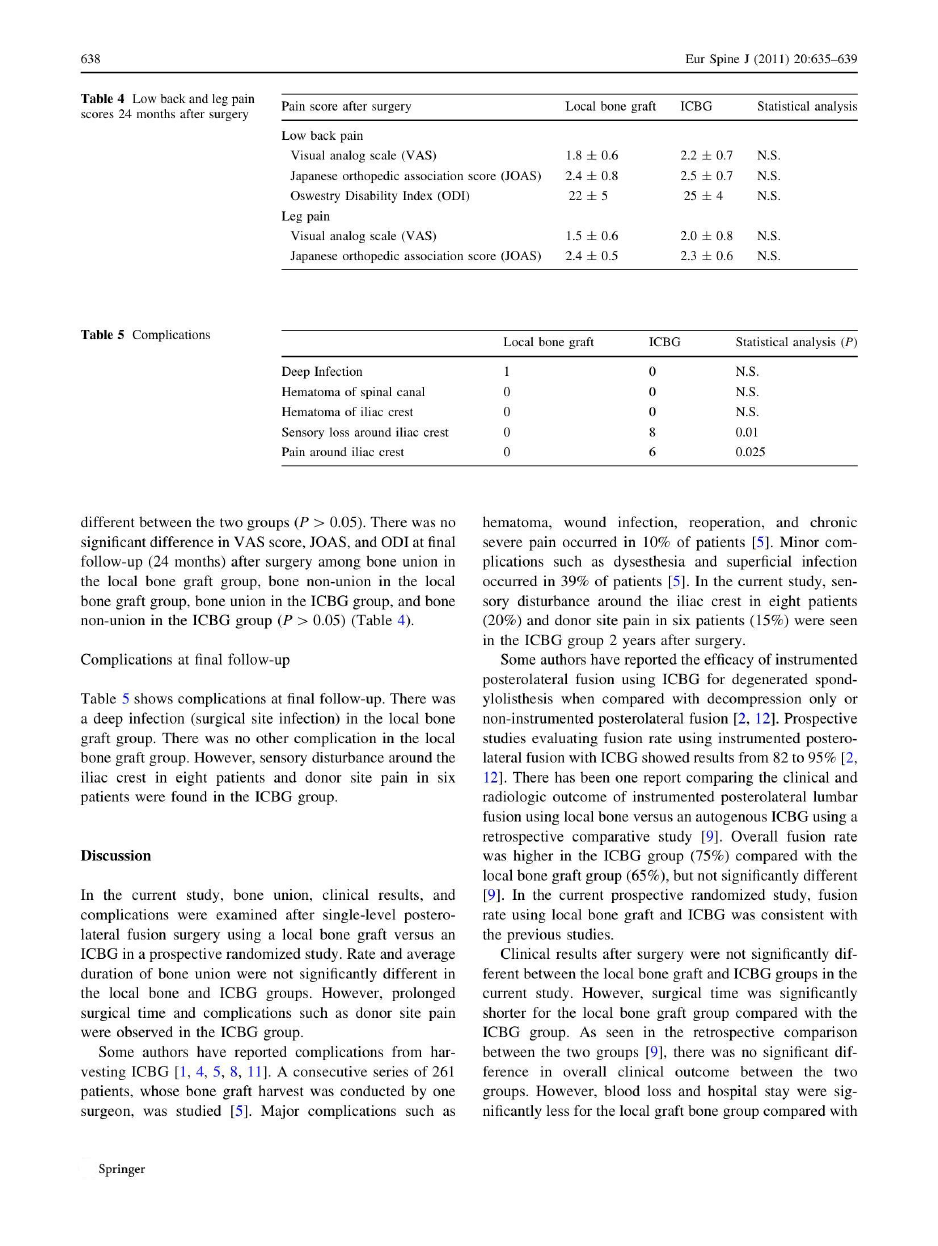}
  \end{ExamplePanel}
\end{minipage}\hfill
\begin{minipage}[t]{0.49\linewidth}
  \begin{ExamplePanel}[height=8.7cm,valign=center,colbacktitle=exampledark,coltitle=white]{Predicted BBox and Category}
    \centering
    \includegraphics[width=\linewidth,height=7.5cm,keepaspectratio]{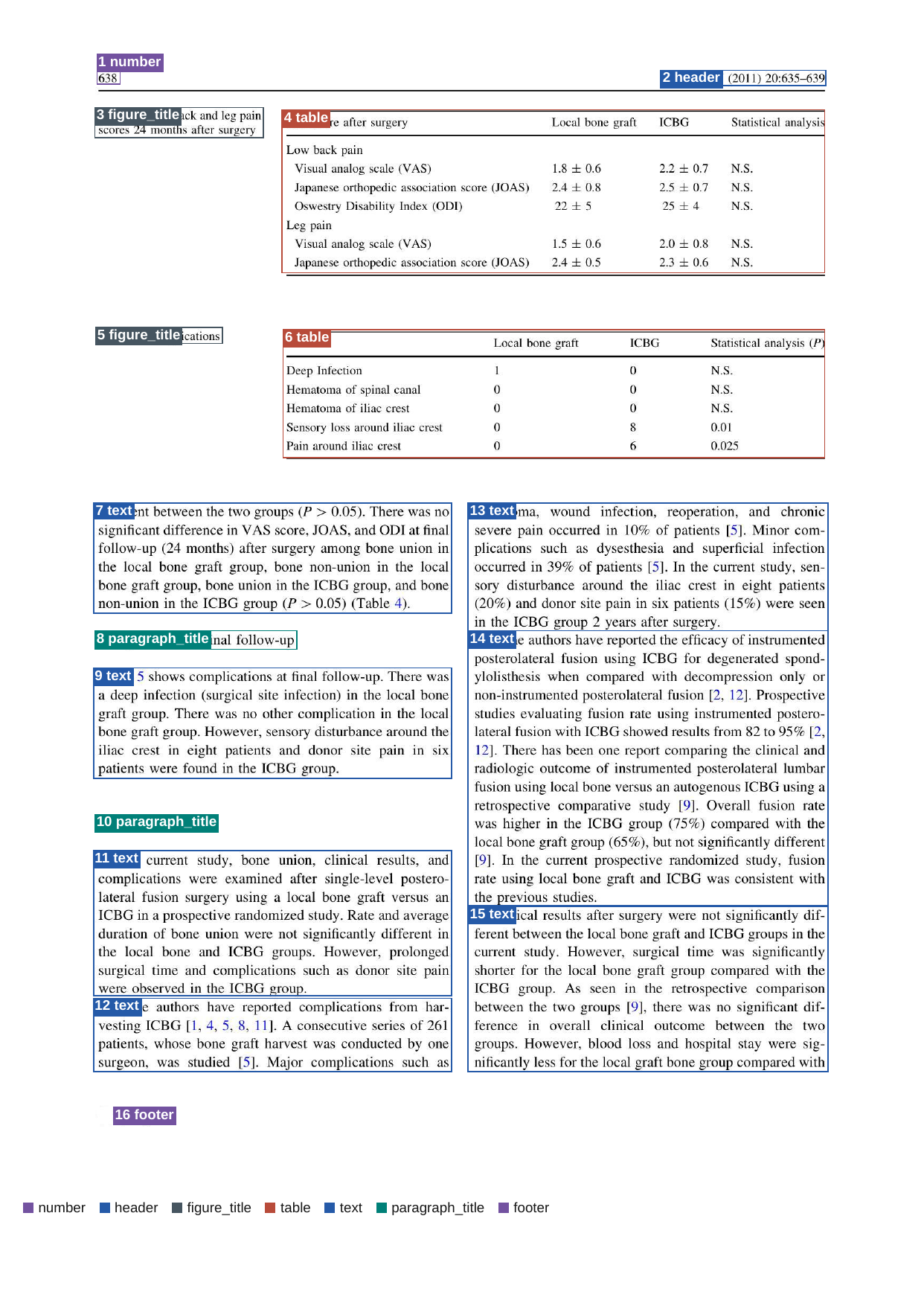}
  \end{ExamplePanel}
\end{minipage}

\vspace{2mm}
\begin{minipage}[t]{0.49\linewidth}
  \begin{ExamplePanel}[height=9cm]{Prompt}
\begin{Verbatim}
Analyze the document structure and extract text content. Return a single JSON array in natural reading order.

Each element follows this schema:
`{"bbox": [x_min, y_min, x_max, y_max], "category": "string", "content": "string"}`

Rules:
1. The bbox coordinates are integers scaled to [0, 1000).
2. The category is exactly one of:
["abstract", "algorithm", "aside_text", "chart", "content", "display_formula", "doc_title", "figure_title", "footer", "footer_image", "footnote", "formula_number", "header", "header_image", "image", "inline_formula", "number", "paragraph_title", "reference_content", "seal", "table", "text", "vertical_text", "vision_footnote"].
3. The content formatting depends on category:
- For "footer_image", "header_image", "image", or "chart": write as an HTML image tag <img src="images/bbox_{x_min}_{y_min}_{x_max}_{y_max}.jpg" />.
- For "display_formula" or "inline_formula": write in LaTeX.
- For "table": write as a standard HTML `<table>...</table>`.
- Otherwise: write in Markdown.
4. Output must be strictly valid JSON. Properly escape quotes and special characters inside content.
\end{Verbatim}
  \end{ExamplePanel}
\end{minipage}\hfill
\begin{minipage}[t]{0.49\linewidth}
  \begin{ExamplePanel}[height=9cm,valign=center,colbacktitle=exampledark,coltitle=white]{OvisOCR3 Output}
    \centering
    \includegraphics[width=\linewidth,height=7.8cm,keepaspectratio]{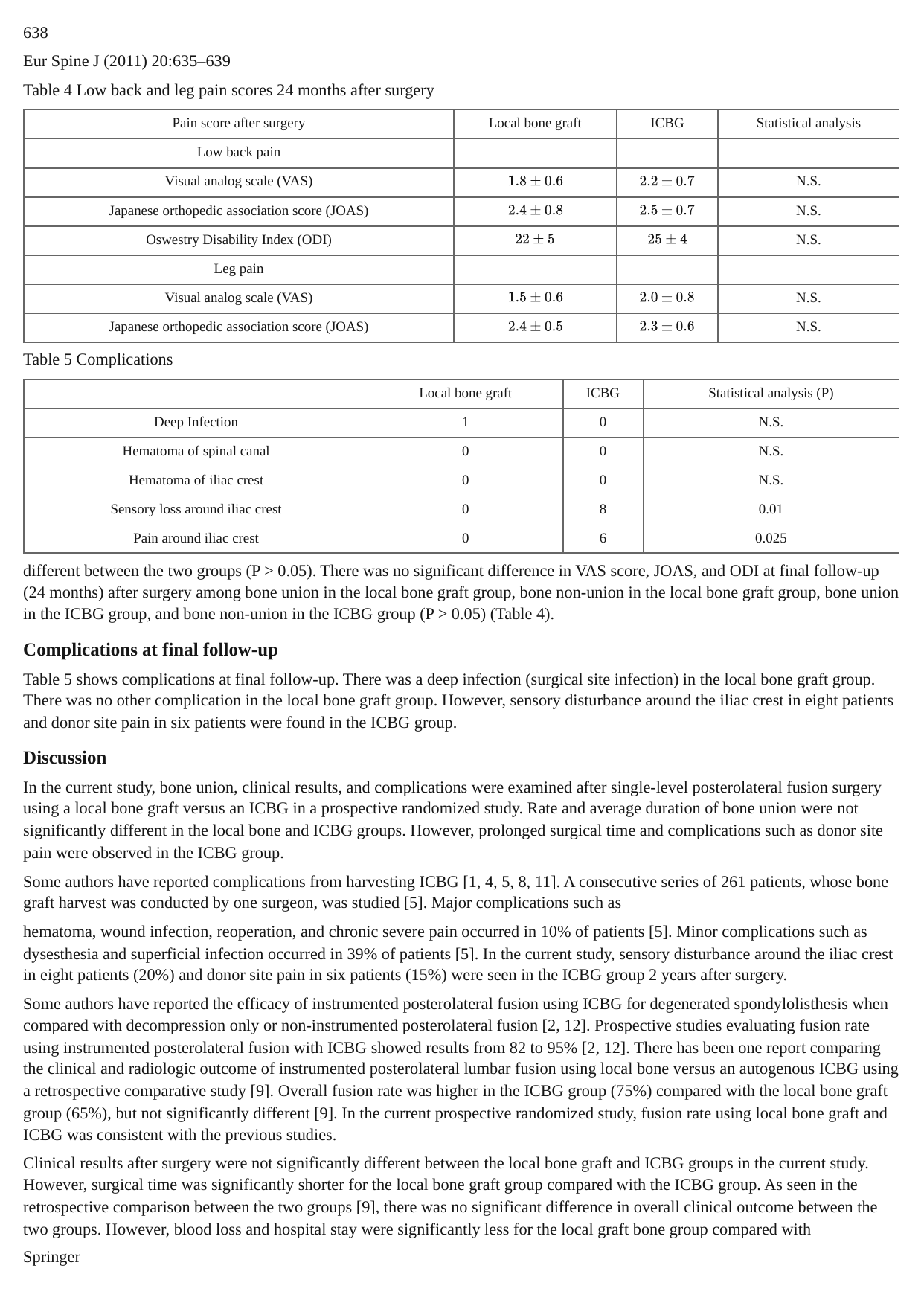}
  \end{ExamplePanel}
\end{minipage}
\par\vspace{1mm}
\caption{Structured document parsing with predicted region boundaries, semantic categories, and rendered page content.}
\label{fig:example_docparse_structured}
\end{figure}

\begin{figure}[!htbp]
  \centering
  \begin{minipage}[t]{0.49\linewidth}
    \vspace{0pt}%
    \begin{ExamplePanel}[equal height group=docparse-layout,valign=top]{Input Image}
      \centering
      \includegraphics[width=\linewidth,height=14.8cm,keepaspectratio]{examples/document_parsing/layout/input.pdf}
    \end{ExamplePanel}
  \end{minipage}\hfill
  \begin{minipage}[t]{0.49\linewidth}
    \vspace{0pt}%
    \begin{ExamplePanel}[equal height group=docparse-layout,valign=top,colbacktitle=exampledark,coltitle=white]{OvisOCR3 Predicted Layout}
      \centering
      \includegraphics[width=\linewidth,height=14.8cm,keepaspectratio]{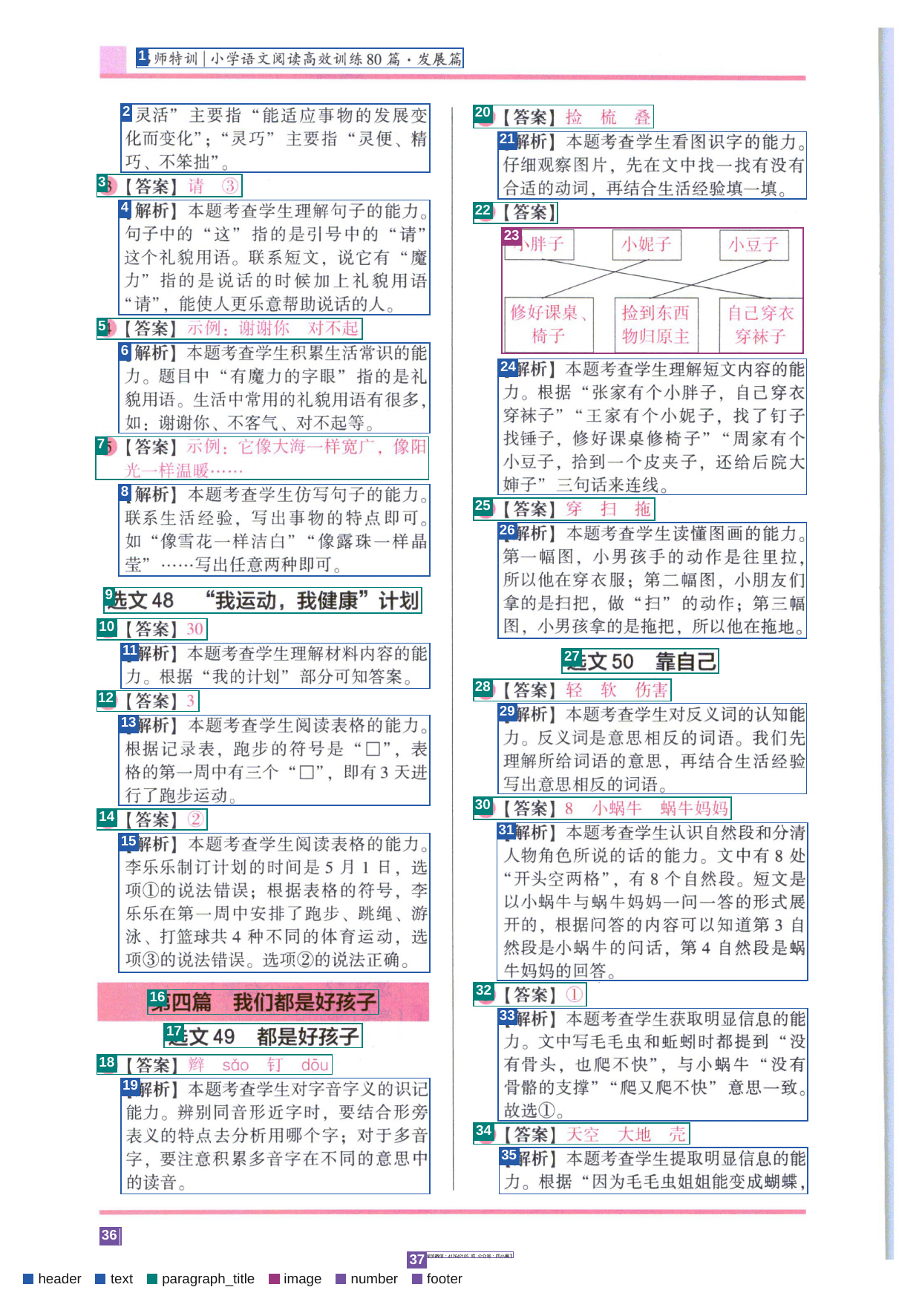}
    \end{ExamplePanel}
  \end{minipage}

  \vspace{2mm}
  \begin{ExamplePanel}{Prompt: Document layout detection}
\begin{Verbatim}
Analyze the document structure and return a single JSON array in natural reading order.

Each element follows this schema:
`{"bbox": [x_min, y_min, x_max, y_max], "category": "string"}`

Rules:
1. The bbox coordinates are integers scaled to [0, 1000).
2. The category is exactly one of:
["abstract", "algorithm", "aside_text", "chart", "content", "display_formula", "doc_title", "figure_title", "footer", "footer_image", "footnote", "formula_number", "header", "header_image", "image", "inline_formula", "number", "paragraph_title", "reference_content", "seal", "table", "text", "vertical_text", "vision_footnote"].
\end{Verbatim}
  \end{ExamplePanel}
  \par\vspace{1mm}
  \caption{Document layout detection. Box colors denote predicted region categories; numeric labels indicate reading order.}
  \label{fig:example_docparse_layout}
  \end{figure}

\begin{figure}[!htbp]
  \centering
  \begin{minipage}[t]{0.49\linewidth}
    \vspace{0pt}%
    \begin{ExamplePanel}[equal height group=docparse-markdown,valign=top]{Input Image}
      \centering
      \includegraphics[width=\linewidth,height=14.8cm,keepaspectratio]{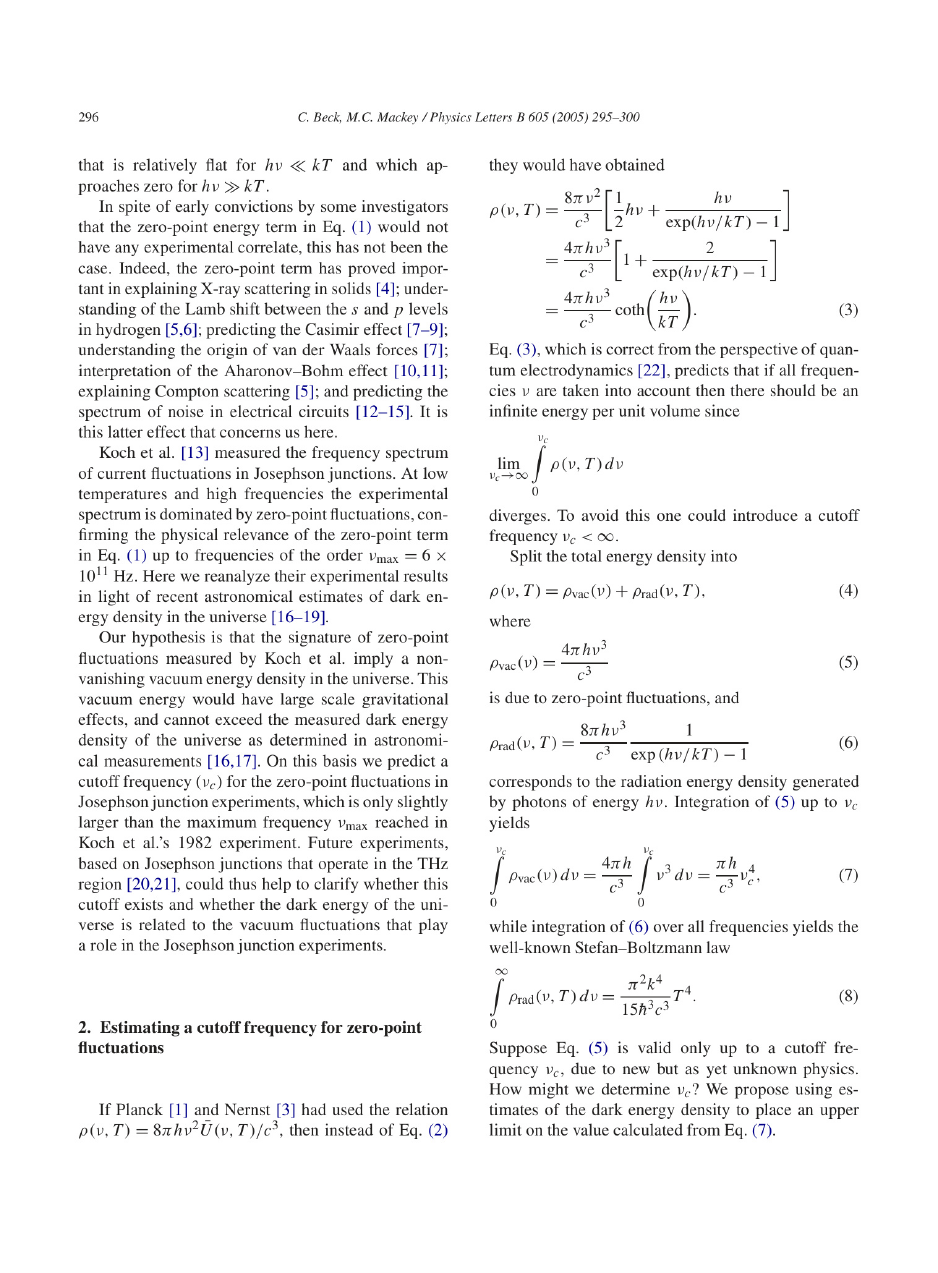}
    \end{ExamplePanel}
  \end{minipage}\hfill
  \begin{minipage}[t]{0.49\linewidth}
    \vspace{0pt}%
    \begin{ExamplePanel}[equal height group=docparse-markdown,valign=top,colbacktitle=exampledark,coltitle=white]{OvisOCR3 Output}
      \centering
      \includegraphics[width=\linewidth,height=14.8cm,keepaspectratio]{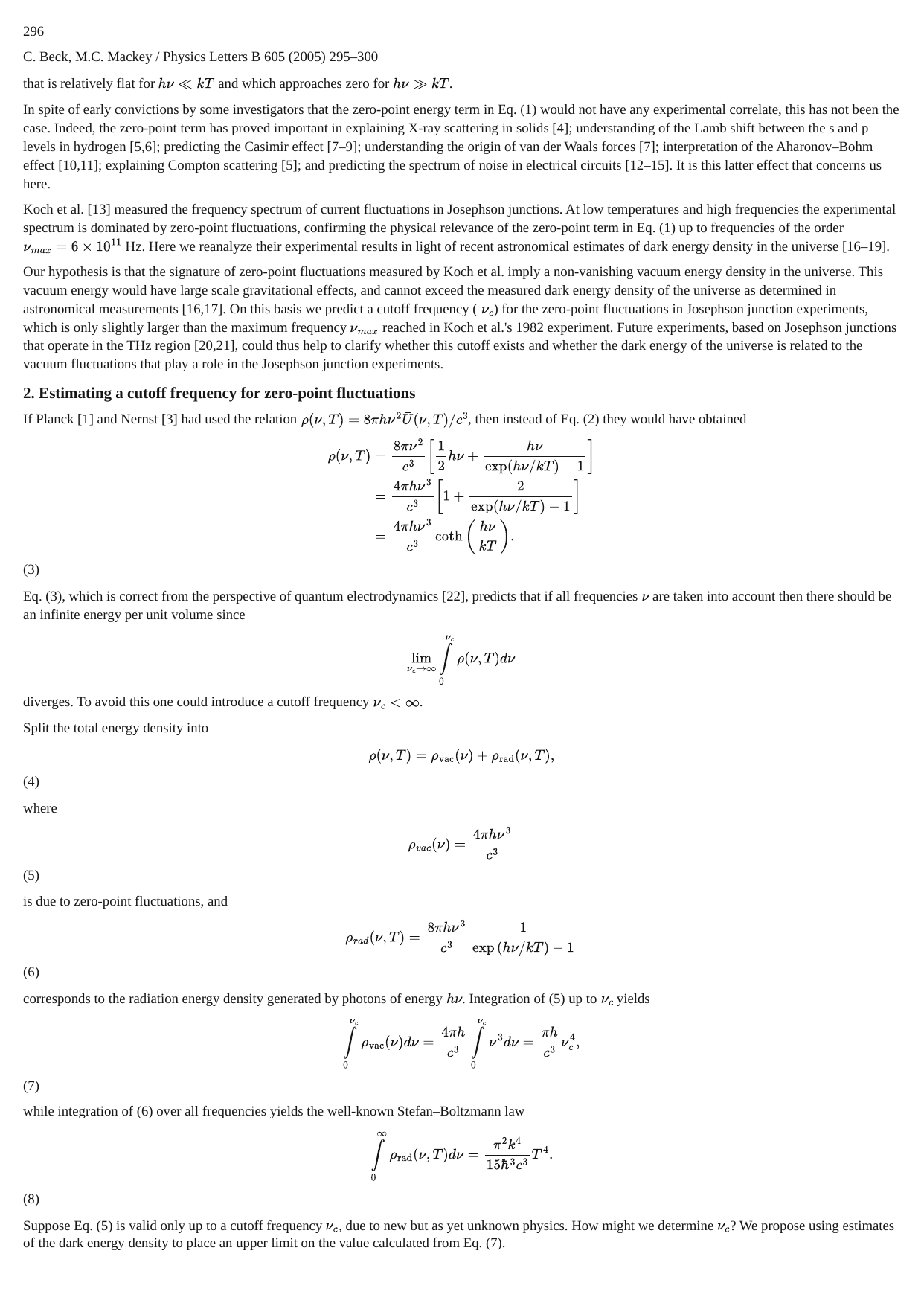}
    \end{ExamplePanel}
  \end{minipage}

  \vspace{2mm}
  \begin{ExamplePanel}{Prompt: Markdown document parsing}
\begin{Verbatim}
Extract all readable content from the image in natural human reading order and output the result as a single Markdown document. For charts or images, represent them using an HTML image tag: <img src="images/bbox_{left}_{top}_{right}_{bottom}.jpg" />, where left, top, right, bottom are bounding box coordinates scaled to [0, 1000). Format formulas as LaTeX. Format tables as HTML: <table>...</table>. Transcribe all other text as standard Markdown. Preserve the original text without translation or paraphrasing.
\end{Verbatim}
  \end{ExamplePanel}
  \par\vspace{1mm}
  \caption{Markdown document parsing, with text in reading order and mathematical expressions rendered from LaTeX.}
  \label{fig:example_docparse_markdown}
  \end{figure}

\begin{figure}[!htbp]
  \centering
  \begin{minipage}[t]{0.49\linewidth}
    \vspace{0pt}%
    \begin{ExamplePanel}[equal height group=docparse-photo_financial,valign=top]{Input Image}
      \centering
      \includegraphics[width=\linewidth,height=14.8cm,keepaspectratio]{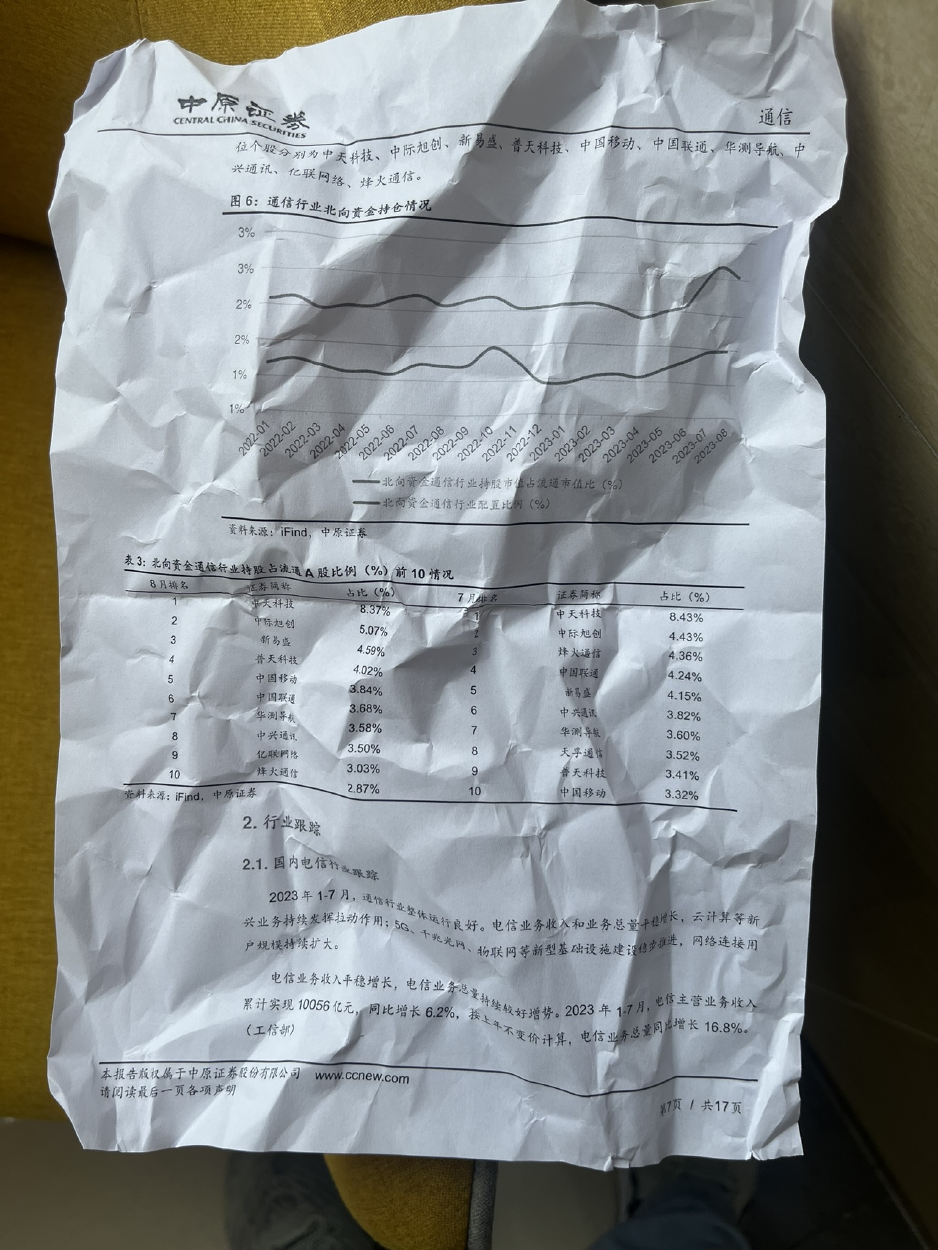}
    \end{ExamplePanel}
  \end{minipage}\hfill
  \begin{minipage}[t]{0.49\linewidth}
    \vspace{0pt}%
    \begin{ExamplePanel}[equal height group=docparse-photo_financial,valign=top,colbacktitle=exampledark,coltitle=white]{OvisOCR3 Output}
      \centering
      \includegraphics[width=\linewidth,height=14.8cm,keepaspectratio]{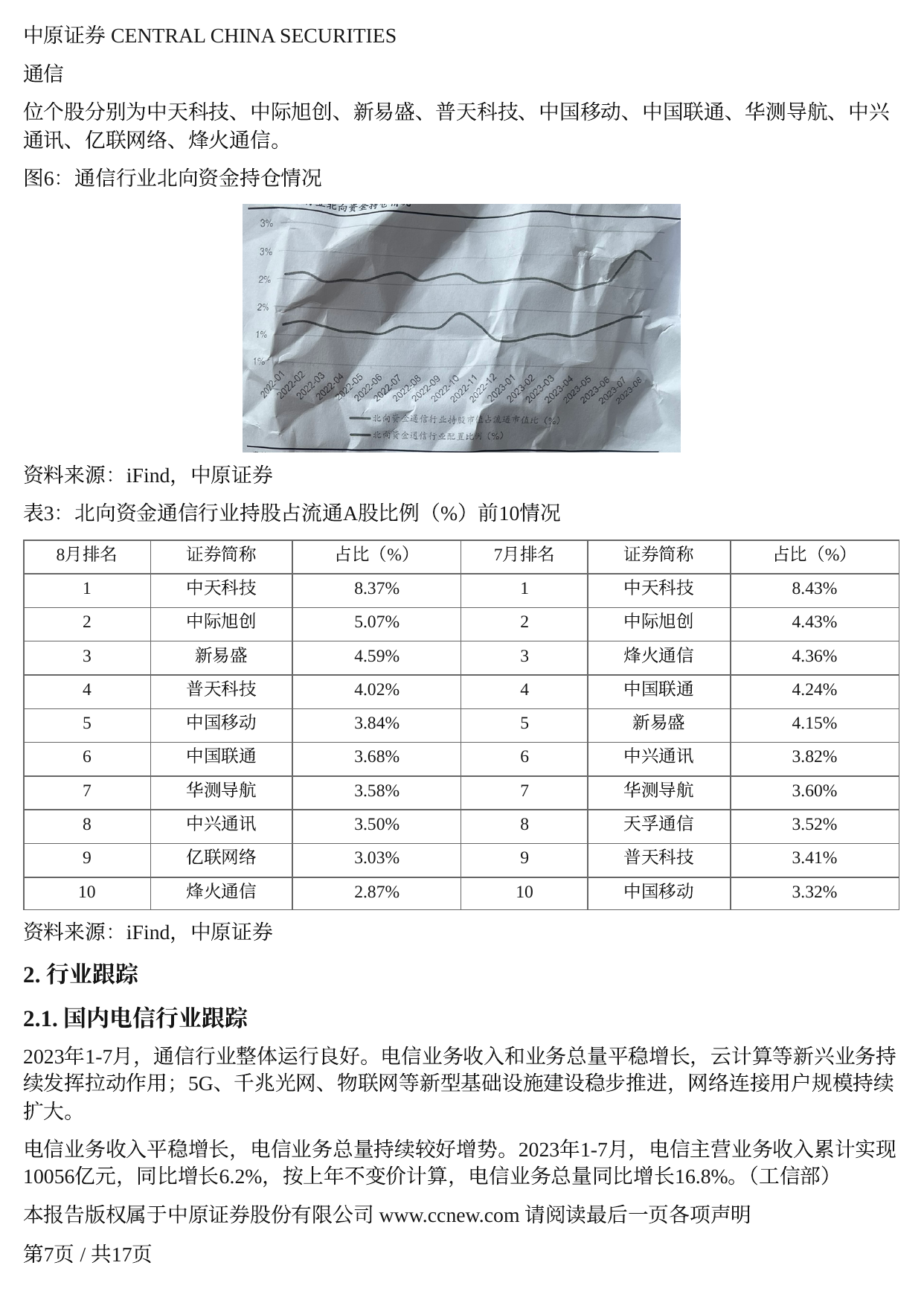}
    \end{ExamplePanel}
  \end{minipage}

  \vspace{2mm}
  \begin{ExamplePanel}{Prompt: Markdown document parsing}
\begin{Verbatim}
Extract all readable content from the image in natural human reading order and output the result as a single Markdown document. For charts or images, represent them using an HTML image tag: <img src="images/bbox_{left}_{top}_{right}_{bottom}.jpg" />, where left, top, right, bottom are bounding box coordinates scaled to [0, 1000). Format formulas as LaTeX. Format tables as HTML: <table>...</table>. Transcribe all other text as standard Markdown. Preserve the original text without translation or paraphrasing.
\end{Verbatim}
  \end{ExamplePanel}
  \par\vspace{1mm}
  \caption{Markdown parsing of a photographed document with page deformation and uneven illumination.}
  \label{fig:example_docparse_photo_financial}
  \end{figure}

\begin{figure}[!htbp]
  \centering
  \begin{minipage}[t]{0.49\linewidth}
    \vspace{0pt}%
    \begin{ExamplePanel}[equal height group=docparse-photo_lowlight,valign=top]{Input Image}
      \centering
      \includegraphics[width=\linewidth,height=14.8cm,keepaspectratio]{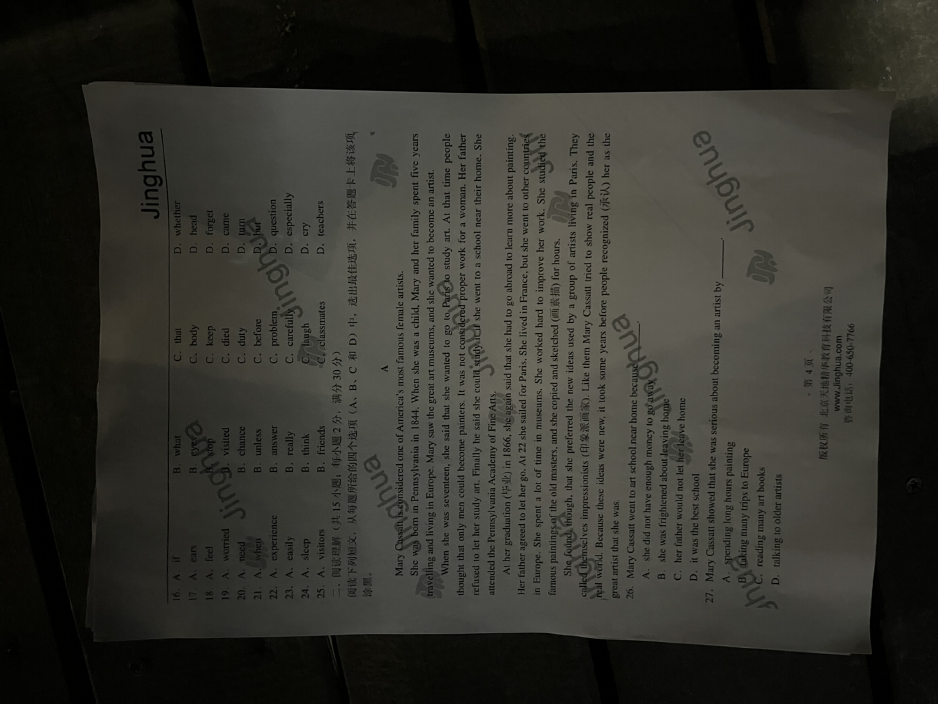}
    \end{ExamplePanel}
  \end{minipage}\hfill
  \begin{minipage}[t]{0.49\linewidth}
    \vspace{0pt}%
    \begin{ExamplePanel}[equal height group=docparse-photo_lowlight,valign=top,colbacktitle=exampledark,coltitle=white]{OvisOCR3 Output}
      \centering
      \includegraphics[width=\linewidth,height=14.8cm,keepaspectratio]{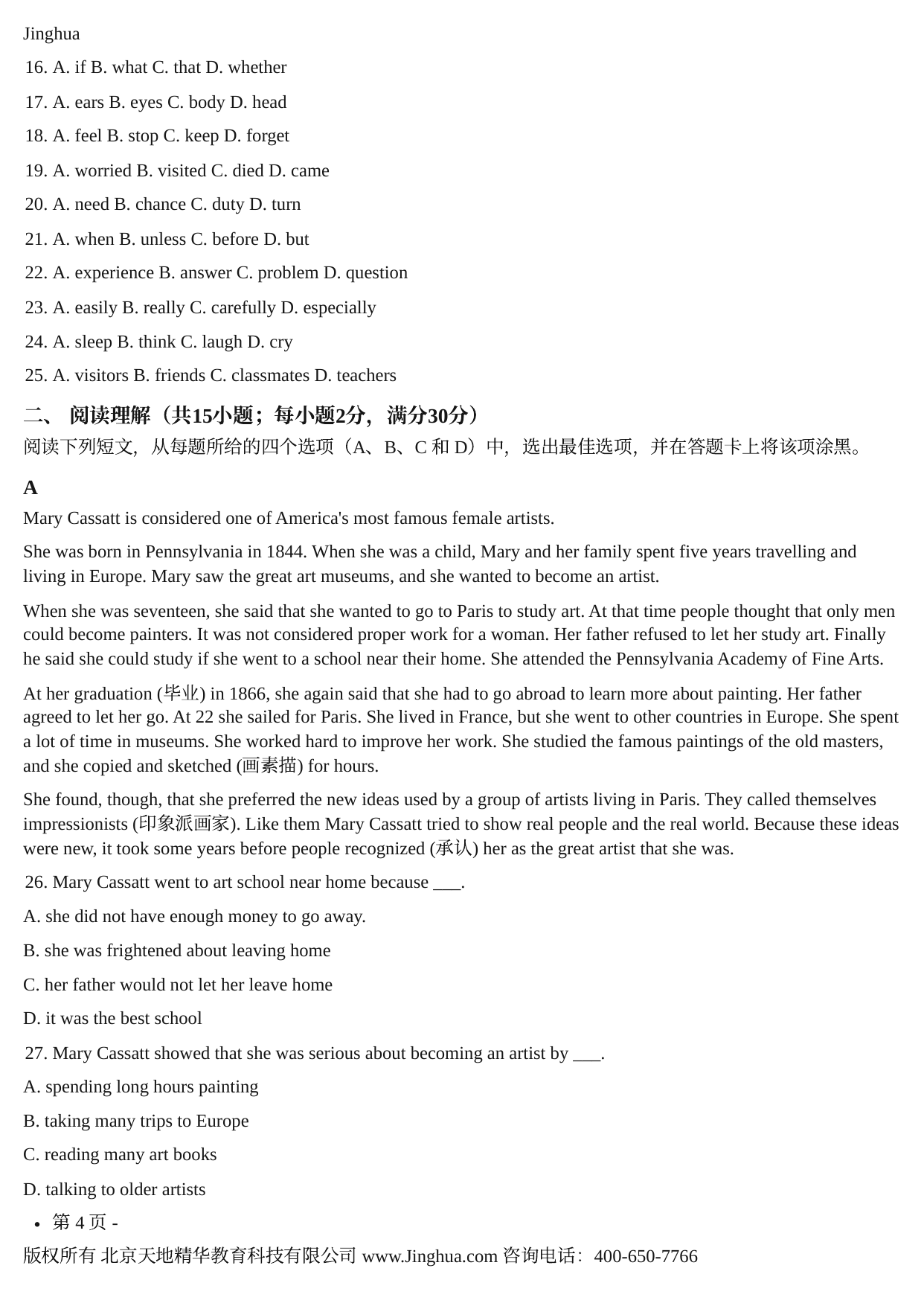}
    \end{ExamplePanel}
  \end{minipage}

  \vspace{2mm}
  \begin{ExamplePanel}{Prompt: Markdown document parsing}
\begin{Verbatim}
Extract all readable content from the image in natural human reading order and output the result as a single Markdown document. For charts or images, represent them using an HTML image tag: <img src="images/bbox_{left}_{top}_{right}_{bottom}.jpg" />, where left, top, right, bottom are bounding box coordinates scaled to [0, 1000). Format formulas as LaTeX. Format tables as HTML: <table>...</table>. Transcribe all other text as standard Markdown. Preserve the original text without translation or paraphrasing.
\end{Verbatim}
  \end{ExamplePanel}
  \par\vspace{1mm}
  \caption{Markdown parsing of a photographed document affected by low light and perspective distortion.}
  \label{fig:example_docparse_photo_lowlight}
  \end{figure}

\begin{figure}[!htbp]
  \centering
  \begin{minipage}[t]{0.49\linewidth}
    \vspace{0pt}%
    \begin{ExamplePanel}[equal height group=docparse-japanese,valign=top]{Input Image}
      \centering
      \includegraphics[width=\linewidth,height=14.8cm,keepaspectratio]{examples/document_parsing/japanese/input.pdf}
    \end{ExamplePanel}
  \end{minipage}\hfill
  \begin{minipage}[t]{0.49\linewidth}
    \vspace{0pt}%
    \begin{ExamplePanel}[equal height group=docparse-japanese,valign=top,colbacktitle=exampledark,coltitle=white]{OvisOCR3 Output}
      \centering
      \includegraphics[width=\linewidth,height=14.8cm,keepaspectratio]{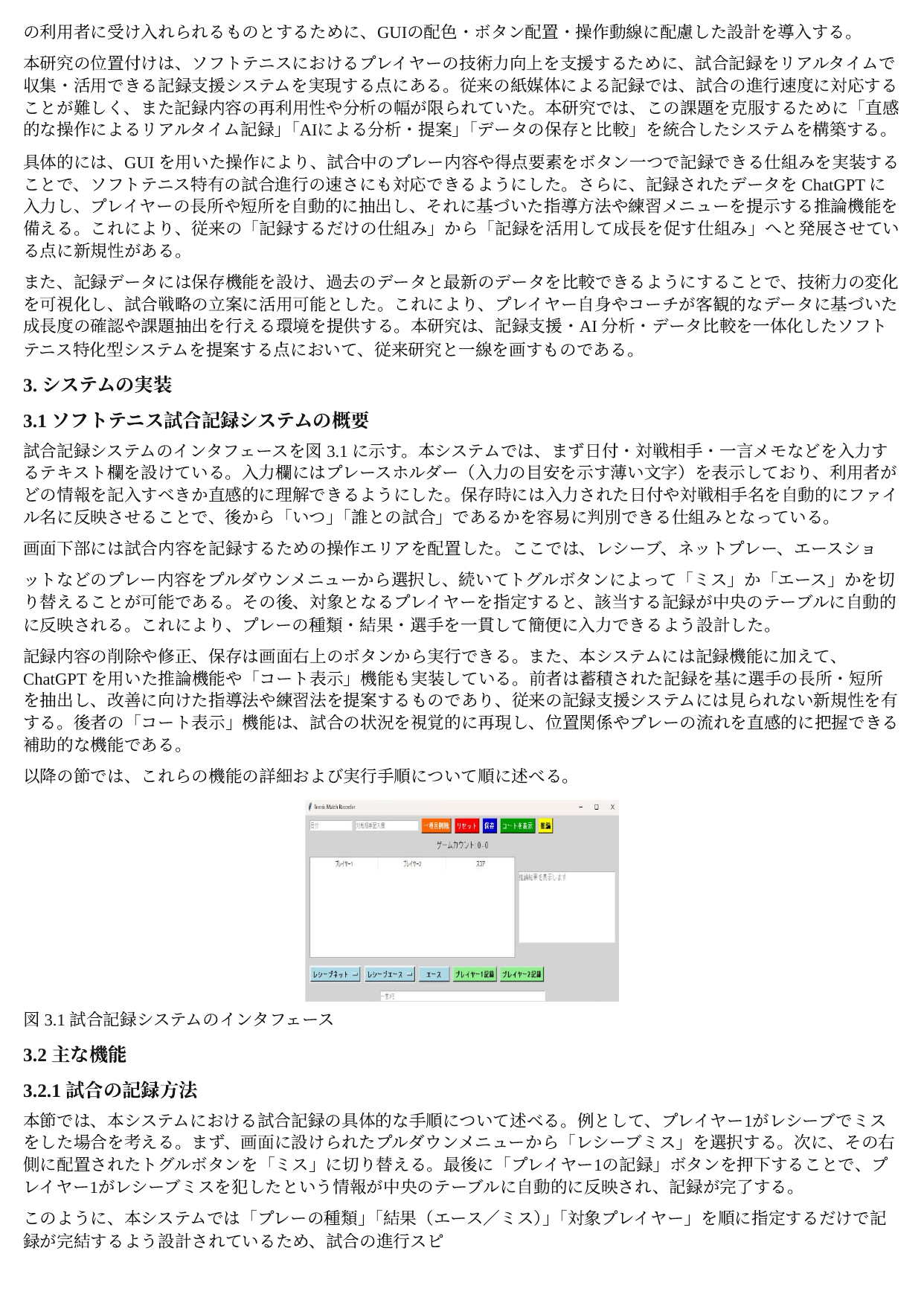}
    \end{ExamplePanel}
  \end{minipage}

  \vspace{2mm}
  \begin{ExamplePanel}{Prompt: Markdown document parsing}
\begin{Verbatim}
Extract all readable content from the image in natural human reading order and output the result as a single Markdown document. For charts or images, represent them using an HTML image tag: <img src="images/bbox_{left}_{top}_{right}_{bottom}.jpg" />, where left, top, right, bottom are bounding box coordinates scaled to [0, 1000). Format formulas as LaTeX. Format tables as HTML: <table>...</table>. Transcribe all other text as standard Markdown. Preserve the original text without translation or paraphrasing.
\end{Verbatim}
  \end{ExamplePanel}
  \par\vspace{1mm}
  \caption{Markdown document parsing of a Japanese page.}
  \label{fig:example_docparse_japanese}
  \end{figure}

Figures~\ref{fig:example_docparse_structured}--\ref{fig:example_docparse_japanese} illustrate the three parsing modes in Section~\ref{sec:capability_document_parsing}, with additional examples of photographed documents and a Japanese page. Each example shows the input, prompt, and complete rendered model output, including image crops referenced in the response.

\FloatBarrier

\clearpage
  \FloatBarrier

\clearpage
  \FloatBarrier

\clearpage
  \FloatBarrier

\clearpage
  \FloatBarrier

\clearpage
  \FloatBarrier

\clearpage

\subsection{General OCR}
\label{sec:qualitative_general_ocr}
\begin{figure}[!htbp]
\centering
\begin{minipage}[t]{0.37\linewidth}
\begin{ExamplePanel}[height=12cm,valign=center]{Input Image}
\centering
\begin{tikzpicture}
\node[anchor=south west,inner sep=0] (page) at (0,0)
  {\includegraphics[width=\linewidth,height=10.3cm,keepaspectratio]{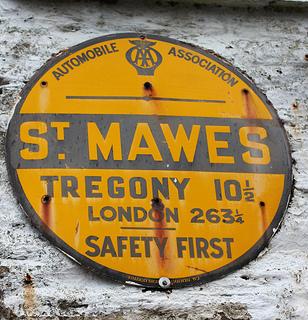}};
\begin{scope}[x={(page.south east)},y={(page.north west)}]
\draw[green,line width=0.8pt] (0.15000,0.74400) rectangle (0.39100,0.87500);
\end{scope}
\end{tikzpicture}
\end{ExamplePanel}
\end{minipage}\hfill
\begin{minipage}[t]{0.61\linewidth}
\begin{ExamplePanel}[height=12cm]{Prompt}
\begin{Verbatim}
Locate every visible region corresponding to the requested reference or references in the image.

Requested references (JSON array, source order):
["AUTOMOBILE"]

Return a single JSON array in the requested reference order.

Each element follows this schema:
`{"bbox": [x_min, y_min, x_max, y_max], "reference": "string"}`

Rules:
1. Each element must represent exactly one requested reference and one corresponding visible region. If one requested reference corresponds to multiple separate regions, output one element for each bbox and repeat the same reference.
2. The bbox coordinates are integers scaled to [0, 1000).
3. Each bbox must tightly enclose its corresponding referenced entity or region and must not include unrelated content.
4. Copy each reference exactly from the requested-references array without translation, paraphrasing, correction, or completion. Do not output any reference that was not requested.
5. Do not add leading or trailing whitespace to reference.
6. Return bounding boxes only. Do not return points, polygons, masks, explanations, or confidence scores.
7. Every requested reference must appear in at least one output element.
8. Output strictly valid JSON only. Do not output Markdown, explanations, or code fences. Properly escape quotes, backslashes, and other special characters inside reference.
\end{Verbatim}
\end{ExamplePanel}
\end{minipage}
\par\vspace{3mm}
\begin{ExamplePanel}[colbacktitle=exampledark,coltitle=white]{OvisOCR3 Output: Reference Grounding}
\exampletext{examples/general_ocr/reference_grounding_8445/output_display.txt}
\end{ExamplePanel}
\par\vspace{2mm}
\caption{Reference grounding for the supplied word \texttt{AUTOMOBILE}, returning its bounding box.}
\label{fig:general_ocr_reference_grounding}
\end{figure}

\begin{figure}[!htbp]
\centering
\begin{minipage}[t]{0.37\linewidth}
\begin{ExamplePanel}[height=12cm,valign=center]{Input Image}
\centering
\begin{tikzpicture}
\node[anchor=south west,inner sep=0] (page) at (0,0)
  {\includegraphics[width=\linewidth,height=10.3cm,keepaspectratio]{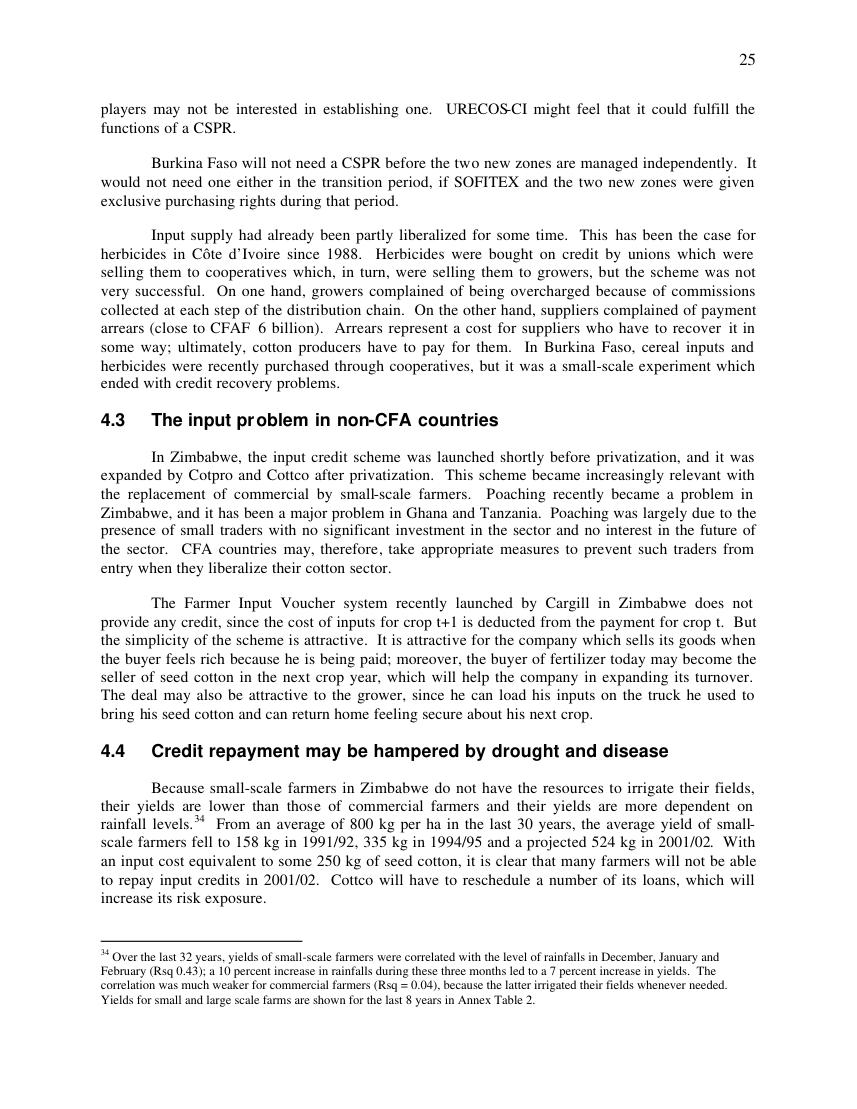}};
\begin{scope}[x={(page.south east)},y={(page.north west)}]
\draw[orange!85!black,line width=0.8pt] (0.11200,0.80700) rectangle (0.89200,0.86600);
\end{scope}
\end{tikzpicture}
\end{ExamplePanel}
\end{minipage}\hfill
\begin{minipage}[t]{0.61\linewidth}
\begin{ExamplePanel}[height=12cm]{Prompt}
\begin{Verbatim}
Transcribe all readable text in the specified image region. Directly output the content.
Target region: [112,134,892,193]

Rules:
1. Read only the visible text inside the target region.
2. Preserve the original text exactly without translation, paraphrasing, correction, completion, or omission.
3. Follow the natural human reading order within the target region.
4. Output plain text only. Use spaces and line breaks only where needed to preserve text separation and reading order.
5. Do not output the target region, bounding boxes, coordinates, category labels, Markdown formatting, HTML, JSON, explanations, or code fences.
\end{Verbatim}
\end{ExamplePanel}
\end{minipage}
\par\vspace{3mm}
\par\medskip\textsf{\small Detail of the supplied region (same original image)}\par\smallskip
\begin{tikzpicture}
\begin{scope}
\clip (0,0) rectangle (1.000000\linewidth,0.097883\linewidth);
\node[anchor=south west,inner sep=0] at (-0.143590\linewidth,-1.338843\linewidth) {\includegraphics[width=1.282051\linewidth]{examples/general_ocr/region_ocr_7113/input.jpg}};
\end{scope}
\pgfresetboundingbox
\path[use as bounding box] (0,0) rectangle (1.000000\linewidth,0.097883\linewidth);
\end{tikzpicture}
\par\vspace{3mm}
\begin{ExamplePanel}[colbacktitle=exampledark,coltitle=white]{OvisOCR3 Output: Region OCR}
\exampletext{examples/general_ocr/region_ocr_7113/output_display.txt}
\end{ExamplePanel}
\par\vspace{2mm}
\caption{Region OCR on a specified paragraph, excluding surrounding text from the response.}
\label{fig:general_ocr_region_ocr}
\end{figure}

\begin{figure}[!htbp]
\centering
\begin{minipage}[t]{0.37\linewidth}
\begin{ExamplePanel}[height=12cm,valign=center]{Input Image}
\centering
\begin{tikzpicture}
\node[anchor=south west,inner sep=0] (page) at (0,0)
  {\includegraphics[width=\linewidth,height=10.3cm,keepaspectratio]{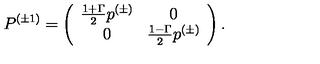}};
\end{tikzpicture}
\end{ExamplePanel}
\end{minipage}\hfill
\begin{minipage}[t]{0.61\linewidth}
\begin{ExamplePanel}[height=12cm]{Prompt}
\begin{Verbatim}
Recognize and transcribe every mathematical expression in the image as LaTeX. Directly output the formula content.

Rules:
1. Preserve every visible symbol, operator, delimiter, superscript, subscript, accent, matrix row, alignment, and line break without translation, paraphrasing, correction, completion, or omission.
2. Wrap each standalone or display expression as `$$ ... $$`.
3. If multiple standalone expressions are visible, output one `$$ ... $$` block per expression in natural reading order, separated by a single line break.
4. Use one decoded backslash for a LaTeX command and two decoded backslashes for a LaTeX row break. JSON serialization may escape these backslashes on disk, but the decoded target must retain valid LaTeX semantics.
5. Output formula content only. Do not output prose, explanations, translations, Markdown code fences, HTML, document-level table markup, or JSON.
\end{Verbatim}
\end{ExamplePanel}
\end{minipage}
\par\vspace{3mm}
\begin{ExamplePanel}[colbacktitle=exampledark,coltitle=white]{OvisOCR3 Output: Formula Recognition}
\exampletext{examples/general_ocr/formula_4835/output_display.txt}
\end{ExamplePanel}
\par\medskip\textsf{\small Rendered LaTeX (same expression)}\par\smallskip
\resizebox{0.94\linewidth}{!}{$\displaystyle P ^ { ( \pm 1 ) } = \left( \begin{array} { c c } { \frac { 1 + \Gamma } { 2 } p ^ { ( \pm ) } } & { 0 } \\ { 0 } & { \frac { 1 - \Gamma } { 2 } p ^ { ( \pm ) } } \\ \end{array} \right) .$}
\par\vspace{2mm}
\caption{Formula recognition with LaTeX output and its rendered matrix expression.}
\label{fig:general_ocr_formula}
\end{figure}

\begin{figure}[!htbp]
\centering
\begin{minipage}[t]{0.37\linewidth}
\begin{ExamplePanel}[height=12cm,valign=center]{Input Image}
\centering
\begin{tikzpicture}
\node[anchor=south west,inner sep=0] (page) at (0,0)
  {\includegraphics[width=\linewidth,height=10.3cm,keepaspectratio]{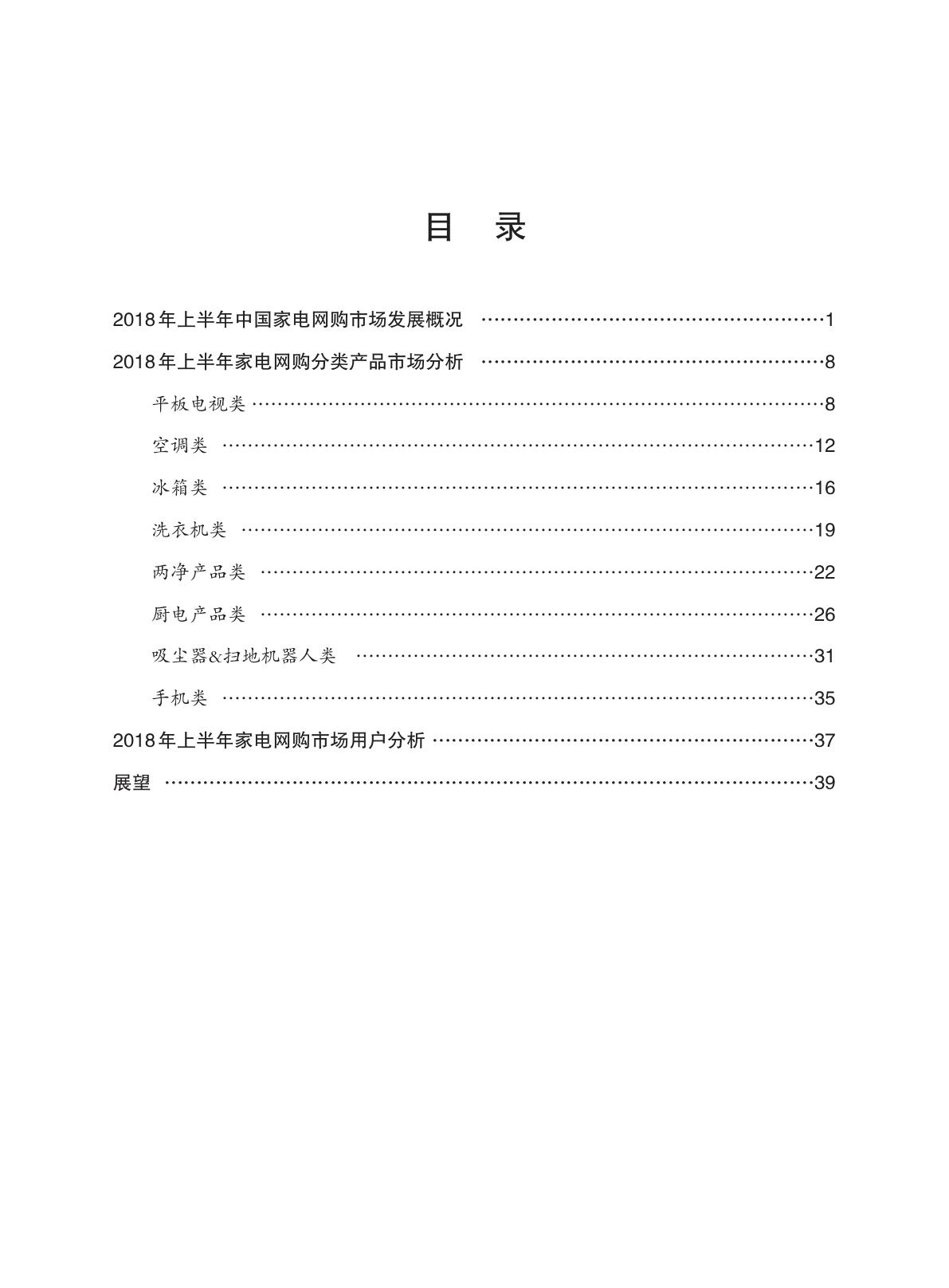}};
\begin{scope}[x={(page.south east)},y={(page.north west)}]
\draw[orange!85!black,line width=0.8pt] (0.15600,0.57700) rectangle (0.24300,0.60800);
\end{scope}
\end{tikzpicture}
\end{ExamplePanel}
\end{minipage}\hfill
\begin{minipage}[t]{0.61\linewidth}
\begin{ExamplePanel}[height=12cm]{Prompt}
\begin{Verbatim}
Translate all readable text in the specified image region into English. Directly output the translated content.
Target region: [156,392,243,423]

Rules:
1. Translate only the visible text inside the target region.
2. Preserve the original meaning, names, numbers, symbols, and line order. Do not add explanations, corrections, or omitted context.
3. Keep content that should not be translated unchanged.
4. Output plain translated text only. Use spaces and line breaks only where needed to preserve the source separation.
5. Do not output the source text, language labels, bounding boxes, Markdown, JSON, or code fences.
\end{Verbatim}
\end{ExamplePanel}
\end{minipage}
\par\vspace{3mm}
\par\medskip\textsf{\small Detail of the supplied region (same original image)}\par\smallskip
\begin{tikzpicture}
\begin{scope}
\clip (0,0) rectangle (0.650000\linewidth,0.314257\linewidth);
\node[anchor=south west,inner sep=0] at (-1.165517\linewidth,-5.849241\linewidth) {\includegraphics[width=7.471264\linewidth]{examples/general_ocr/translation_9739/input.jpg}};
\end{scope}
\pgfresetboundingbox
\path[use as bounding box] (0,0) rectangle (0.650000\linewidth,0.314257\linewidth);
\end{tikzpicture}
\par\vspace{3mm}
\begin{ExamplePanel}[colbacktitle=exampledark,coltitle=white]{OvisOCR3 Output: Text Translation}
\exampletext{examples/general_ocr/translation_9739/output_display.txt}
\end{ExamplePanel}
\par\vspace{2mm}
\caption{Text translation from Chinese to English within the selected region.}
\label{fig:general_ocr_translation}
\end{figure}

\begin{figure}[!htbp]
\centering
\begin{minipage}[t]{0.37\linewidth}
\begin{ExamplePanel}[height=12cm,valign=center]{Input Image}
\centering
\begin{tikzpicture}
\node[anchor=south west,inner sep=0] (page) at (0,0)
  {\includegraphics[width=\linewidth,height=10.3cm,keepaspectratio]{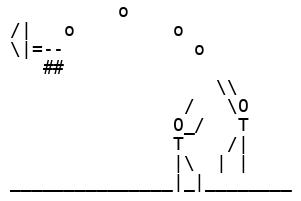}};
\end{tikzpicture}
\end{ExamplePanel}
\end{minipage}\hfill
\begin{minipage}[t]{0.61\linewidth}
\begin{ExamplePanel}[height=12cm]{Prompt}
\begin{Verbatim}
Classify the ASCII art in the image. Select exactly one label from the candidate list.

Candidate labels: ["chess", "boxing", "basketball", "bicycle"]

Directly output the selected label exactly as written. Do not output explanations, punctuation, JSON, Markdown, or any other text.
\end{Verbatim}
\end{ExamplePanel}
\end{minipage}
\par\vspace{3mm}
\begin{ExamplePanel}[colbacktitle=exampledark,coltitle=white]{OvisOCR3 Output: Image Classification}
\exampletext{examples/general_ocr/classification_330/output_display.txt}
\end{ExamplePanel}
\par\vspace{2mm}
\caption{Image classification of ASCII art. The response selects \texttt{basketball} from the supplied candidate labels.}
\label{fig:general_ocr_classification}
\end{figure}

Figures~\ref{fig:general_ocr_plain_ocr}--\ref{fig:general_ocr_position_vqa} illustrate the ten General OCR tasks in Section~\ref{sec:capability_general_ocr}. Each example shows the input image, task prompt, and model response. Green boxes denote predicted locations; orange boxes denote regions specified in the input.

\begin{figure}[!htbp]
\centering
\begin{minipage}[t]{0.37\linewidth}
\begin{ExamplePanel}[height=12cm,valign=center]{Input Image}
\centering
\begin{tikzpicture}
\node[anchor=south west,inner sep=0] (page) at (0,0)
  {\includegraphics[width=\linewidth,height=10.3cm,keepaspectratio]{examples/general_ocr/plain_ocr_5882/input.jpg}};
\end{tikzpicture}
\end{ExamplePanel}
\end{minipage}\hfill
\begin{minipage}[t]{0.61\linewidth}
\begin{ExamplePanel}[height=12cm]{Prompt}
\begin{Verbatim}
Read all the text in the image. Directly output the content.

Rules:
1. Preserve the original text exactly without translation, paraphrasing, correction, or completion.
2. Follow the natural human reading order.
3. Output plain text only. Use spaces and line breaks only to preserve text separation and reading order.
4. Do not output Markdown, HTML, JSON, bounding boxes, category labels, explanations, or code fences.
\end{Verbatim}
\end{ExamplePanel}
\end{minipage}
\par\vspace{3mm}
\begin{ExamplePanel}[colbacktitle=exampledark,coltitle=white]{OvisOCR3 Output: Plain OCR}
\exampletext{examples/general_ocr/plain_ocr_5882/output_display.txt}
\end{ExamplePanel}
\par\vspace{2mm}
\caption{Plain OCR on a comic page, transcribing speech-balloon text in reading order.}
\label{fig:general_ocr_plain_ocr}
\end{figure}
\FloatBarrier

\clearpage
\begin{figure}[!htbp]
\centering
\begin{minipage}[t]{0.37\linewidth}
\begin{ExamplePanel}[height=12cm,valign=center]{Input Image}
\centering
\begin{tikzpicture}
\node[anchor=south west,inner sep=0] (page) at (0,0)
  {\includegraphics[width=\linewidth,height=10.3cm,keepaspectratio]{examples/general_ocr/word_spotting_9520/input.jpg}};
\begin{scope}[x={(page.south east)},y={(page.north west)}]
\draw[green,line width=0.8pt] (0.27700,0.61900) rectangle (0.45200,0.65800);
\draw[green,line width=0.8pt] (0.47300,0.61100) rectangle (0.67500,0.65200);
\draw[green,line width=0.8pt] (0.12000,0.43100) rectangle (0.28500,0.56500);
\draw[green,line width=0.8pt] (0.32000,0.38200) rectangle (0.57200,0.43500);
\draw[green,line width=0.8pt] (0.62100,0.40600) rectangle (0.82800,0.54500);
\end{scope}
\end{tikzpicture}
\end{ExamplePanel}
\end{minipage}\hfill
\begin{minipage}[t]{0.61\linewidth}
\begin{ExamplePanel}[height=12cm]{Prompt}
\begin{Verbatim}
Locate and transcribe every readable word or text token in the image. Return a single JSON array in natural reading order.

Each element follows this schema:
`{"bbox": [x_min, y_min, x_max, y_max], "text": "string"}`

Rules:
1. Use word-level annotations. Each element must represent exactly one visible word or text token. Do not merge multiple words into a text line, paragraph, label, or table cell.
2. The bbox coordinates are integers scaled to [0, 1000).
3. Each bbox must tightly enclose its corresponding visible word or text token and must not include adjacent words or unrelated text regions.
4. Preserve the original text exactly without translation, paraphrasing, correction, or completion.
5. Do not add leading or trailing whitespace to text.
6. Include every readable word or text token and exclude non-text visual content.
7. Output strictly valid JSON only. Do not output Markdown, explanations, or code fences. Properly escape quotes, backslashes, and other special characters inside text.
\end{Verbatim}
\end{ExamplePanel}
\end{minipage}
\par\vspace{3mm}
\begin{ExamplePanel}[colbacktitle=exampledark,coltitle=white]{OvisOCR3 Output: Word-Level Text Spotting}
\exampletext{examples/general_ocr/word_spotting_9520/output_display.txt}
\end{ExamplePanel}
\par\vspace{2mm}
\caption{Word-level text spotting. Each of the five recognized words is paired with a predicted bounding box.}
\label{fig:general_ocr_word_spotting}
\end{figure}
\FloatBarrier

\clearpage
\FloatBarrier

\clearpage
\FloatBarrier

\clearpage
\FloatBarrier

\clearpage
\begin{figure}[!htbp]
\centering
\begin{minipage}[t]{0.37\linewidth}
\begin{ExamplePanel}[height=12cm,valign=center]{Input Image}
\centering
\begin{tikzpicture}
\node[anchor=south west,inner sep=0] (page) at (0,0)
  {\includegraphics[width=\linewidth,height=10.3cm,keepaspectratio]{examples/general_ocr/chart_2031/input.jpg}};
\end{tikzpicture}
\end{ExamplePanel}
\end{minipage}\hfill
\begin{minipage}[t]{0.61\linewidth}
\begin{ExamplePanel}[height=12cm]{Prompt}
\begin{Verbatim}
Parse all key information in the chart and output one valid JSON object using exactly this top-level schema:
{"title":"","source":"","x_title":"","y_title":"","values":{}}

Rules:
1. Preserve visible title, source, axis titles, series or category labels, x-axis labels, and values without explanation or inference.
2. Use the literal string "None" when the chart has no visible value for title, source, or an axis title. For a dual-axis chart, `y_title` may be an array of two strings in visual order; otherwise it must be one string.
3. `values` must be a non-empty object. Preserve the chart hierarchy and encode every scalar data value as a JSON string. An empty series name may be represented by the key "" when the chart has no visible series label.
4. Output JSON only. Do not output Python syntax, Markdown fences, comments, prose, or extra keys.
\end{Verbatim}
\end{ExamplePanel}
\end{minipage}
\par\vspace{3mm}
\begin{ExamplePanel}[colbacktitle=exampledark,coltitle=white]{OvisOCR3 Output: Chart Parsing}
\exampletext{examples/general_ocr/chart_2031/output_display.txt}
\end{ExamplePanel}
\par\vspace{2mm}
\caption{Chart parsing with metadata and per-category values organized as JSON. Numerical values are serialized as strings.}
\label{fig:general_ocr_chart}
\end{figure}
\FloatBarrier

\clearpage
\FloatBarrier

\clearpage
\FloatBarrier

\clearpage
\begin{figure}[!htbp]
\centering
\begin{minipage}[t]{0.37\linewidth}
\begin{ExamplePanel}[height=12cm,valign=center]{Input Image}
\centering
\begin{tikzpicture}
\node[anchor=south west,inner sep=0] (page) at (0,0)
  {\includegraphics[width=\linewidth,height=10.3cm,keepaspectratio]{examples/general_ocr/vqa_2505/input.jpg}};
\end{tikzpicture}
\end{ExamplePanel}
\end{minipage}\hfill
\begin{minipage}[t]{0.61\linewidth}
\begin{ExamplePanel}[height=12cm]{Prompt}
\begin{Verbatim}
Who wrote this book?

Please reason step by step, and put your final answer within \boxed{}.
\end{Verbatim}
\end{ExamplePanel}
\end{minipage}
\par\vspace{3mm}
\begin{ExamplePanel}[colbacktitle=exampledark,coltitle=white]{OvisOCR3 Output: Visual Question Answering}
\exampletext{examples/general_ocr/vqa_2505/output_display.txt}
\end{ExamplePanel}
\par\vspace{2mm}
\caption{Visual question answering from a book cover, with a brief explanation followed by the author's name as the final answer.}
\label{fig:general_ocr_vqa}
\end{figure}
\FloatBarrier

\clearpage
\begin{figure}[!htbp]
\centering
\begin{minipage}[t]{0.37\linewidth}
\begin{ExamplePanel}[height=12cm,valign=center]{Input Image}
\centering
\begin{tikzpicture}
\node[anchor=south west,inner sep=0] (page) at (0,0)
  {\includegraphics[width=\linewidth,height=10.3cm,keepaspectratio]{examples/general_ocr/position_vqa_1630/input.jpg}};
\begin{scope}[x={(page.south east)},y={(page.north west)}]
\draw[green,line width=0.8pt] (0.35100,0.74200) rectangle (0.48900,0.81500);
\end{scope}
\end{tikzpicture}
\end{ExamplePanel}
\end{minipage}\hfill
\begin{minipage}[t]{0.61\linewidth}
\begin{ExamplePanel}[height=12cm]{Prompt}
\begin{Verbatim}
what does the yellow sign say? Output the answer with 'answer' and 'bbox'. 'bbox' refers to the bounding box position of the 'answer' content in the image. The output format is "answer:gt, bbox:(x1,y1,x2,y2)", where the bbox is the coordinates of the top-left corner and the bottom-right corners. The 'bbox' should be normalized coordinates ranging from 0 to 1000 by the image width and height.
Your answer should be in the JSON format:
{
    "answer": "..",  # The answer
    "bbox": "(x1,y1,x2,y2)"     # The bounding box position of the 'answer'
}
\end{Verbatim}
\end{ExamplePanel}
\end{minipage}
\par\vspace{3mm}
\begin{ExamplePanel}[colbacktitle=exampledark,coltitle=white]{OvisOCR3 Output: Position-Aware VQA}
\exampletext{examples/general_ocr/position_vqa_1630/output_display.txt}
\end{ExamplePanel}
\par\vspace{2mm}
\caption{Position-aware VQA on a yellow sign. The response pairs the answer text with its predicted bounding box.}
\label{fig:general_ocr_position_vqa}
\end{figure}
\FloatBarrier

\clearpage
\subsection{Key Information Extraction}
\label{sec:qualitative_kie}

Figures~\ref{fig:example_kie_open_151}--\ref{fig:example_kie_schema_157} illustrate the three KIE modes in Section~\ref{sec:capability_kie}. Open extraction determines the keys from the document; specified-field extraction uses a supplied field list; schema-based extraction follows a supplied JSON template. Source images and response content are unchanged; only JSON indentation is adjusted for readability.

\begin{figure}[!htbp]
\centering
\begin{minipage}[t]{0.49\linewidth}
  \begin{ExamplePanel}[height=10.5cm,valign=center]{Input Image}
    \centering
    \includegraphics[width=\linewidth,height=9.2cm,keepaspectratio]{examples/kie_open_151/input.jpg}
  \end{ExamplePanel}
\end{minipage}\hfill
\begin{minipage}[t]{0.49\linewidth}
  \begin{ExamplePanel}[height=10.5cm]{Prompt}
    \exampletext{examples/kie_open_151/prompt.txt}
  \end{ExamplePanel}
\end{minipage}

\vspace{3mm}
\begin{ExamplePanel}[colbacktitle=exampledark,coltitle=white]{OvisOCR3 Output}
  \exampletext{examples/kie_open_151/output_display.json}
\end{ExamplePanel}
\par\vspace{2mm}
\caption{Open extraction from a CC-OCR~\citep{yang2024ccocr} receipt. The model extracts item amounts and payment totals using keys drawn from the receipt, without a supplied field list or JSON template. Blurring is present in the source image.}
\label{fig:example_kie_open_151}
\end{figure}

\clearpage
\begin{figure}[!htbp]
\centering
\begin{minipage}[t]{0.49\linewidth}
  \begin{ExamplePanel}[height=10.5cm,valign=center]{Input Image}
    \centering
    \includegraphics[width=\linewidth,height=9.2cm,keepaspectratio]{examples/kie_fields_1250/input.jpg}
  \end{ExamplePanel}
\end{minipage}\hfill
\begin{minipage}[t]{0.49\linewidth}
  \begin{ExamplePanel}[height=10.5cm]{Prompt}
    \exampletext{examples/kie_fields_1250/prompt.txt}
  \end{ExamplePanel}
\end{minipage}

\vspace{3mm}
\begin{ExamplePanel}[colbacktitle=exampledark,coltitle=white]{OvisOCR3 Output}
  \exampletext{examples/kie_fields_1250/output_display.json}
\end{ExamplePanel}
\par\vspace{2mm}
\caption{Specified-field extraction from an OCRBench v2~\citep{fu2025ocrbenchv2} registration form. The response contains exactly the five requested keys. Missing applicant telephone and fax values remain empty, rather than being filled with the organizer's contact details.}
\label{fig:example_kie_fields_1250}
\end{figure}

\clearpage
\begin{figure}[!htbp]
\centering
\begin{minipage}[t]{0.49\linewidth}
  \begin{ExamplePanel}[height=10.5cm,valign=center]{Input Image}
    \centering
    \includegraphics[width=\linewidth,height=9.2cm,keepaspectratio]{examples/kie_schema_157/input.jpg}
  \end{ExamplePanel}
\end{minipage}\hfill
\begin{minipage}[t]{0.49\linewidth}
  \begin{ExamplePanel}[height=10.5cm]{Prompt}
    \exampletext{examples/kie_schema_157/prompt.txt}
  \end{ExamplePanel}
\end{minipage}

\vspace{3mm}
\begin{ExamplePanel}[colbacktitle=exampledark,coltitle=white]{OvisOCR3 Output}
  \exampletext{examples/kie_schema_157/output_display.json}
\end{ExamplePanel}
\par\vspace{2mm}
\caption{Schema-based extraction from a CC-OCR~\citep{yang2024ccocr} receipt. Each item's name, quantity, and price are grouped in one record; the total amount, cash payment, and change are stored in a separate nested object. Blurring is present in the source image.}
\label{fig:example_kie_schema_157}
\end{figure}

\clearpage

\subsection{Document Figure Captioning}
\label{sec:qualitative_captioning}
\begin{figure}[!htbp]
\centering
\includegraphics[width=\linewidth,height=.82\textheight,keepaspectratio]{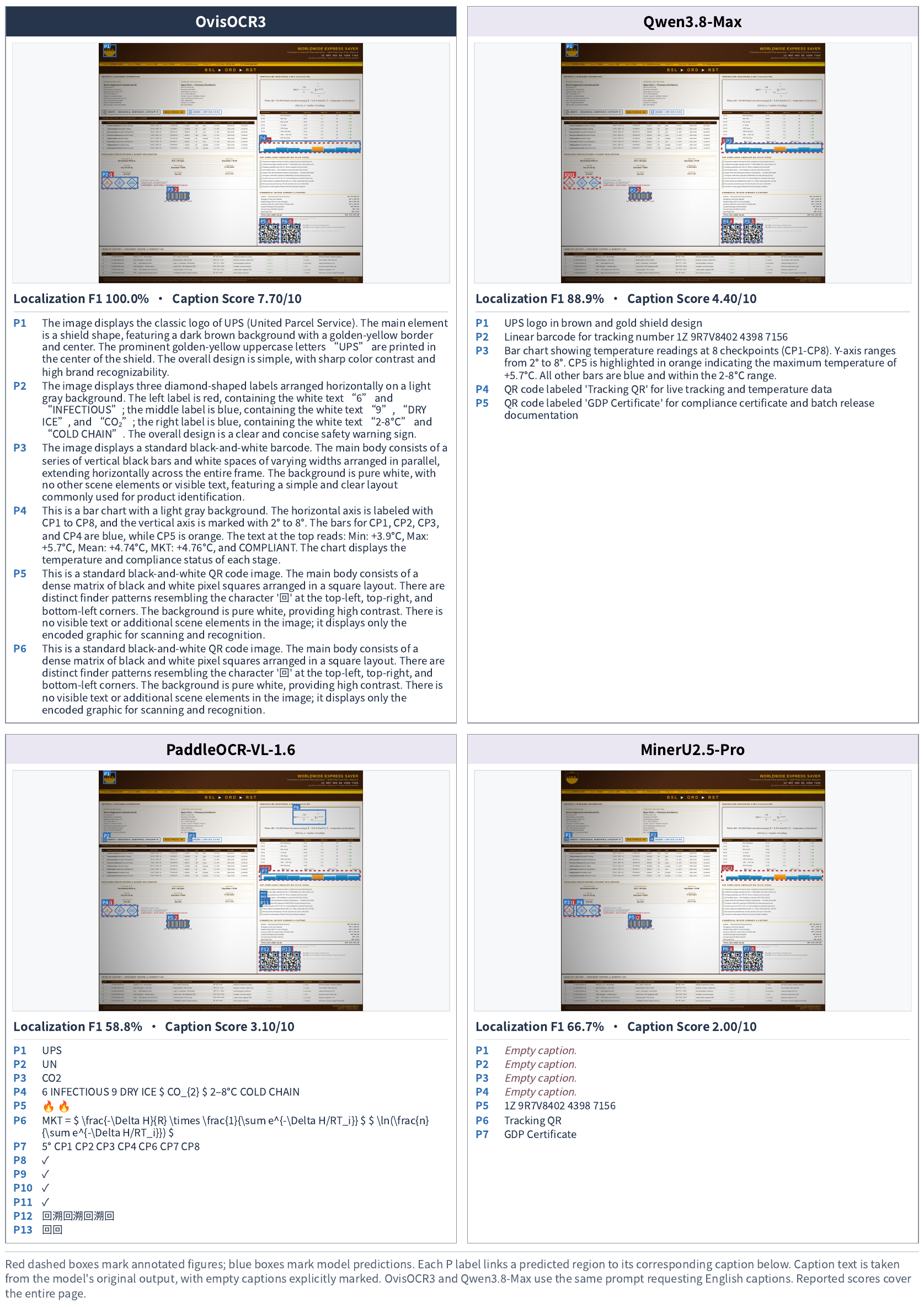}
\caption{Comparison on a page with multiple visual regions.}
\label{fig:docfigcap_comparison_0512}
\end{figure}

\begin{figure}[!htbp]
\centering
\includegraphics[width=\linewidth,height=.82\textheight,keepaspectratio]{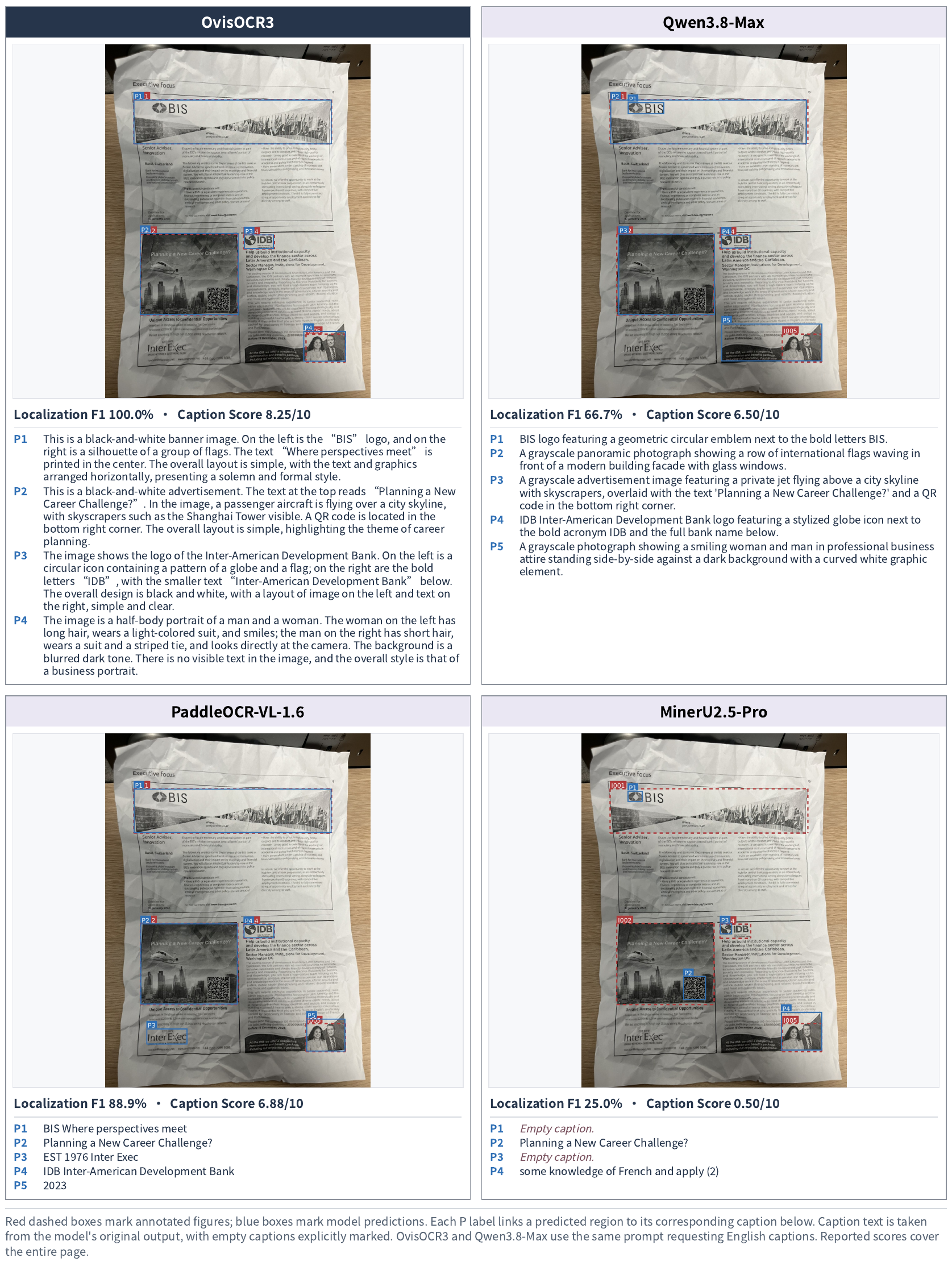}
\caption{Comparison on a photographed magazine page.}
\label{fig:docfigcap_comparison_0346}
\end{figure}

\begin{figure}[!htbp]
\centering
\includegraphics[width=\linewidth,height=.82\textheight,keepaspectratio]{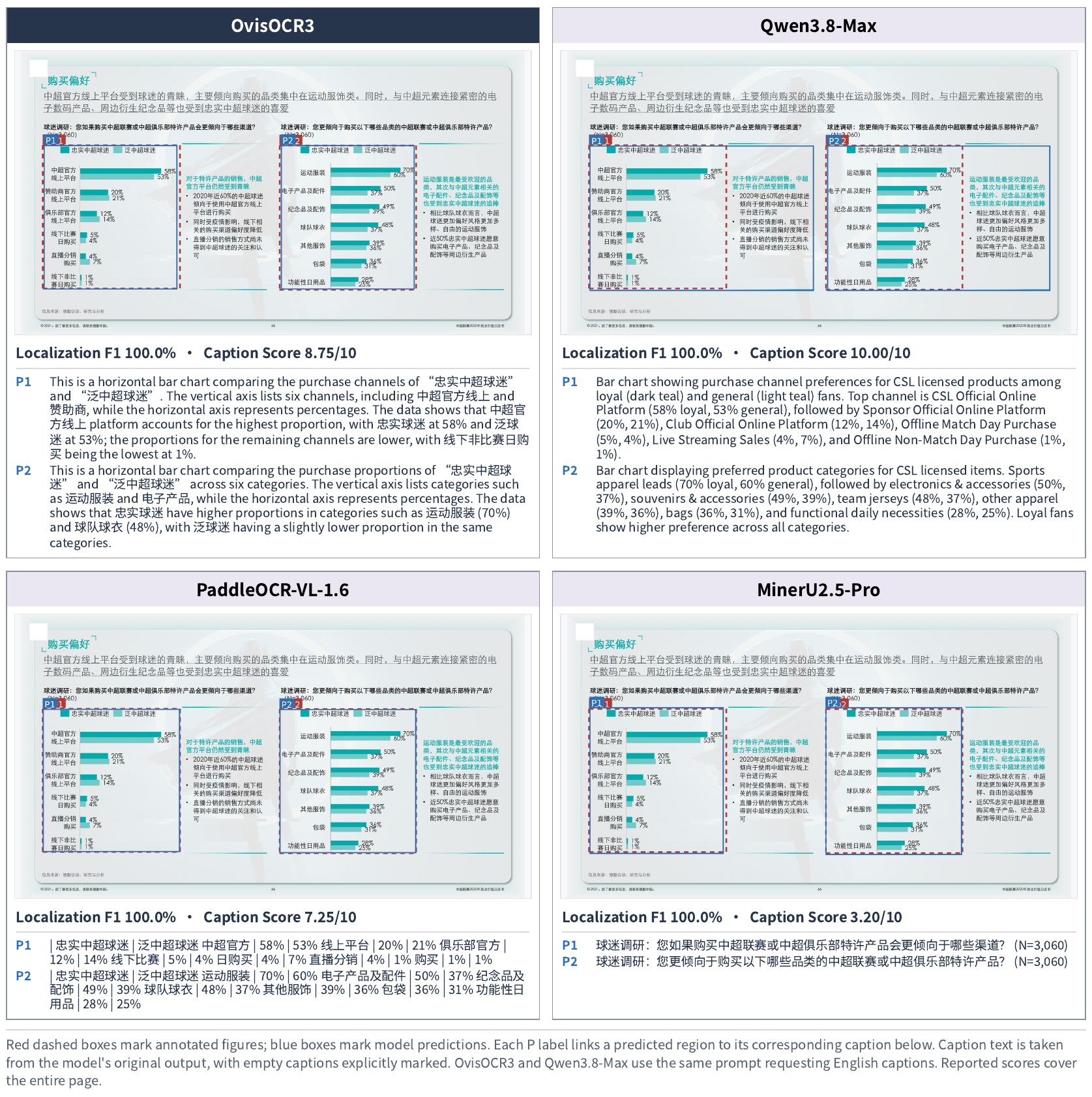}
\caption{Comparison on a page with two charts.}
\label{fig:docfigcap_comparison_0022}
\end{figure}

\begin{figure}[!htbp]
\centering
\includegraphics[width=\linewidth,height=.82\textheight,keepaspectratio]{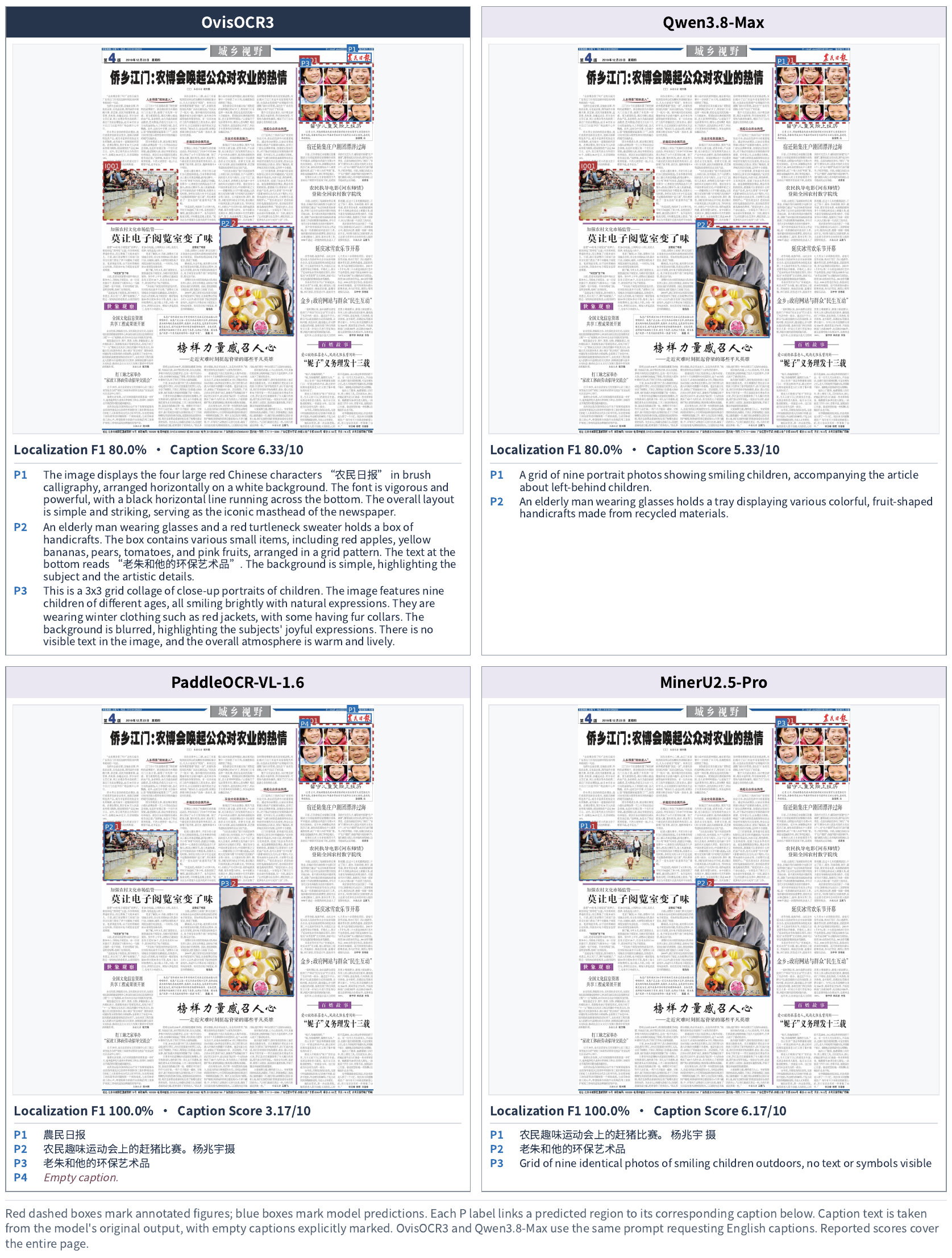}
\caption{Comparison on a newspaper page with individual images and an image group.}
\label{fig:docfigcap_comparison_0224}
\end{figure}

\begin{figure}[!htbp]
\centering
\includegraphics[width=\linewidth,height=.82\textheight,keepaspectratio]{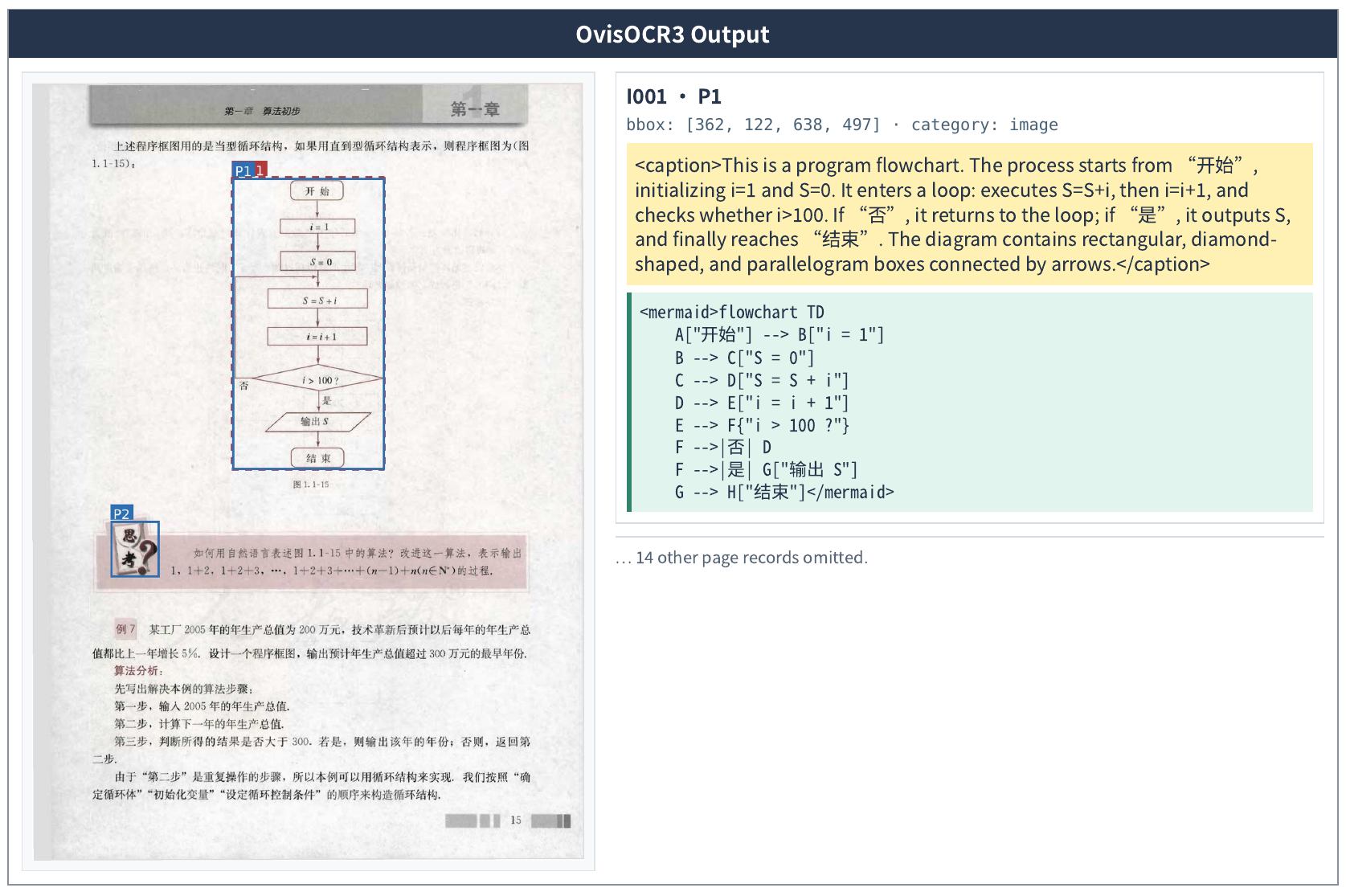}
\caption{\ours{} generates a caption and a Mermaid representation for an algorithm flowchart. The other 14 records in the full-page response are omitted from this display.}
\label{fig:docfigcap_flowchart_0112}
\end{figure}

\begin{figure}[!htbp]
\centering
\includegraphics[width=\linewidth,height=.82\textheight,keepaspectratio]{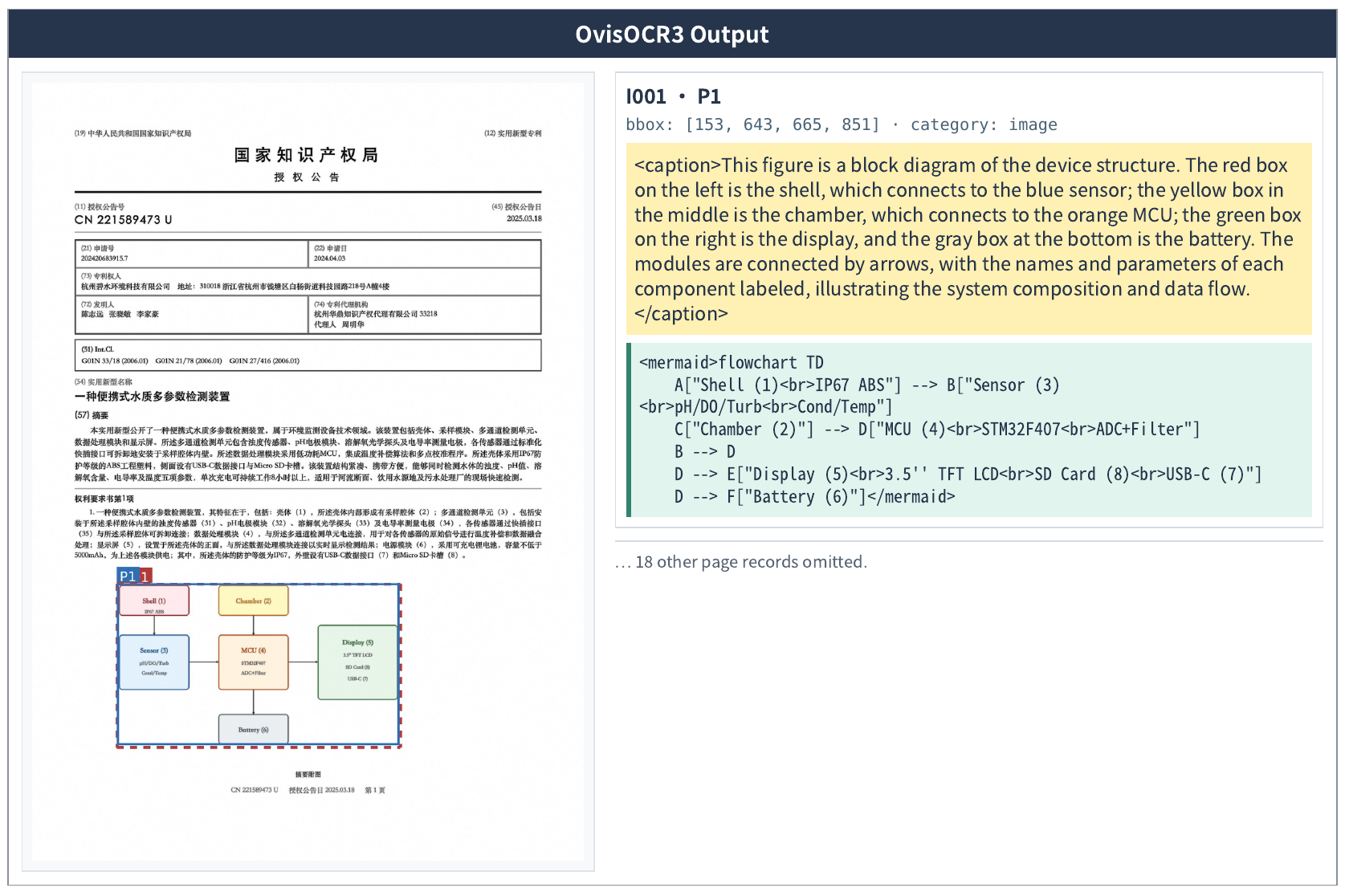}
\caption{\ours{} generates a caption and a Mermaid representation for a device block diagram. The other 18 records in the full-page response are omitted from this display.}
\label{fig:docfigcap_flowchart_0050}
\end{figure}

\begin{figure}[!htbp]
\centering
\includegraphics[width=\linewidth,height=.82\textheight,keepaspectratio]{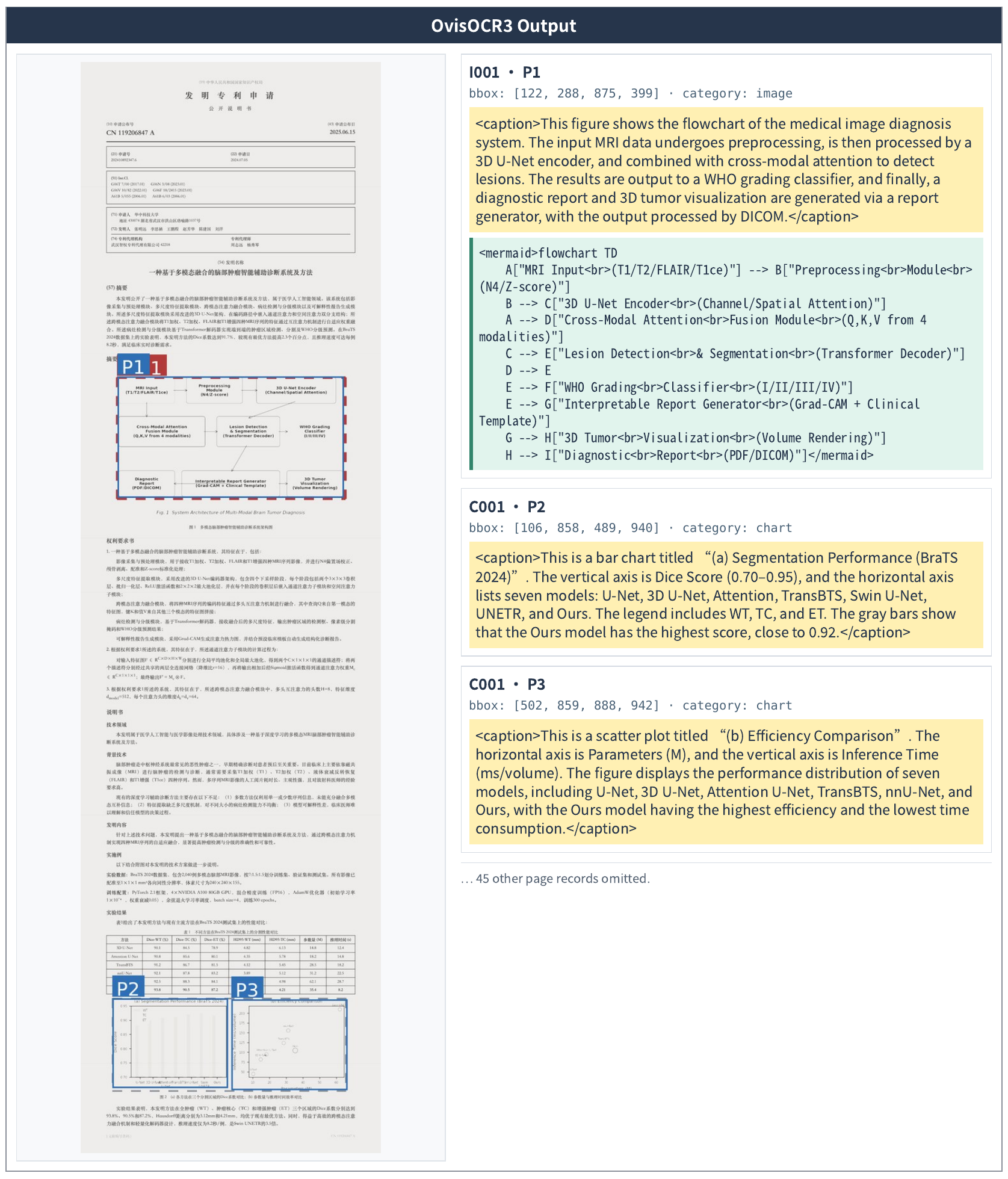}
\caption{\ours{} generates a caption and a Mermaid representation for a system architecture diagram. The other 45 records in the full-page response are omitted from this display.}
\label{fig:docfigcap_flowchart_0171}
\end{figure}

Figures~\ref{fig:docfigcap_comparison_0512}--\ref{fig:docfigcap_comparison_0224} compare \ours{} with three baselines on figure localization and captioning. Figures~\ref{fig:docfigcap_flowchart_0112}--\ref{fig:docfigcap_flowchart_0171} show \ours{} captions and Mermaid code for flowcharts and block diagrams.


\FloatBarrier

\clearpage
\subsection{Cross-Page Table Reconstruction}
\label{sec:qualitative_crosspage}

\begin{figure}[!htbp]
\centering
\begin{ExamplePanel}{Complete Input Pages}
  \begin{minipage}[t]{0.49\linewidth}
    \centering\textsf{\small Input page 1}\par\smallskip
    \includegraphics[width=\linewidth,height=9cm,keepaspectratio]{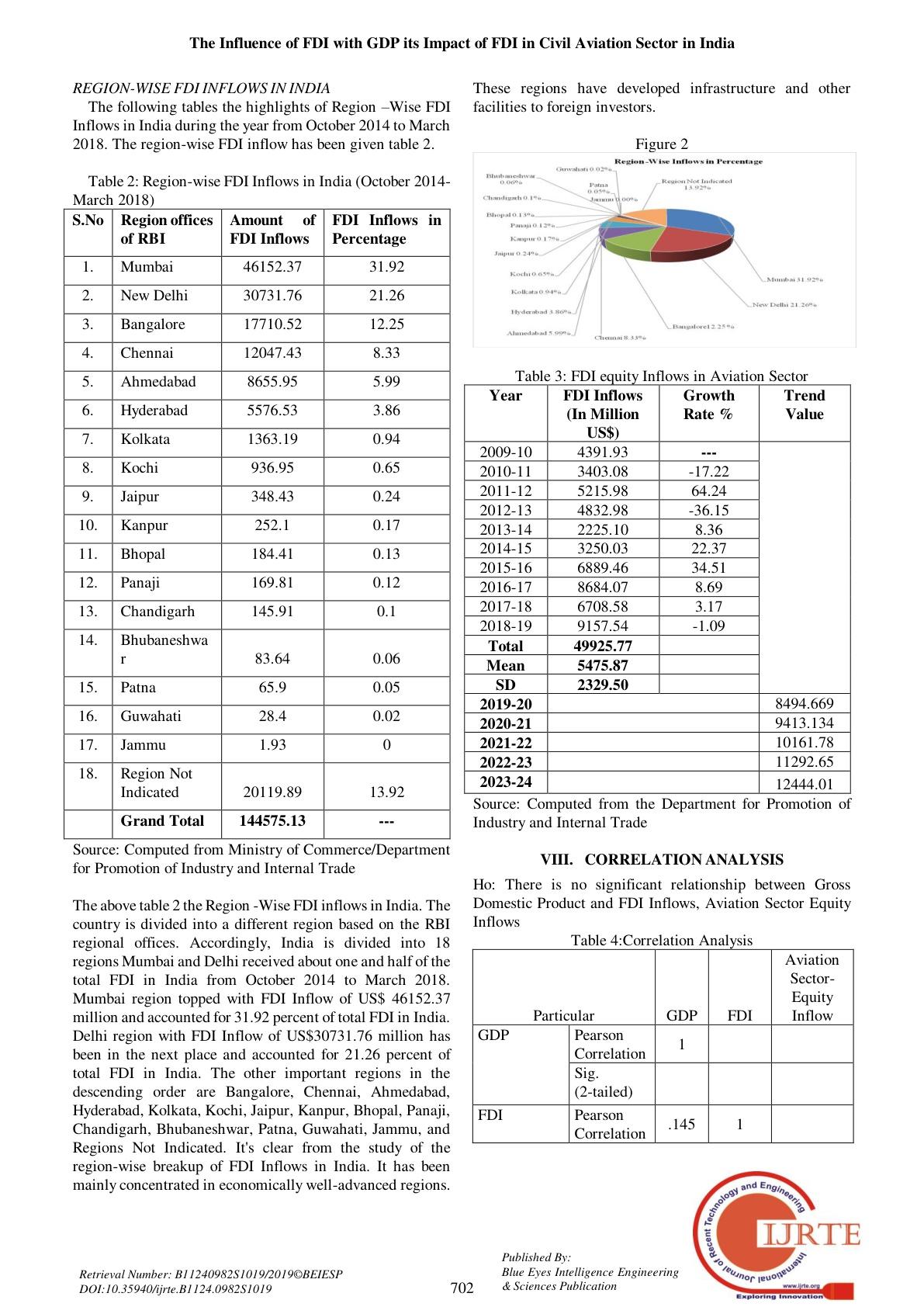}
  \end{minipage}\hfill
  \begin{minipage}[t]{0.49\linewidth}
    \centering\textsf{\small Input page 2}\par\smallskip
    \includegraphics[width=\linewidth,height=9cm,keepaspectratio]{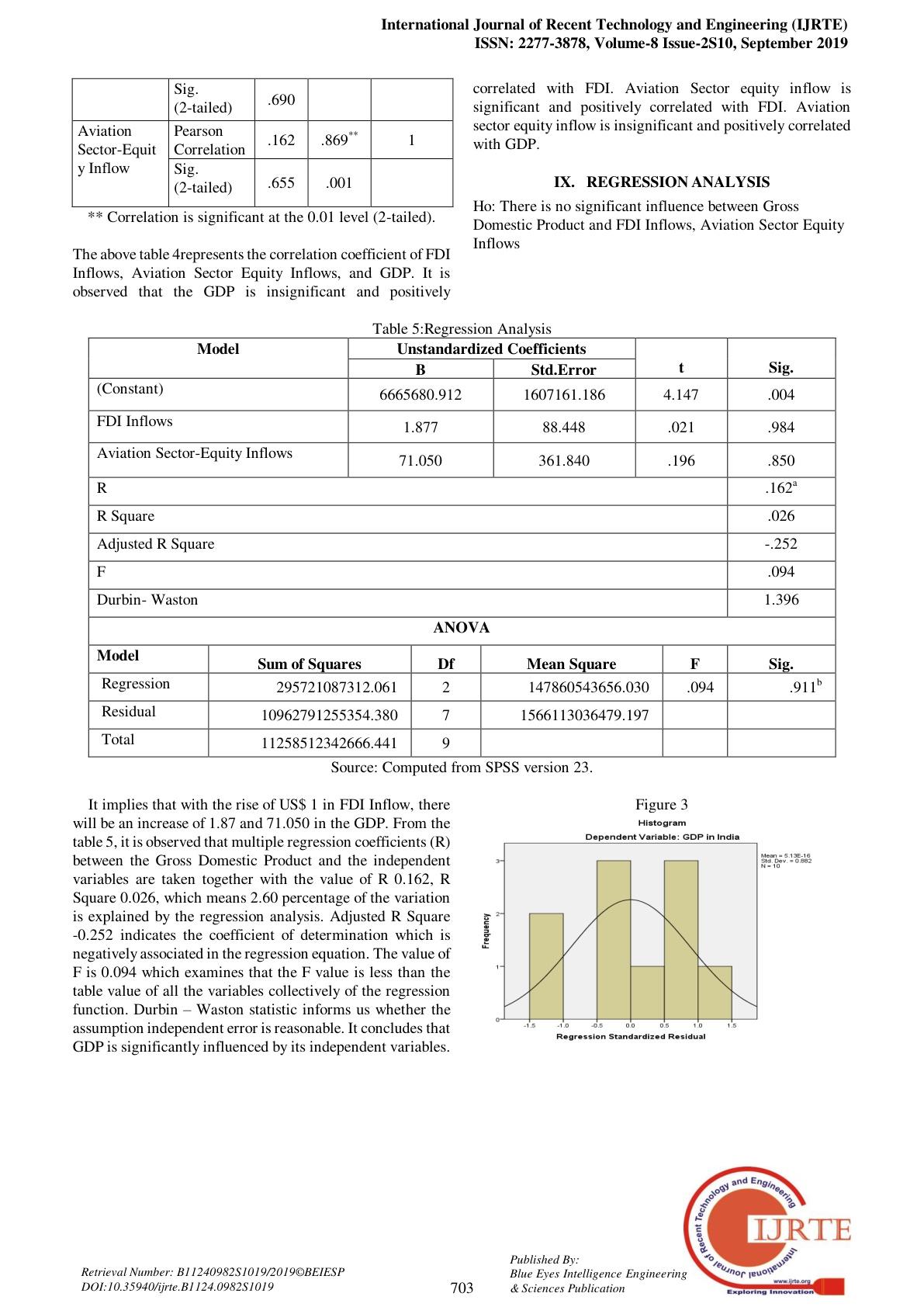}
  \end{minipage}
\end{ExamplePanel}
\par\vspace{3mm}
\begin{ExamplePanel}{Table Fragments (Enlarged)}
\begin{minipage}[t]{0.49\linewidth}\centering\textsf{\small Page 1 fragment}\par\smallskip
\includegraphics[width=\linewidth]{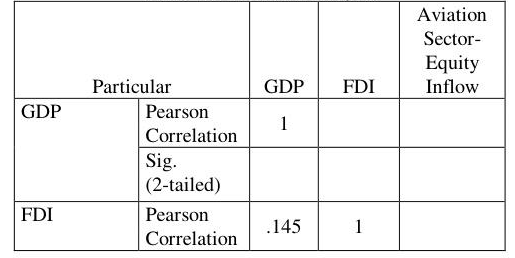}
\end{minipage}\hfill
\begin{minipage}[t]{0.49\linewidth}\centering\textsf{\small Page 2 fragment}\par\smallskip
\includegraphics[width=\linewidth]{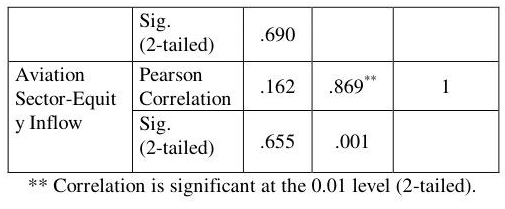}
\end{minipage}
\end{ExamplePanel}
\caption{Cross-page correlation table from MPDocBench-Parse (document pages 4--5). The FDI label spans two rows separated by the page boundary. Full pages and enlarged fragments show the source structure; Figure~\ref{fig:crosspage_comparison_1} compares the reconstructions.}
\label{fig:crosspage_inputs_1}
\end{figure}

\begin{figure}[p]
\centering
\begin{minipage}[t]{0.49\linewidth}
\begin{ExamplePanel}[height=8.8cm,valign=center]{MinerU2.5-Pro}
\centering
\includegraphics[width=\linewidth,height=7.1cm,keepaspectratio]{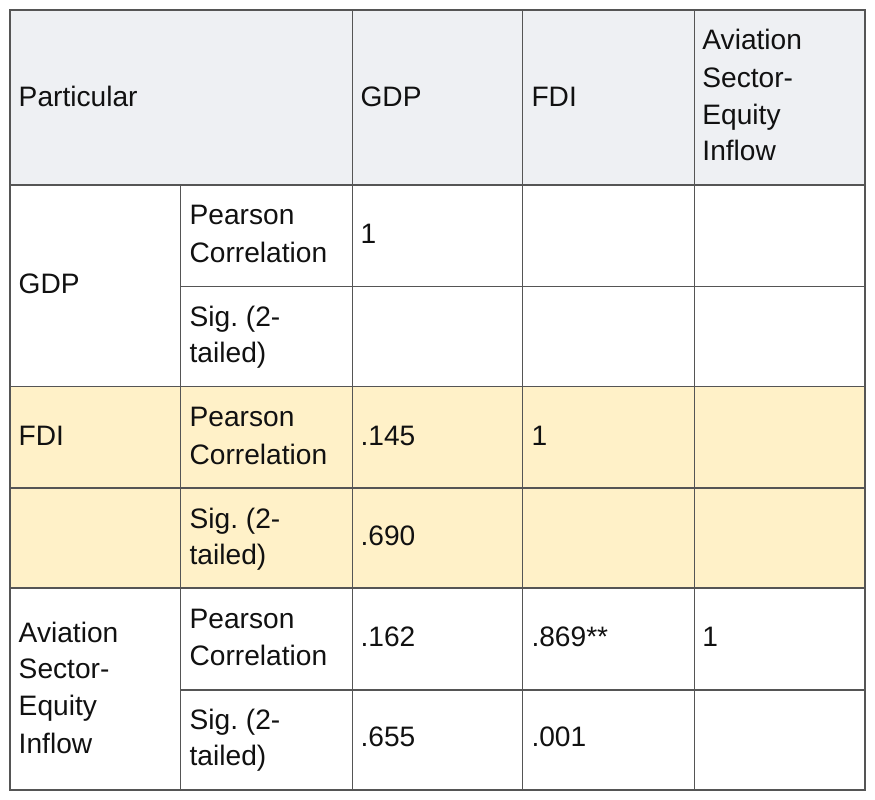}\par\vspace{3mm}
\textsf{\footnotesize Splits the shared FDI label cell.}
\end{ExamplePanel}
\end{minipage}\hfill
\begin{minipage}[t]{0.49\linewidth}
\begin{ExamplePanel}[height=8.8cm,valign=center]{Qwen3.8-Max}
\centering
\includegraphics[width=\linewidth,height=7.1cm,keepaspectratio]{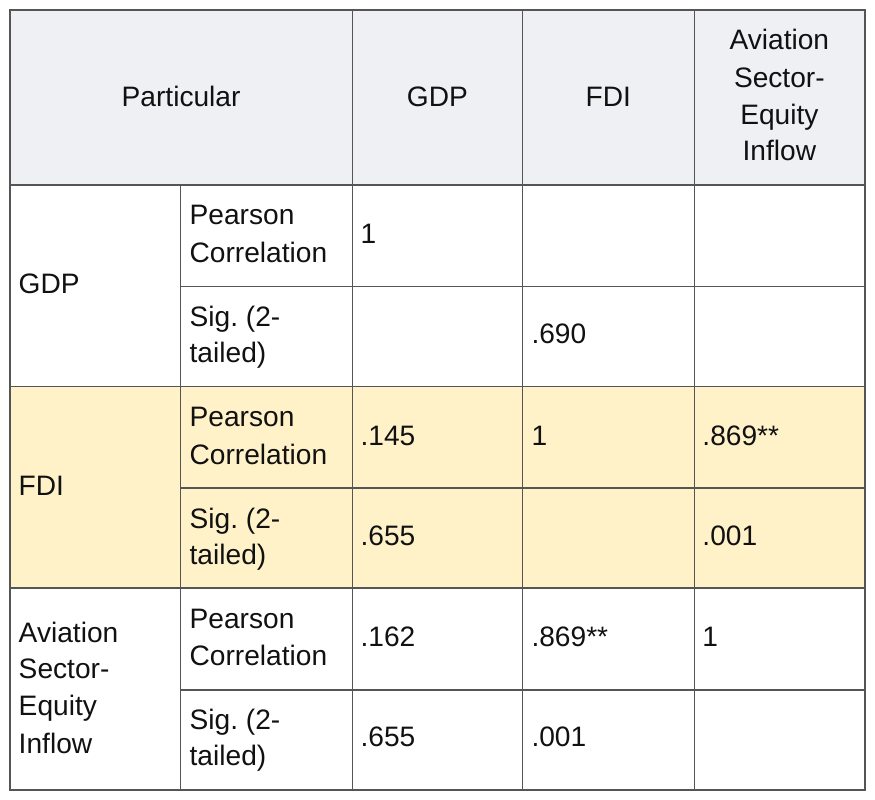}\par\vspace{3mm}
\textsf{\footnotesize Misplaces and duplicates numerical values.}
\end{ExamplePanel}
\end{minipage}
\par\vspace{3mm}
\begin{minipage}[t]{0.49\linewidth}
\begin{ExamplePanel}[height=8.8cm,valign=center]{Logics-Parsing-V3}
\centering
\includegraphics[width=\linewidth,height=7.1cm,keepaspectratio]{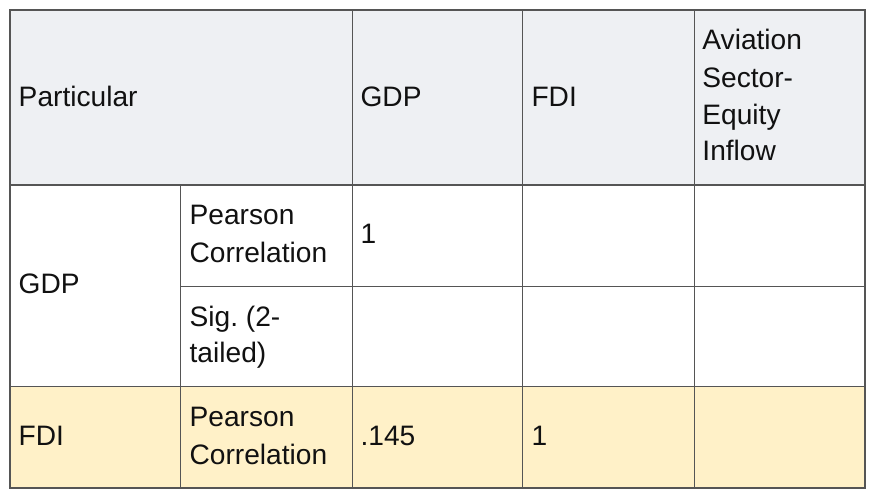}\par\vspace{3mm}
\textsf{\footnotesize Omits the second-page continuation.}
\end{ExamplePanel}
\end{minipage}\hfill
\begin{minipage}[t]{0.49\linewidth}
\begin{ExamplePanel}[height=8.8cm,valign=center]{OvisOCR3}
\centering
\includegraphics[width=\linewidth,height=7.1cm,keepaspectratio]{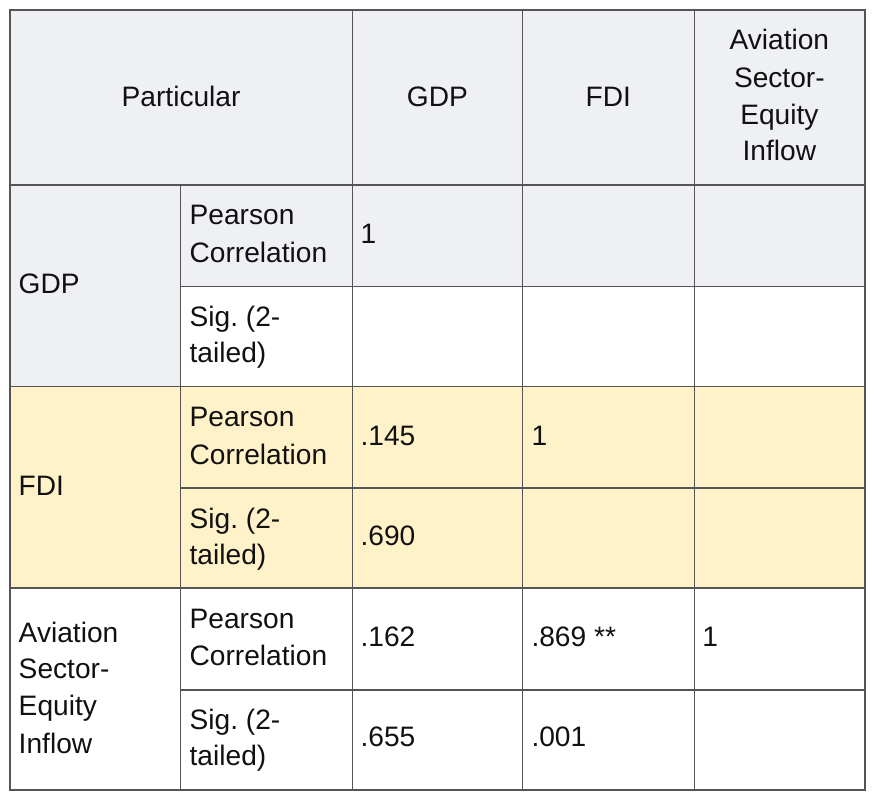}\par\vspace{3mm}
\textsf{\footnotesize Preserves the shared cell and value alignment.}
\end{ExamplePanel}
\end{minipage}
\par\vspace{3mm}
\caption{Reconstructions of the correlation table in Figure~\ref{fig:crosspage_inputs_1}. \mbox{MinerU2.5-Pro} splits the shared FDI label cell, while \mbox{Qwen3.8-Max} misplaces and duplicates numerical values. \mbox{Logics-Parsing-V3} omits the second-page fragment. \mbox{\ours{}} preserves both the shared FDI cell and the placement of numerical values. Shading marks the rows spanning the page boundary.}
\label{fig:crosspage_comparison_1}
\end{figure}

\begin{figure}[p]
\centering
\begin{ExamplePanel}{Complete Input Pages}
  \begin{minipage}[t]{0.49\linewidth}
    \centering\textsf{\small Input page 1}\par\smallskip
    \includegraphics[width=\linewidth,height=8cm,keepaspectratio]{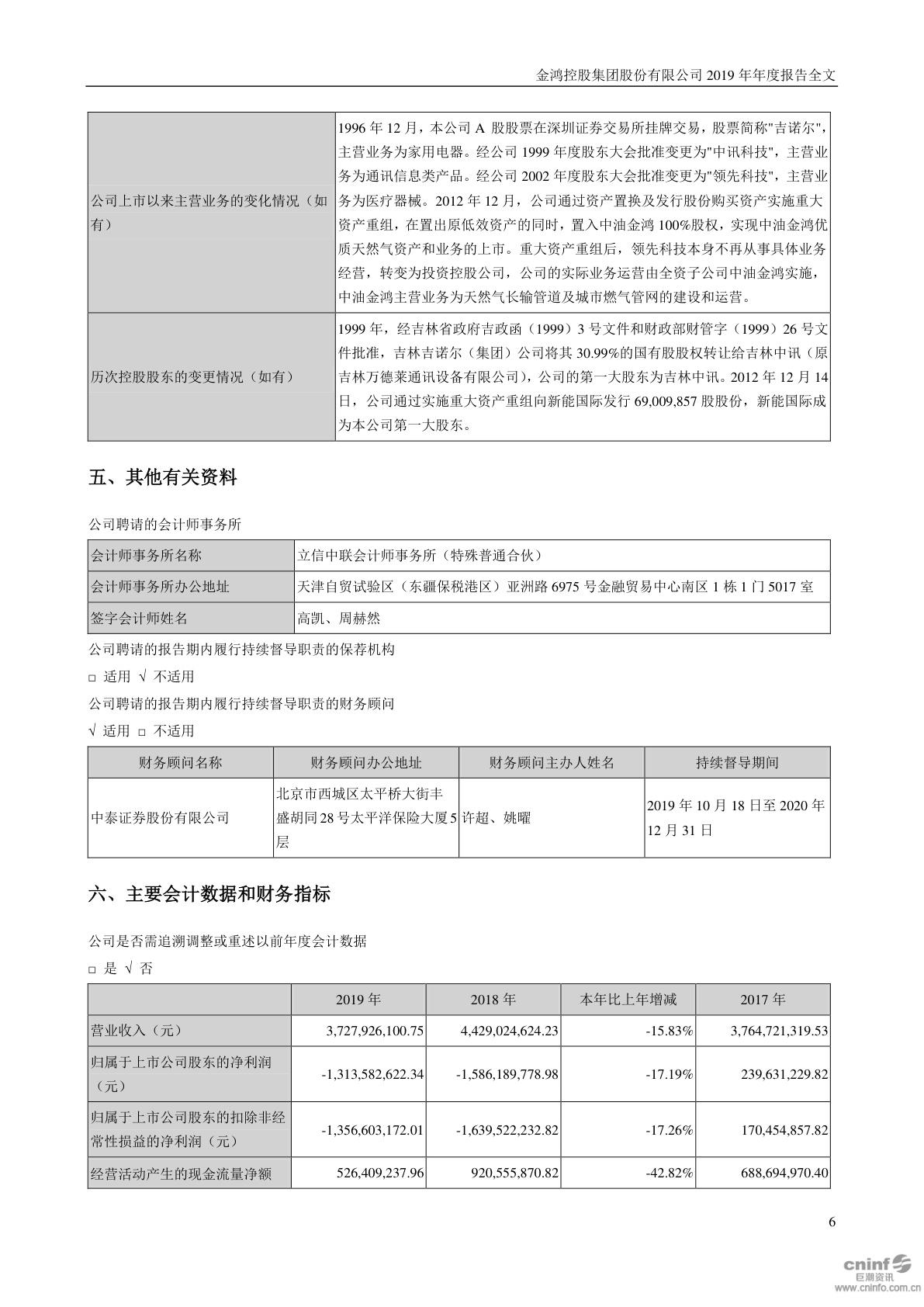}
  \end{minipage}\hfill
  \begin{minipage}[t]{0.49\linewidth}
    \centering\textsf{\small Input page 2}\par\smallskip
    \includegraphics[width=\linewidth,height=8cm,keepaspectratio]{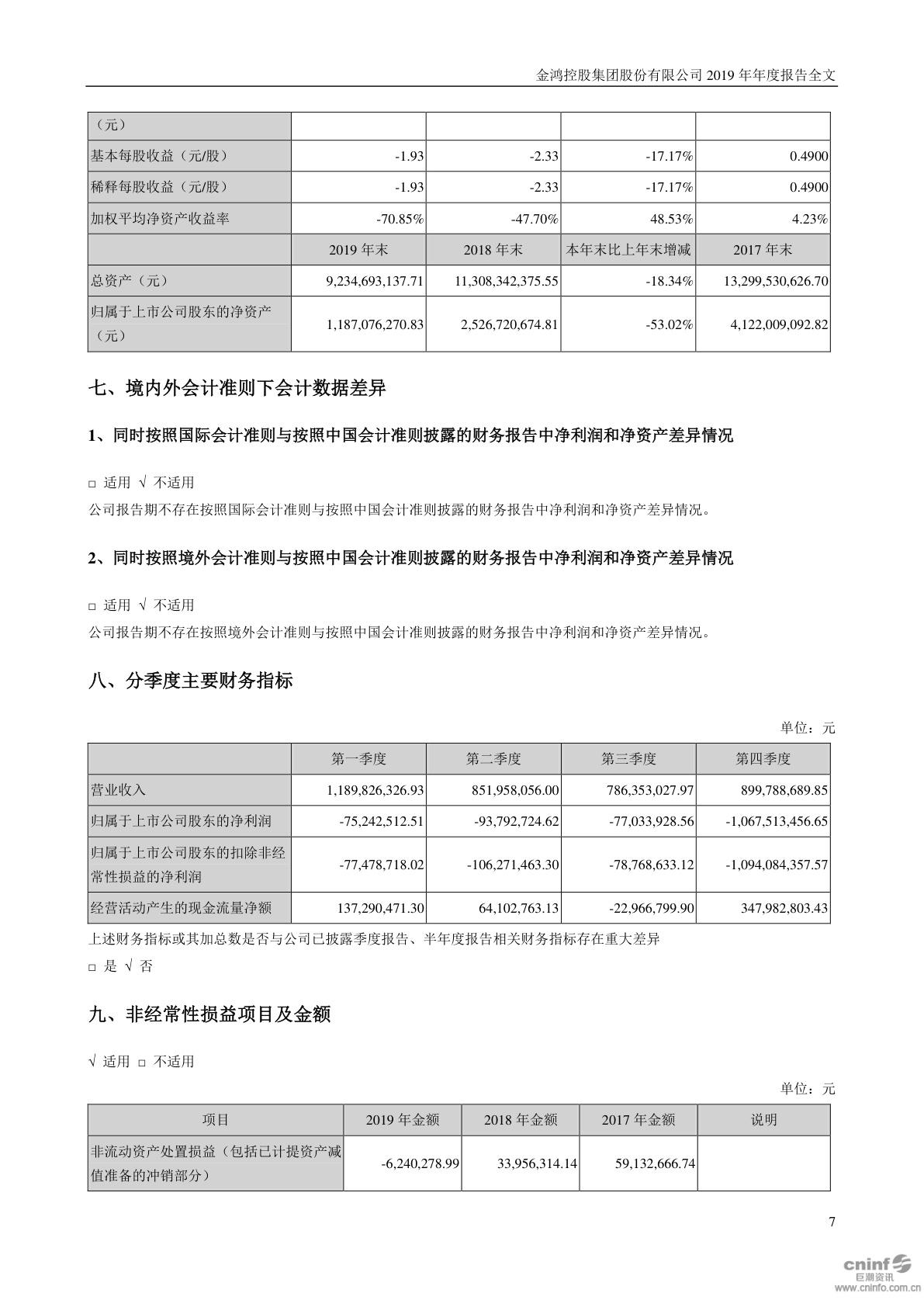}
  \end{minipage}
\end{ExamplePanel}
\par\vspace{3mm}
\begin{ExamplePanel}{Table Fragments (Enlarged)}
\textsf{\small Page 1 fragment}\par\smallskip
\includegraphics[width=0.88\linewidth,height=5.3cm,keepaspectratio]{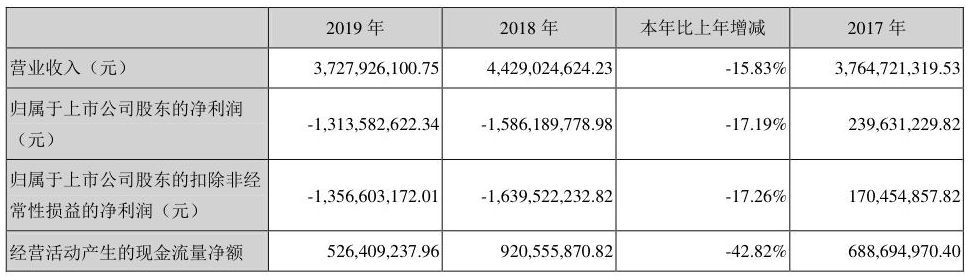}\par
\textsf{\small Page 2 fragment}\par\smallskip
\includegraphics[width=0.88\linewidth,height=5.3cm,keepaspectratio]{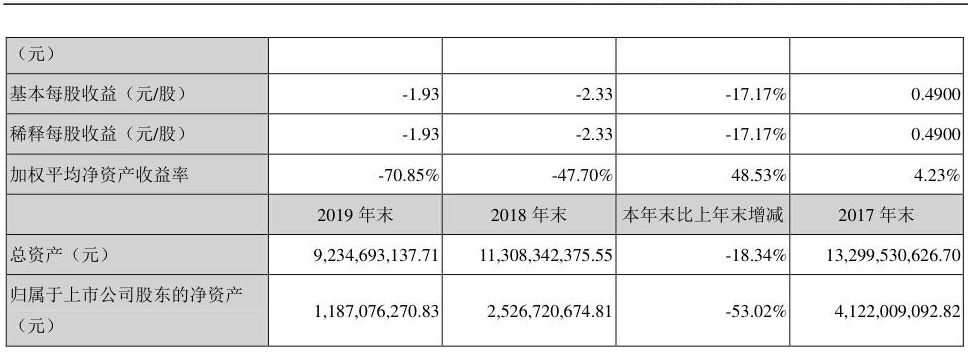}\par
\end{ExamplePanel}
\caption{Cross-page financial table from MPDocBench-Parse (document pages 6--7). The currency unit at the start of the second page completes the cash-flow label on the first page. Full pages and enlarged fragments show the source structure; Figure~\ref{fig:crosspage_comparison_2} compares the reconstructions.}
\label{fig:crosspage_inputs_2}
\end{figure}

\begin{figure}[p]
\centering
\begin{minipage}[t]{0.49\linewidth}
\begin{ExamplePanel}[height=8.8cm,valign=center]{MinerU2.5-Pro}
\centering
\includegraphics[width=\linewidth,height=7.1cm,keepaspectratio]{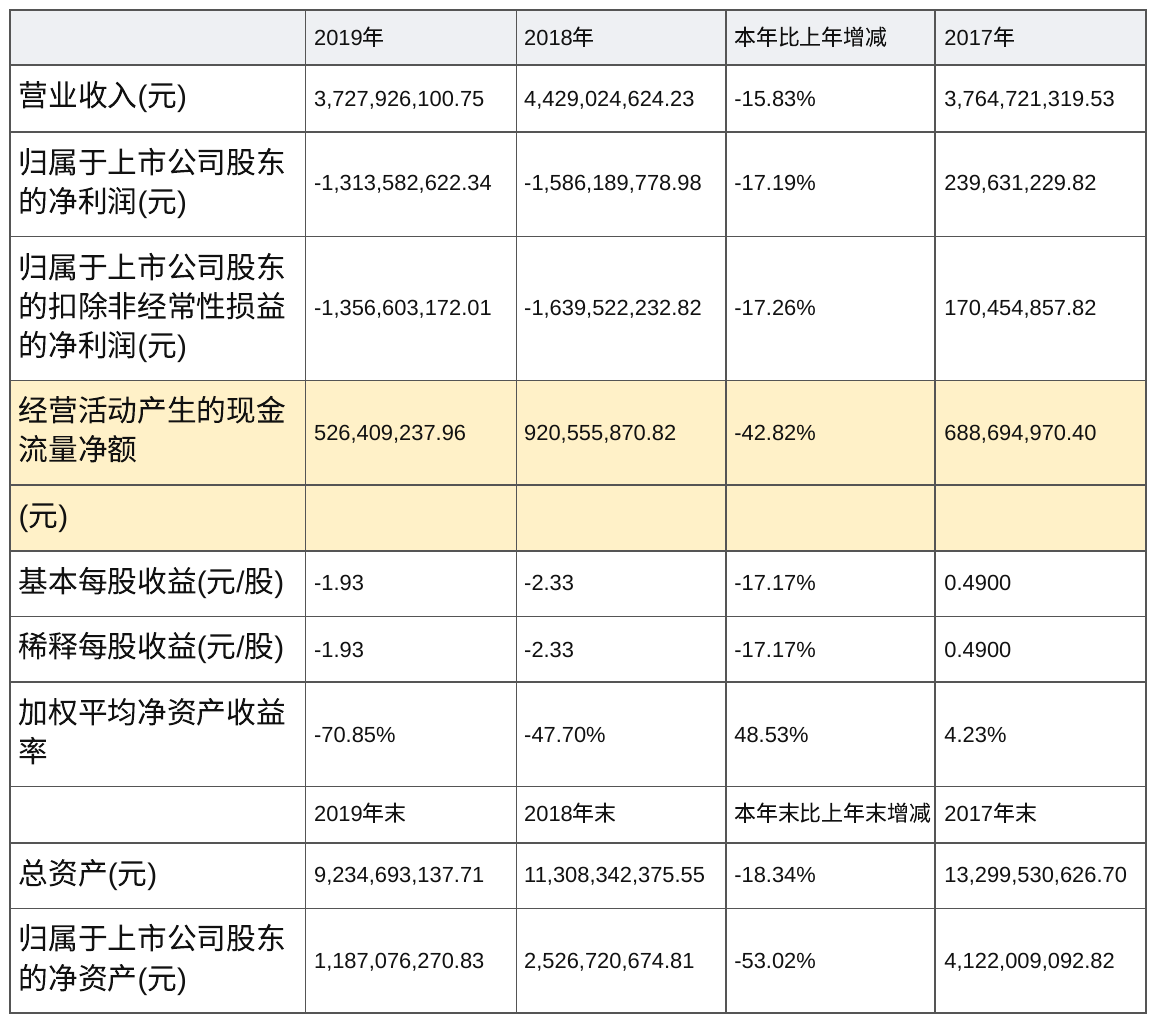}\par\vspace{3mm}
\textsf{\footnotesize Treats the currency unit as a separate row.}
\end{ExamplePanel}
\end{minipage}\hfill
\begin{minipage}[t]{0.49\linewidth}
\begin{ExamplePanel}[height=8.8cm,valign=center]{Qwen3.8-Max}
\centering
\includegraphics[width=\linewidth,height=7.1cm,keepaspectratio]{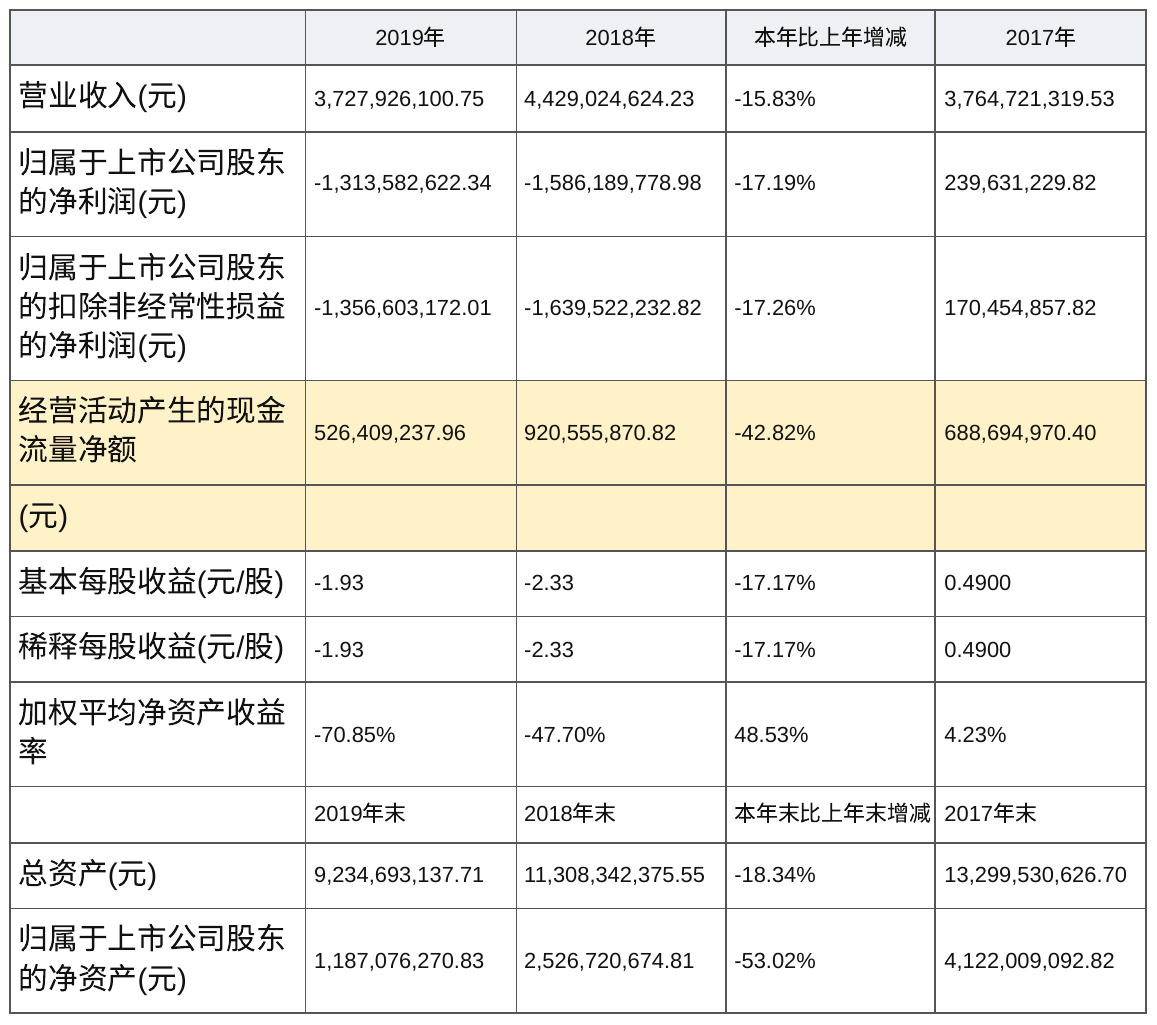}\par\vspace{3mm}
\textsf{\footnotesize Treats the currency unit as a separate row.}
\end{ExamplePanel}
\end{minipage}
\par\vspace{3mm}
\begin{minipage}[t]{0.49\linewidth}
\begin{ExamplePanel}[height=8.8cm,valign=center]{Logics-Parsing-V3}
\centering
\includegraphics[width=\linewidth,height=7.1cm,keepaspectratio]{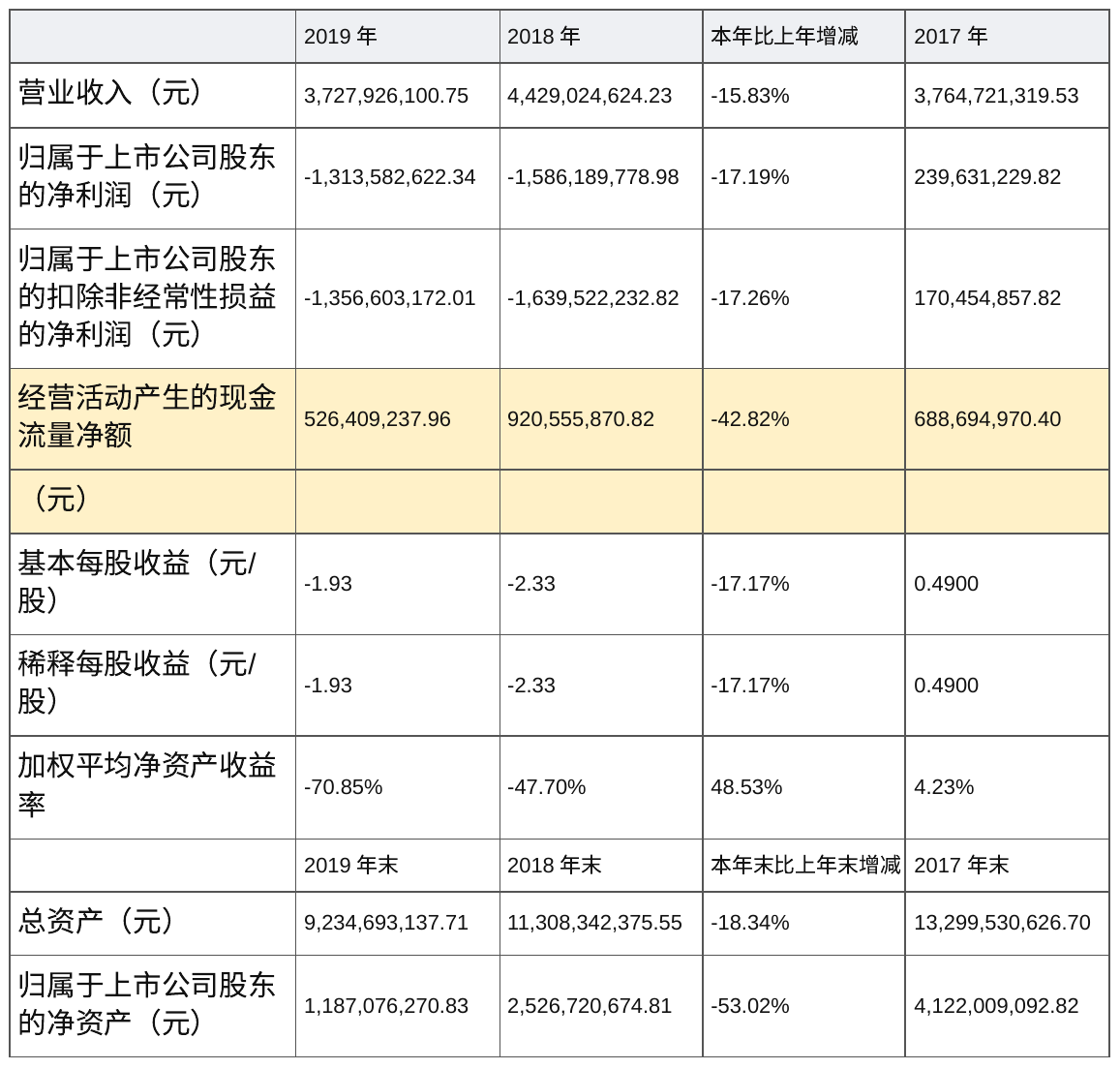}\par\vspace{3mm}
\textsf{\footnotesize Treats the currency unit as a separate row.}
\end{ExamplePanel}
\end{minipage}\hfill
\begin{minipage}[t]{0.49\linewidth}
\begin{ExamplePanel}[height=8.8cm,valign=center]{OvisOCR3}
\centering
\includegraphics[width=\linewidth,height=7.1cm,keepaspectratio]{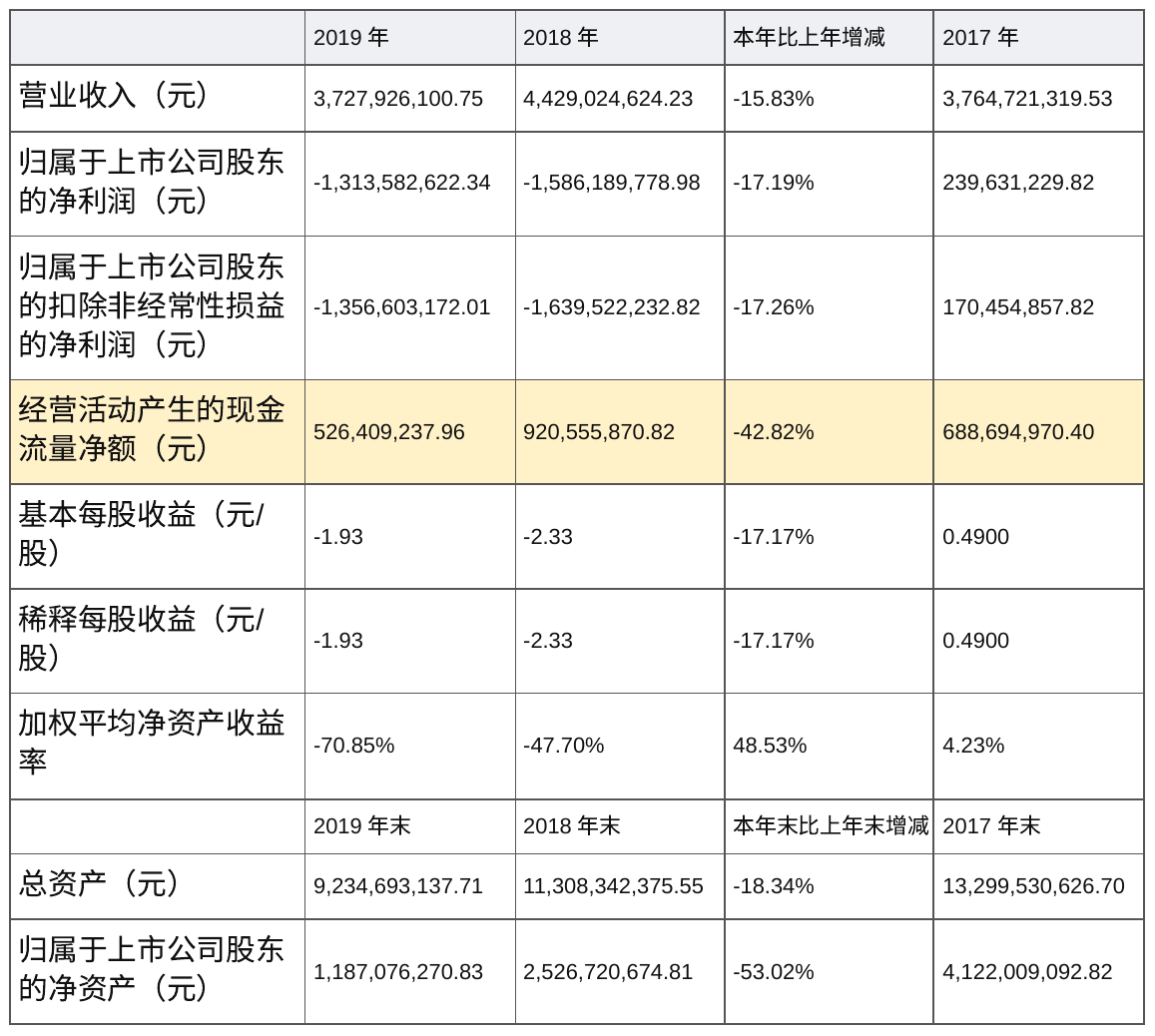}\par\vspace{3mm}
\textsf{\footnotesize Merges the unit into the cash-flow label.}
\end{ExamplePanel}
\end{minipage}
\par\vspace{3mm}
\caption{Reconstructions of the financial table in Figure~\ref{fig:crosspage_inputs_2}. \mbox{MinerU2.5-Pro}, \mbox{Qwen3.8-Max}, and \mbox{Logics-Parsing-V3} each place the currency unit in a separate row, leaving four numerical cells empty. \mbox{\ours{}} attaches the unit to the preceding cash-flow label, avoiding the extra row. Shading marks the content split by the page boundary.}
\label{fig:crosspage_comparison_2}
\end{figure}

Figures~\ref{fig:crosspage_inputs_1}--\ref{fig:crosspage_comparison_2} compare reconstructions of two cross-page tables from MPDocBench-Parse. Full source pages and enlarged fragments accompany the four-model comparisons, highlighting differences in merged-cell structure and numerical alignment.

\clearpage
\clearpage

\clearpage
\clearpage
\clearpage

\end{document}